\documentclass{article}

\usepackage[preprint]{neurips_2026}

\usepackage[utf8]{inputenc} 
\usepackage[T1]{fontenc}    
\usepackage{hyperref}       
\usepackage{url}            
\usepackage{booktabs}       
\usepackage{amsfonts}       
\usepackage{nicefrac}       
\usepackage{microtype}      
\usepackage{xcolor}         
\usepackage{microtype}
\usepackage{graphicx}
\usepackage{subcaption}
\usepackage{multirow}
\usepackage{amsmath}
\usepackage{amssymb}
\usepackage{mathtools}
\usepackage{amsthm}
\usepackage{adjustbox}

\title{Asymmetric Peak-Aware Loss for Peak-Critical Time Series Forecasting}

\author{%
    \textbf{Theivaprakasham Hari}$^{1}$\thanks{Corresponding author.}, \textbf{Yanan Xin}$^{1}$, \textbf{Winnie Daamen}$^{1}$,\\
    \textbf{Serge Paul Hoogendoorn}$^{1}$, \textbf{Sascha Hoogendoorn-Lanser}$^{2}$\\
    $^{1}$ Department of Transport and Planning, Faculty of Civil Engineering and Geosciences\\
    $^{2}$ Mobility Innovation Centre\\
    Delft University of Technology, Delft, The Netherlands 2628 CN\\
    \texttt{t.hari@tudelft.nl; yanan.xin@tudelft.nl; w.daamen@tudelft.nl;}\\
    \texttt{s.p.hoogendoorn@tudelft.nl; s.hoogendoorn-lanser@tudelft.nl}
}

\begin{document}

\maketitle

\begin{abstract}
In many operational time-series forecasting applications, such as crowd demand forecasting, the risk related to under-prediction is substantially higher than that of over-prediction. Accurate prediction of rare demand spikes plays a critical role in downstream tasks. Yet most time-series forecasters are trained with symmetric objectives (e.g., MSE, MAE) and evaluated primarily on aggregate error, which can mask failures in extreme-values and peak-timing predictions. We introduce \textbf{Asymmetric Peak-Aware Loss (APAL)}, a simple, model-agnostic objective that (i) penalizes under-predictions more heavily and (ii) increases the training weight of peak regions within each forecast window. We further propose a peak-critical evaluation protocol that complements MAE/MSE with channel-wise tail error (Top-10\% and Top-1\%) and peak metrics (precision, recall, F1 under timing tolerance, and peak timing error).
We evaluate APAL on long-horizon multivariate forecasting across five state-of-the-art backbones, with a focus on pedestrian demand forecasting using (i) a production-ready subset of the City of Melbourne pedestrian hourly count dataset and (ii) a beach visitor count dataset. The generality of the loss function for time-series forecasting is tested on additional benchmarks. Across peak-critical datasets and settings, APAL improves tail accuracy and peak-prediction quality while exposing a controllable trade-off with aggregate error, making it a practical solution when peak-prediction failures are the dominant operational concern.
\end{abstract}

\section{Introduction}
Accurate long-horizon time-series forecasting is central to operational and strategic decision-making across domains such as urban mobility, crowd management, energy planning, weather, financial risk and resource allocation~\cite{Zhou_Zhang_Peng_Zhang_Li_Xiong_Zhang_2021,schve_forecast, liu2024itransformer,9922512,Fan19}.
In many of these cases, the consequences of errors are asymmetric. Under-predicting demand might cause unsafe congestion, stampedes, staffing shortages, and decreased service quality, while moderate over-prediction is generally more acceptable~\cite{Berechman2018,9922512}.
Moreover, the operational risk arising from forecast-based decisions is often dominated by a small fraction of high-impact periods, such as demand spikes during commuting hours or large crowd gatherings (e.g., concerts, sporting events, or political demonstrations), where forecasting failures incur disproportionate costs.
These properties motivate the design of peak-critical forecasting objectives and evaluations that prioritise performance on extreme values and peaks.

Despite these requirements, most time-series forecasting models are predominantly trained with symmetric point loss functions such as Mean Squared Error (MSE) or Mean Absolute Error (MAE)~\cite{Zhou_Zhang_Peng_Zhang_Li_Xiong_Zhang_2021, liu2024itransformer, das2023tide,autoformer, wang2024tssurvey}.
Such objectives treat over- and under-prediction equally and allocate most training signal to the abundant non-extreme regions of the series, causing models to systematically smooth rare peaks~\cite{Forcastdia}.
As a result, models can achieve low average error while still systematically underestimating rare peaks or misplacing their timing, which is precisely where operational costs or risks are highest.
This mismatch arises broadly in peak-critical domains such as traffic flow, energy demand, and urban planning, and is particularly pronounced in pedestrian and crowd count forecasting, where distributions are heavy-tailed, include many zeros, and exhibit strong temporal structure~\cite{crowdcount}.

Recent work has proposed specialised objectives to shape forecasts beyond average pointwise accuracy \cite{dtw,softdtw,dilateloss,patchloss,tildeqloss,fredfloss,dbloss}.
However, evaluation is often reported primarily via MAE/MSE, which can obscure tail behavior and event-level peak quality \cite{Forcastdia}.
Peak-critical applications therefore require both (i) objectives aligned with asymmetric, extreme-value risk and (ii) evaluation that directly measures tail and peak-event behavior.

In this work, we introduce \textbf{Asymmetric Peak-Aware Loss (APAL)}, a simple, model-agnostic training objective for point forecasting.
APAL combines two principles: \textbf{(1) asymmetric cost}, which penalises under-predictions more than over-predictions, and \textbf{(2) peak emphasis}, which upweights errors in peak regions of the ground-truth forecast window to discourage peak smoothing.
The framework is flexible: the asymmetry direction can be reversed when over-prediction carries a higher cost, and the emphasis mechanism can target other regions of interest.
APAL requires only element-wise operations and is controlled by three interpretable hyperparameters.

\begin{figure}[ht]
\centering
\includegraphics[width=0.55\linewidth]{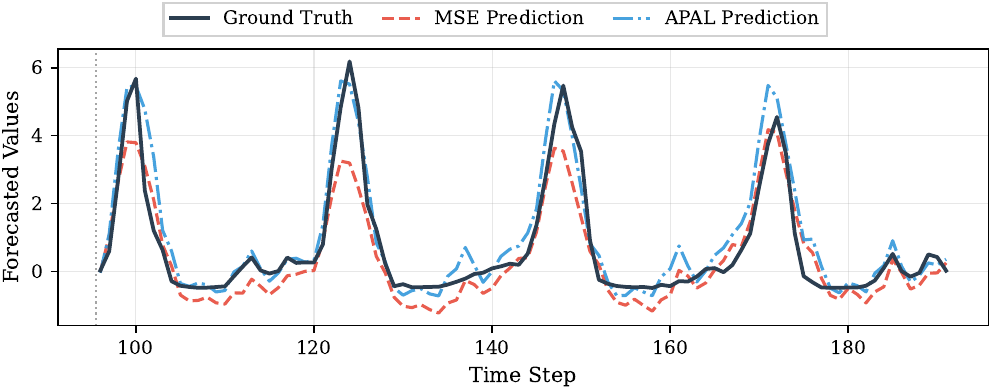}
\caption{
Qualitative comparison of 96-step forecasts on a single sensor of the Melbourne Pedestrian Count Dataset. 
Ground truth (solid black) versus MSE-trained (dashed orange) and APAL-trained (dash-dot blue) predictions. Values shown are standardized (z-score normalized).
}
\label{fig:intro_peaks}
\end{figure}

Beyond the loss, we propose a \textbf{peak-critical evaluation protocol} that reports standard metrics (MAE and MSE) alongside tail and event-based peak metrics.
We compute channel-wise tail error on the Top-10\% and Top-1\% of ground-truth values, and we treat peaks as events scored by precision, recall, and F1 under a timing tolerance, together with peak timing error.
Figure~\ref{fig:intro_peaks} illustrates that MSE-trained models can shrink peaks, while APAL improves peak magnitude and timing.

We study pedestrian demand forecasting motivated by staffing, public safety, transport planning, and crowd management~\cite{crowdcount}.
Our main datasets are (i) a production-ready subset of the City of Melbourne pedestrian hourly counts (16 sensor locations, 2010 to 2017) and (ii) a proprietary beach visitor dataset (16 areas, hourly, 2021 to 2023; not publicly released due to data-sharing restrictions).
To test generality, we additionally benchmark public datasets from other domains (energy, weather, finance, traffic), aiming to characterise where APAL offers clear benefits and where it does not.
We expect APAL to improve tail accuracy and peak-event quality on datasets with recurrent peak structure (e.g., pedestrian, traffic, electricity), where under-prediction of extremes is costly; conversely, on series without identifiable, learnable, or actionable peak structure (e.g., exchange rates), symmetric losses may remain preferable.

\paragraph{Contributions:}
\begin{enumerate}
\item \textbf{Loss:} We introduce APAL, a simple asymmetric and peak-aware objective that improves peak-critical forecasting while remaining model-agnostic and easy to deploy.
\item \textbf{Evaluation:} We propose a peak-critical metric suite, including channel-wise tail error (Top-10\%, Top-1\%), peak-event detection rate, and peak timing metrics, alongside standard MAE/MSE.
\item \textbf{Applicability diagnostic} We develop a principled pretraining diagnostic tool to quantify whether dataset peaks are structurally distinct and learnable and stable and operationally meaningful giving practitioners a clear rule for when to apply APAL.
\item \textbf{Peak-critical forecasting evidence:} Across five forecasting backbones and ten datasets, APAL improves tail accuracy and peak-event quality, exposing a clear and tunable tradeoff with aggregate error.
\end{enumerate}

\section{Related Work}
\label{sec:related}
In this section, we review our contributions within three areas: long-horizon forecasting architectures (Section~\ref{sec:rw-forecasting}), cost-sensitive and peak-aware objectives (Section~\ref{sec:rw-objectives}), and tail and event-based evaluation (Section~\ref{sec:rw-evaluation}).

\subsection{Long-horizon time-series forecasting}
\label{sec:rw-forecasting}
Long-horizon forecasting has been advanced by Transformer variants that handle long contexts and multivariate dependencies, as well as strong linear and MLP baselines.
Informer~\cite{Zhou_Zhang_Peng_Zhang_Li_Xiong_Zhang_2021} introduced efficient attention for long sequences, while Autoformer~\cite{autoformer} incorporated series decomposition and autocorrelation-style aggregation.
PatchTST~\cite{nie2023patchtst} improved efficiency and accuracy via patching and channel independence, and iTransformer~\cite{liu2024itransformer} proposed variate-centric attention to better model cross-channel structure.
In parallel, simpler architectures such as DLinear~\cite{zeng2023dlinear}, TiDE~\cite{das2023tide}, and TSMixer~\cite{TSMixer} have demonstrated that carefully designed linear and MLP-based models remain highly competitive on long-horizon benchmarks. To isolate the effect of the training objective from architectural choices, our experiments use a representative subset spanning linear, MLP-based, and Transformer-based families (Section~\ref{sec:models}).

\subsection{Cost-sensitive, asymmetric, and peak-aware objectives}
\label{sec:rw-objectives}
Most time-series forecasting models are trained with symmetric point losses (MSE, MAE), which optimise average errors but can be misaligned when the operational cost of errors is asymmetric. For example, when under-prediction is substantially more costly than over-prediction, or vice versa.
Classical cost-sensitive objectives include \emph{quantile} (pinball) and \emph{expectile} losses, widely used in risk-sensitive and probabilistic forecasting~\cite{SMYL2026103637, 10230996}, including DeepAR~\cite{deepar} and the Temporal Fusion Transformer~\cite{tft}.
These target conditional quantiles rather than peak-critical point forecasts, and apply a uniform asymmetric penalty across all time steps, so the gradient signal on rare extremes is still diluted by abundant non-extreme points.
In computer vision, Focal loss~\cite{lin2017focal} reweights by classification confidence rather than prediction direction or temporal structure.
Imbalanced-regression methods~\cite{yang2021delving, steininger2021density} reweight by label density but encode neither directional cost nor temporal peak structure.

A recent work targets \emph{structural} agreement between predicted and true trajectories.
Alignment-based objectives include differentiable DTW variants such as SoftDTW~\cite{softdtw} and DILATE~\cite{dilateloss}, which penalise both value mismatch and temporal misalignment.
Other methods aim to match local or transformation-robust structure.
\textbf{TILDE-Q}~\cite{tildeqloss} is a lightweight shape-aware loss designed to reduce sensitivity to certain distortions by jointly accounting for amplitude and phase distortions under a transformation-invariant design rationale.
\textbf{Patch-wise Structural (PS)} loss~\cite{patchloss} compares predicted and true series at the patch level using local statistics such as correlation, mean, and variance, and can be combined with a pointwise term to improve local structural alignment.
Finally, \textbf{DBLoss}~\cite{dbloss} is a decomposition-based loss that uses exponential moving averages to decompose targets within the forecasting horizon into trend and seasonal components and applies separate, weighted losses to these components.

These structure-oriented objectives can improve the \emph{overall} shape of forecasts, including the placement of local extrema, but they are not specifically designed for peak-critical applications.
First, they are typically \emph{symmetric} with respect to under-versus over-prediction, so they do not encode asymmetric operational risk.
Second, many structural terms aggregate errors over the full horizon or across patches, so the gradient contribution of rare extreme points (peaks) can still be diluted when extremes occupy a small fraction of time steps.
Third, several structural comparisons rely on normalised summaries (e.g., correlation-like statistics), which can reward correct \emph{relative} shape while tolerating unacceptable \emph{absolute} magnitude shrinkage at sharp extremes.
In peak-critical settings, a slightly smoothed peak (or dip) may remain structurally plausible yet be operationally costly if it underestimates the extreme magnitude or misses the event under a tight timing tolerance. These limitations motivate an objective that explicitly targets (i) asymmetric cost and (ii) increased emphasis on extreme regions within each forecast window, which we achieve with APAL.

\subsection{Tail and event-based evaluation for peak-critical forecasting}
\label{sec:rw-evaluation}
Standard forecasting benchmarks predominantly report aggregate metrics (MAE, MSE), which average errors across all time steps and can mask poor performance on rare but critical extreme peaks~\cite{Forcastdia}.
This evaluation gap is problematic for peak-critical applications where the majority of operational risk concentrates in a small fraction of high-value periods.

Several lines of work address this limitation, though each covers only part of the requirements for peak-critical evaluation.
\emph{Tail-focused metrics} restrict error computation to extreme quantiles of the target distribution, a practice common in financial risk forecasting and extreme weather prediction but underutilized in general time-series benchmarks~\cite{liu2024itransformer, Zhou_Zhang_Peng_Zhang_Li_Xiong_Zhang_2021, autoformer,das2023tide}; however, they measure only magnitude accuracy and do not assess whether peaks are detected as events or correctly timed.
\emph{Alignment-aware metrics} such as DTW~\cite{dtw} and differentiable variants~\cite{softdtw} penalize temporal misalignment but do not distinguish peak from non-peak regions, so timing errors on critical peaks are averaged with errors elsewhere.
\emph{Event-based scoring} treats peaks as discrete events and evaluates detection quality using precision, recall, and F1 under timing tolerance~\cite{dtw_f1}, analogous to object detection evaluation in computer vision; yet it does not directly measure the magnitude accuracy of matched peaks.

We synthesize these complementary perspectives into a unified \emph{peak-critical evaluation protocol}.
Drawing on tail-focused evaluation from financial risk forecasting~\cite{Forcastdia}, event-based scoring from anomaly and peak detection~\cite{dtw_f1}, and timing-aware metrics from temporal alignment literature~\cite{dtw, softdtw}, our protocol reports: (i) channel-wise tail errors on the Top-10\% and Top-1\% of ground-truth values, capturing magnitude accuracy at extremes; (ii) peak-event detection metrics (precision, recall, F1) with tolerance matching, measuring whether peaks are correctly identified; and (iii) peak timing error, quantifying temporal displacement of detected peaks.
We additionally report the Pearson Correlation Coefficient (PCC) and the Temporal Distortion Index (TDI)~\cite{FRIASPAREDES2016180}.
The choice of tail thresholds (10\%, 1\%) implicitly assumes peaks occur at least this frequently; when true peaks are rarer, the tail set may include non-peak high values. We discuss threshold selection, sensitivity to peak frequency, and robustness to data errors in Appendix~\ref{app:tail_assumptions}.

\section{Problem Formulation and Method}
\label{sec:prelim}
This section formalises the long-horizon multivariate forecasting problem (Section~\ref{sec:setup}), defines tail points and peak events (Section~\ref{sec:peaks}), and presents APAL (Section~\ref{sec:apal}).

\subsection{Problem Setup and Notation}
\label{sec:setup}
We consider long-horizon multivariate time-series forecasting.
Let $\mathbf{x}_t \in \mathbb{R}^{C}$ denote the observation at time $t$, where $C$ is the number of channels (variates).
Given an input context window of length $L$, the input is
$\mathbf{X}_{t-L+1:t} \in \mathbb{R}^{L \times C}$ and the goal is to predict the next $H$ steps
$\mathbf{Y}_{t+1:t+H} \in \mathbb{R}^{H \times C}$.
A forecasting model $f_{\theta}$ produces a point forecast
$\widehat{\mathbf{Y}} = f_{\theta}(\mathbf{X}) \in \mathbb{R}^{H \times C}$.

Throughout, indices $h \in \{1,\dots,H\}$ and $c \in \{1,\dots,C\}$ denote forecast time steps and channels, respectively.
For a sample, let $\hat{y}_{h,c}$ and $y_{h,c}$ denote prediction and ground truth.
We define the signed error $e_{h,c} = \hat{y}_{h,c} - y_{h,c}$ so that $e_{h,c} < 0$ indicates under-prediction. Throughout, we use $\mathbb{I}[\cdot]$ to denote the indicator function, which equals one when its argument is true and zero otherwise.

\subsection{Peak-Critical Forecasting: Tails and Peak Events}
\label{sec:peaks}
Peak-critical applications place particular emphasis on high-demand periods and their timing.
We therefore distinguish two notions that are complementary in evaluation.

\paragraph{Tail points (extreme values).}
Tail metrics restrict error computation to the largest target values.
For a given channel $c$, let $\mathcal{I}^{(q)}_c$ denote the indices of the top-$q$ fraction of ground-truth values among all evaluated points in that channel, where $q \in \{0.10,0.01\}$ 
corresponds to Top-10\% and Top-1\%.
Tail errors ($\text{MSE}_{10}$, $\text{MSE}_1$) measure magnitude accuracy at these extremes.

\paragraph{Peak events (local maxima).}
Tail points capture large values but do not directly encode event structure or timing.
Event-based peak metrics treat peaks as local maxima in each forecast window and score whether predicted peaks occur near true peaks.
We detect peaks using a standard local-maximum operator with an amplitude threshold (defined in Appendix~\ref{app:peak_metrics}), and we allow a tolerance window of $\pm \Delta$ steps (with $\Delta=3$ by default) when matching predicted and true peaks. The peak timing is computed via the argmax function for each tolerance window for both predicted and ground truth peaks. The difference between the two is then computed as peak timing error (PTE).
This yields event precision, recall, F1 (Peak F1) and PTE.

Together, tail metrics assess \emph{how accurately} extreme values are predicted, while event metrics assess \emph{whether} peaks are detected and \emph{when} they occur. We additionally report PCC and TDI~\cite{FRIASPAREDES2016180}, as introduced in Section~\ref{sec:rw-evaluation}.

\subsection{Asymmetric Peak-Aware Loss}
\label{sec:apal}

Building on the motivation in Section~\ref{sec:rw-objectives}, we present APAL as a weighted point-forecast objective with two multiplicative terms: an asymmetric term penalizing under-predictions and a peak-emphasis term upweighting high-magnitude regions.

\paragraph{Asymmetric under-prediction penalty.}
We upweight under-predictions multiplicatively:
\begin{equation}
w^{\text{asym}}_{h,c} = 1 + (\lambda_u - 1)\,\mathbb{I}[e_{h,c} < 0],
\end{equation}
where $\lambda_u \ge 1$ controls the strength of the penalty.

\paragraph{Peak-region emphasis.}
For each training sample and channel, we define the within-horizon ground-truth maximum $y^{\max}_{c} = \max_{h} y_{h,c}$ and mark peak regions as values exceeding a fraction $\tau \in (0,1)$ of this maximum:
\begin{equation}
w^{\text{peak}}_{h,c} = 1 + (\lambda_p - 1)\,\mathbb{I}[y_{h,c} \ge \tau\,y^{\max}_{c}],
\end{equation}
where $\lambda_p \ge 1$ controls peak emphasis. Per-window normalisation makes the mask adaptive across heterogeneous channels and ensures at least one point is emphasised even in flat windows. APAL therefore targets settings where high values correspond to meaningful, recurrent, learnable structure rather than isolated sensor artefacts.

\paragraph{Final objective.}
APAL is the weighted absolute error averaged over the horizon and channels:
\begin{equation}
\mathcal{L}_{\text{APAL}}(\widehat{\mathbf{Y}},\mathbf{Y})
= \frac{1}{HC}\sum_{h=1}^{H}\sum_{c=1}^{C}
\left|e_{h,c}\right|\, w^{\text{asym}}_{h,c}\, w^{\text{peak}}_{h,c}
\end{equation}

\paragraph{Special cases and interpretation.}
When $\lambda_u = \lambda_p = 1$, APAL reduces to MAE.
Increasing $\lambda_u$ encodes asymmetric operational risk by giving under-predictions weight $\lambda_u$ while over-predictions retain weight 1, biasing the model upward to avoid the heavier penalty.
Increasing $\lambda_p$ concentrates the learning signal on peak regions and reduces peak smoothing. Since $w^{\text{peak}}$ depends only on ground truth, APAL is a drop-in objective. The indicator-based weights are piecewise constant and yield well-defined subgradients almost everywhere, sufficient for standard first-order optimisation. We use weighted absolute error to retain stable gradients on large-magnitude targets and to avoid disproportionate sensitivity to outliers under squared error. Combining APAL with squared error can over-amplify rare points and destabilise optimisation, so we leave systematic comparison of base error functions to future work.

\paragraph{Theoretical motivation.}
APAL's multiplicative weighting reflects two independent sources of sample importance: (i) \emph{direction-dependent cost} (under- vs.\ over-prediction) and (ii) \emph{magnitude-dependent cost} (peak vs.\ non-peak), which factorise as $w^{\text{asym}} \cdot w^{\text{peak}}$ under separability~\cite{zellner1986linex}. We view APAL as a cost-sensitive empirical-risk objective, not a likelihood-optimal estimator under a universal noise model. From a gradient perspective, for a single sample with peak indicator $p = \mathbb{I}[y \geq \tau y^{\max}]$ and under-prediction indicator $u = \mathbb{I}[\hat{y} < y]$:
\begin{equation}
\frac{\partial \mathcal{L}_{\text{APAL}}}{\partial \hat{y}} = \mathrm{sign}(\hat{y} - y) \cdot \bigl(1 + (\lambda_u - 1)u\bigr)\bigl(1 + (\lambda_p - 1)p\bigr)
\end{equation}
Peak under-predictions thus receive gradients scaled by $\lambda_u \lambda_p$, focusing optimisation on the regions of highest operational concern. Unlike focal loss~\cite{lin2017focal}, which reweights by prediction \emph{confidence}, APAL reweights by ground-truth \emph{structure} and prediction \emph{direction}, suiting it for asymmetric peak-critical settings.

\section{Experimental setup}
\label{sec:experiments}

\subsection{Dataset}
\label{sec:data}
We focus on two pedestrian datasets: an open, production-ready subset of the City of Melbourne pedestrian count dataset (16 sensor locations, hourly sampling rate, from 2010 to 2017) and a Beach visitor count dataset (16 different areas, hourly sampling rate, from 2021 to 2023).
We further benchmark on standard public multivariate forecasting datasets commonly used in long-horizon evaluation, namely electricity transformer monitoring (ETT), power consumption (ECL), currency exchange (Exchange), road traffic (Traffic), and meteorology (Weather).
Detailed dataset descriptions, preprocessing, and split construction are provided in Appendix~\ref{app:datasets}.

\subsection{Backbones}
\label{sec:models}
We evaluate APAL across five backbones spanning linear (\textbf{DLinear}~\cite{zeng2023dlinear}), MLP-based (\textbf{TSMixer}~\cite{TSMixer}, \textbf{TiDE}~\cite{das2023tide}), and Transformer-based (\textbf{PatchTST}~\cite{nie2023patchtst}, \textbf{iTransformer}~\cite{liu2024itransformer}) families to test whether APAL transfers across architectures.
Model descriptions and hyperparameters are reported in Appendix~\ref{app:models}.

\subsection{Implementation Details}
\label{sec:impl}

We use input length $L=96$ and horizons $H \in \{96,192,336,720\}$ under chronological splits with training-only normalisation and inverse-transform evaluation.
All backbones use direct multi-step forecasting (the model outputs the entire horizon in one forward pass), so our experiments test APAL's fixed-horizon trade-off but do not evaluate recursive deployments where directional bias could accumulate.
Unless stated otherwise, results use the checkpoint with best validation performance under the selection criterion, with the same optimiser, early stopping, and training budget across all losses to isolate the effect of the objective. 
APAL hyperparameters are tuned on validation only (Appendix~\ref{app:tuning}).
Experiments use PyTorch 2.4 on an NVIDIA L4 GPU (24\,GB) using the Time Series Library (TSLib)~\cite{wang2024tssurvey}.
Computational overhead is reported in Section~\ref{sec:compute}.

\subsection{Evaluation Metrics}
\label{sec:metrics}
We report standard (MSE, MAE) and peak-critical metrics (Tail $\text{MSE}_{10}$ and Tail $\text{MSE}_{1}$) across all horizons. Peak-critical metrics are computed per sensor location (i.e., per channel/variate) and then averaged across all locations to prevent high-variance sensors from dominating the aggregate. We additionally report channel-wise breakdowns to reflect heterogeneity across locations in section~\ref{sec:channelwise-results}.

\section{Experimental Results and Discussion}
\label{sec:results}


\subsection{Main Peak-Critical Analysis on Pedestrian Forecasting}
\label{sec:main_result}

We compare APAL against symmetric pointwise losses (MSE, MAE) and specialised objectives (TILDE-Q, DBLoss, PS) across all backbones and horizons (Table~\ref{tab:combined_results}). Symmetric objectives spread training signal uniformly so rare extremes are underweighted; APAL counteracts this by concentrating gradients on peak regions ($\lambda_p$) and penalising under-prediction ($\lambda_u$). Specialised baselines improve structural alignment but remain symmetric.

The two datasets exhibit distinct characteristics: Pedestrian has regular weekly peaks predictable from history, while Beach has irregular peaks tied to weather and events. On Pedestrian, APAL achieves Peak~F1 $>$0.79 across all settings, while baseline Peak~F1 spans a very wide range across backbones (from below 0.05 at long horizons on iTransformer up to about 0.78 on PatchTST/TiDE). Excluding the iTransformer outlier, baseline Peak~F1 lies roughly in 0.42--0.78. On Beach, $\text{MSE}_1$ improvements are largest at shorter horizons and on the stronger backbones, reaching up to about 31\% on TSMixer (and $\sim$29\% on PatchTST) at $H=96$, but attenuate at $H=720$ where APAL is comparable to or slightly worse than the MSE baseline on some backbones. These tail gains come at an intentional, controllable cost: aggregate MSE rises relative to symmetric losses (Section~\ref{sec:ablation}).

Across both pedestrian-domain datasets, APAL achieves the best $\text{MSE}_1$ in 39 of 40 dataset--model--horizon combinations, consistently, suggesting a systematic rather than random improvement on peak-critical metrics. The advantage is most pronounced at $H \in \{96, 192\}$ and attenuates at $H=720$ due to forecast uncertainty, though APAL still attains the best tail metrics. Among backbones, iTransformer is more variable, likely because its variate-centric attention partially overlaps with APAL's peak emphasis; for architectures without explicit cross-channel modelling, gains are more uniform. Architectural and qualitative analyses are in Appendix~\ref{app:additional}.

\begin{table*}[ht]
\centering
\caption{Comparison of loss functions for multivariate long-term forecasting. Prediction horizons $H \in \{96, 192, 336, 720\}$. The proposed APAL is highlighted in bold. Best prediction results are \textbf{bolded}; second-best are \underline{underlined}.}
\label{tab:combined_results}
\resizebox{\textwidth}{!}{%
\begin{tabular}{c|c|l|ccccc|ccccc|ccccc|ccccc}
\toprule
Dataset & Model & Loss & \multicolumn{5}{c|}{96} & \multicolumn{5}{c|}{192} & \multicolumn{5}{c|}{336} & \multicolumn{5}{c}{720} \\
\cmidrule(lr){4-8} \cmidrule(lr){9-13} \cmidrule(lr){14-18} \cmidrule(lr){19-23}
 &  &  & MSE & MSE$_{10}$ & MSE$_{1}$ & Peak F1 & PTE & MSE & MSE$_{10}$ & MSE$_{1}$ & Peak F1 & PTE & MSE & MSE$_{10}$ & MSE$_{1}$ & Peak F1 & PTE & MSE & MSE$_{10}$ & MSE$_{1}$ & Peak F1 & PTE \\
\midrule
\multirow{30}{*}{Pedestrian} & \multirow{6}{*}{DLinear} & MSE & \underline{0.433} & 1.190 & 4.230 & 0.442 & \underline{0.987} & \underline{0.395} & 1.133 & 4.230 & 0.517 & 0.929 & \textbf{0.402} & 1.161 & 4.328 & 0.513 & 0.919 & \underline{0.426} & 1.220 & 4.480 & 0.469 & 0.956 \\
 &  & MAE & 0.440 & 1.171 & \underline{4.150} & \underline{0.573} & 1.028 & 0.402 & 1.121 & \underline{4.172} & 0.626 & 0.967 & 0.409 & 1.148 & \underline{4.262} & 0.612 & 0.954 & 0.432 & 1.210 & 4.430 & 0.560 & 1.000 \\
 &  & TildeQ & 0.451 & 1.229 & 4.273 & 0.425 & 1.141 & 0.408 & 1.148 & 4.234 & 0.529 & 1.035 & 0.407 & 1.160 & 4.309 & 0.543 & \underline{0.908} & 0.434 & 1.234 & 4.492 & 0.465 & \underline{0.930} \\
 &  & DBLoss & \textbf{0.431} & \textbf{1.157} & 4.156 & 0.568 & 0.991 & \textbf{0.395} & \textbf{1.108} & \underline{4.172} & \underline{0.637} & \underline{0.920} & \underline{0.403} & \textbf{1.136} & \underline{4.262} & \underline{0.624} & 0.908 & \textbf{0.425} & \underline{1.196} & \underline{4.422} & \underline{0.573} & 0.944 \\
 &  & PS & 0.434 & 1.167 & 4.191 & 0.555 & 1.037 & 0.398 & \underline{1.119} & 4.207 & 0.622 & 0.991 & 0.406 & 1.146 & 4.297 & 0.608 & 0.981 & 0.427 & 1.204 & 4.445 & 0.553 & 1.013 \\
 &  & \textbf{APAL} & 0.537 & \underline{1.166} & \textbf{3.929} & \textbf{0.832} & \textbf{0.721} & 0.476 & 1.122 & \textbf{4.013} & \textbf{0.823} & \textbf{0.731} & 0.481 & \underline{1.143} & \textbf{4.064} & \textbf{0.808} & \textbf{0.740} & 0.499 & \textbf{1.181} & \textbf{4.206} & \textbf{0.803} & \textbf{0.786} \\
\cline{2-23}
 & \multirow{6}{*}{PatchTST} & MSE & \textbf{0.253} & \underline{0.868} & \underline{3.952} & 0.741 & 0.896 & \textbf{0.257} & \textbf{0.883} & \underline{4.006} & 0.756 & 0.934 & \textbf{0.292} & 0.967 & \underline{4.198} & 0.716 & 0.882 & \textbf{0.312} & 1.011 & \underline{4.305} & 0.702 & 0.920 \\
 &  & MAE & 0.258 & 0.909 & 4.120 & 0.721 & \textbf{0.807} & 0.265 & 0.936 & 4.199 & 0.705 & \underline{0.831} & 0.301 & 0.996 & 4.293 & 0.699 & 0.848 & 0.322 & 1.047 & 4.418 & 0.664 & \underline{0.856} \\
 &  & TildeQ & 0.263 & 0.892 & 4.027 & 0.774 & 1.002 & 0.268 & 0.907 & 4.070 & \underline{0.780} & 0.990 & \underline{0.297} & \underline{0.962} & 4.207 & \underline{0.768} & \underline{0.826} & 0.321 & 1.014 & 4.328 & \underline{0.741} & 0.859 \\
 &  & DBLoss & \underline{0.256} & 0.903 & 4.090 & 0.715 & 0.823 & \underline{0.260} & 0.923 & 4.148 & 0.695 & 0.850 & 0.298 & 0.999 & 4.305 & 0.658 & 0.834 & 0.320 & 1.050 & 4.418 & 0.621 & 0.873 \\
 &  & PS & 0.260 & 0.883 & 4.059 & \underline{0.781} & \underline{0.813} & 0.263 & 0.899 & 4.113 & 0.776 & 0.844 & 0.299 & 0.966 & 4.246 & 0.749 & 0.865 & \underline{0.318} & \underline{1.009} & 4.352 & 0.732 & 0.887 \\
 &  & \textbf{APAL} & 0.299 & \textbf{0.855} & \textbf{3.811} & \textbf{0.841} & 0.817 & 0.307 & \underline{0.888} & \textbf{3.919} & \textbf{0.835} & \textbf{0.794} & 0.348 & \textbf{0.962} & \textbf{4.044} & \textbf{0.813} & \textbf{0.712} & 0.363 & \textbf{1.001} & \textbf{4.197} & \textbf{0.808} & \textbf{0.763} \\
\cline{2-23}
 & \multirow{6}{*}{TSMixer} & MSE & 0.453 & \underline{1.226} & \underline{4.219} & 0.579 & 1.372 & 0.415 & \underline{1.165} & \underline{4.190} & 0.635 & 1.354 & 0.423 & \underline{1.198} & \underline{4.282} & 0.610 & 1.337 & 0.452 & \underline{1.270} & \underline{4.428} & 0.543 & 1.384 \\
 &  & MAE & \textbf{0.448} & 1.247 & 4.365 & 0.561 & \underline{1.321} & \underline{0.409} & 1.186 & 4.325 & 0.610 & 1.296 & \underline{0.417} & 1.213 & 4.402 & 0.600 & 1.283 & \textbf{0.445} & 1.288 & 4.537 & 0.528 & 1.335 \\
 &  & TildeQ & 0.481 & 1.345 & 4.373 & 0.506 & 1.420 & 0.425 & 1.218 & 4.271 & 0.586 & 1.393 & 0.425 & 1.235 & 4.332 & 0.571 & 1.368 & 0.460 & 1.329 & 4.506 & 0.475 & 1.392 \\
 &  & DBLoss & \underline{0.453} & 1.264 & 4.346 & 0.550 & 1.345 & \textbf{0.409} & 1.184 & 4.295 & 0.615 & \underline{1.293} & \textbf{0.417} & 1.214 & 4.382 & 0.592 & \underline{1.276} & \underline{0.446} & 1.291 & 4.531 & 0.516 & \underline{1.324} \\
 &  & PS & 0.470 & 1.265 & 4.339 & \underline{0.638} & 1.358 & 0.422 & 1.194 & 4.300 & \underline{0.668} & 1.301 & 0.430 & 1.225 & 4.381 & \underline{0.649} & 1.289 & 0.461 & 1.299 & 4.507 & \underline{0.589} & 1.339 \\
 &  & \textbf{APAL} & 0.545 & \textbf{1.147} & \textbf{3.970} & \textbf{0.832} & \textbf{1.199} & 0.481 & \textbf{1.107} & \textbf{4.032} & \textbf{0.823} & \textbf{1.133} & 0.481 & \textbf{1.139} & \textbf{4.161} & \textbf{0.809} & \textbf{1.136} & 0.514 & \textbf{1.195} & \textbf{4.311} & \textbf{0.802} & \textbf{1.256} \\
\cline{2-23}
 & \multirow{6}{*}{TiDE} & MSE & \textbf{0.429} & 1.143 & 4.148 & 0.582 & 0.917 & \textbf{0.390} & \textbf{1.094} & 4.164 & 0.637 & 0.847 & \textbf{0.399} & 1.126 & 4.270 & 0.625 & 0.838 & \textbf{0.424} & 1.188 & 4.430 & 0.590 & 0.883 \\
 &  & MAE & 0.463 & 1.165 & \underline{4.078} & \underline{0.707} & \underline{0.876} & 0.425 & 1.128 & \underline{4.125} & \underline{0.735} & \underline{0.818} & 0.432 & 1.151 & \underline{4.203} & \underline{0.721} & \underline{0.815} & 0.455 & 1.205 & \underline{4.371} & \underline{0.688} & \underline{0.859} \\
 &  & TildeQ & 0.451 & 1.171 & 4.152 & 0.637 & 1.058 & 0.407 & 1.106 & 4.156 & 0.690 & 0.951 & 0.408 & \textbf{1.122} & 4.230 & 0.688 & 0.819 & 0.435 & 1.191 & 4.410 & 0.630 & 0.888 \\
 &  & DBLoss & \underline{0.436} & \underline{1.141} & 4.102 & 0.642 & 0.960 & \underline{0.399} & \underline{1.097} & 4.133 & 0.696 & 0.880 & \underline{0.408} & \underline{1.125} & 4.219 & 0.681 & 0.870 & \underline{0.431} & \textbf{1.183} & 4.383 & 0.640 & 0.907 \\
 &  & PS & 0.440 & \textbf{1.140} & 4.105 & 0.679 & 0.971 & 0.405 & 1.104 & 4.141 & 0.716 & 0.909 & 0.414 & 1.131 & 4.223 & 0.703 & 0.898 & 0.436 & \underline{1.185} & 4.379 & 0.669 & 0.934 \\
 &  & \textbf{APAL} & 0.570 & 1.239 & \textbf{4.009} & \textbf{0.825} & \textbf{0.690} & 0.501 & 1.186 & \textbf{4.085} & \textbf{0.822} & \textbf{0.686} & 0.513 & 1.213 & \textbf{4.134} & \textbf{0.807} & \textbf{0.692} & 0.542 & 1.264 & \textbf{4.292} & \textbf{0.800} & \textbf{0.736} \\
\cline{2-23}
 & \multirow{6}{*}{iTransformer} & MSE & 0.454 & 1.282 & 4.465 & 0.175 & 0.788 & 0.533 & 1.532 & 4.742 & 0.026 & 1.035 & 0.665 & 1.884 & 5.176 & 0.004 & 1.062 & 0.740 & 2.049 & 5.426 & 0.007 & 1.074 \\
 &  & MAE & 0.373 & 1.087 & 4.223 & 0.437 & 0.922 & 0.513 & 1.489 & 4.699 & 0.038 & 1.018 & 0.573 & 1.635 & 4.922 & 0.031 & 1.091 & 0.652 & 1.856 & 5.238 & 0.011 & 1.177 \\
 &  & TildeQ & \textbf{0.354} & \underline{1.032} & \underline{4.168} & \underline{0.661} & 0.873 & \textbf{0.345} & \underline{1.035} & \underline{4.203} & \underline{0.703} & \underline{0.669} & \textbf{0.339} & \underline{1.033} & \underline{4.297} & \underline{0.739} & \textbf{0.579} & \textbf{0.369} & 1.114 & 4.469 & \underline{0.699} & \textbf{0.673} \\
 &  & DBLoss & 0.391 & 1.137 & 4.305 & 0.390 & 0.812 & 0.495 & 1.433 & 4.637 & 0.062 & 0.970 & 0.480 & 1.383 & 4.664 & 0.105 & 0.935 & \underline{0.375} & \underline{1.092} & \underline{4.426} & 0.676 & 0.832 \\
 &  & PS & 0.390 & 1.105 & 4.250 & 0.567 & \textbf{0.721} & 0.361 & 1.061 & 4.250 & 0.676 & \textbf{0.593} & \underline{0.364} & 1.075 & 4.332 & 0.697 & \underline{0.617} & 0.403 & 1.153 & 4.473 & 0.616 & \underline{0.741} \\
 &  & \textbf{APAL} & \underline{0.370} & \textbf{0.977} & \textbf{3.905} & \textbf{0.811} & \underline{0.778} & \underline{0.351} & \textbf{0.968} & \textbf{3.966} & \textbf{0.825} & 0.728 & 0.378 & \textbf{1.015} & \textbf{4.042} & \textbf{0.815} & 0.723 & 0.389 & \textbf{1.050} & \textbf{4.231} & \textbf{0.792} & 0.854 \\
\midrule
\multirow{29}{*}{Beach} & \multirow{6}{*}{DLinear} & MSE & \textbf{0.753} & \underline{4.081} & \underline{11.432} & \underline{0.456} & \underline{1.677} & \textbf{0.753} & \underline{4.034} & \underline{10.862} & \underline{0.418} & \textbf{1.691} & \textbf{0.770} & \underline{4.088} & \underline{10.480} & \textbf{0.392} & \textbf{1.700} & \textbf{0.798} & \textbf{4.204} & \underline{11.073} & \textbf{0.395} & \underline{1.674} \\
 &  & MAE & 0.793 & 4.601 & 12.663 & 0.358 & 1.712 & 0.793 & 4.530 & 12.024 & 0.298 & 1.718 & 0.814 & 4.625 & 11.665 & 0.280 & 1.728 & 0.846 & 4.760 & 12.323 & 0.297 & 1.713 \\
 &  & TildeQ & 0.772 & 4.356 & 12.188 & 0.383 & 1.680 & 0.770 & 4.262 & 11.453 & 0.334 & 1.703 & 0.789 & 4.340 & 11.086 & 0.309 & 1.710 & 0.821 & 4.527 & 11.867 & 0.299 & \textbf{1.672} \\
 &  & DBLoss & 0.777 & 4.391 & 12.214 & 0.388 & 1.688 & 0.777 & 4.350 & 11.590 & 0.314 & 1.699 & 0.796 & 4.430 & 11.207 & 0.300 & 1.706 & 0.828 & 4.571 & 11.880 & 0.308 & 1.683 \\
 &  & PS & \underline{0.770} & 4.283 & 12.091 & 0.415 & \textbf{1.675} & \underline{0.770} & 4.223 & 11.430 & 0.354 & \underline{1.696} & \underline{0.788} & 4.295 & 11.078 & 0.327 & \underline{1.702} & \underline{0.817} & 4.416 & 11.710 & \underline{0.346} & 1.680 \\
 &  & \textbf{APAL} & 0.888 & \textbf{3.590} & \textbf{9.259} & \textbf{0.487} & 1.688 & 0.826 & \textbf{3.685} & \textbf{9.111} & \textbf{0.418} & 1.709 & 0.812 & \textbf{3.897} & \textbf{9.348} & \underline{0.328} & 1.717 & 0.830 & \underline{4.321} & \textbf{11.017} & 0.285 & 1.706 \\
\cline{2-23}
 & \multirow{5}{*}{PatchTST} & MSE & \textbf{0.743} & \underline{3.955} & \underline{10.711} & \underline{0.478} & 1.696 & \textbf{0.753} & \underline{4.049} & \underline{10.506} & \underline{0.405} & 1.703 & \textbf{0.789} & \underline{4.195} & \underline{10.461} & \underline{0.368} & \underline{1.712} & \textbf{0.831} & \underline{4.426} & \underline{11.383} & \underline{0.349} & \underline{1.699} \\
 &  & TildeQ & 0.771 & 4.210 & 11.548 & 0.274 & 1.700 & 0.782 & 4.205 & 10.988 & 0.271 & 1.724 & 0.828 & 4.376 & 11.012 & 0.249 & 1.736 & 0.890 & 4.632 & 12.021 & 0.224 & 1.727 \\
 &  & DBLoss & 0.774 & 4.443 & 12.184 & 0.366 & 1.693 & 0.784 & 4.503 & 11.844 & 0.257 & 1.712 & 0.821 & 4.688 & 11.911 & 0.246 & 1.721 & 0.861 & 4.886 & 12.727 & 0.229 & 1.708 \\
 &  & PS & \underline{0.761} & 4.189 & 11.627 & 0.442 & \textbf{1.682} & \underline{0.768} & 4.186 & 11.134 & 0.380 & \textbf{1.701} & \underline{0.805} & 4.331 & 11.066 & 0.346 & \textbf{1.708} & \underline{0.845} & 4.513 & 11.882 & 0.346 & \textbf{1.692} \\
 &  & \textbf{APAL} & 0.994 & \textbf{3.670} & \textbf{7.599} & \textbf{0.537} & \underline{1.689} & 0.920 & \textbf{3.792} & \textbf{7.410} & \textbf{0.500} & \underline{1.702} & 0.897 & \textbf{3.885} & \textbf{7.881} & \textbf{0.463} & 1.719 & 0.858 & \textbf{4.236} & \textbf{10.043} & \textbf{0.363} & 1.704 \\
\cline{2-23}
 & \multirow{6}{*}{TSMixer} & MSE & \textbf{0.751} & \underline{3.885} & \underline{10.748} & \underline{0.441} & \textbf{1.667} & 0.770 & \underline{3.943} & \underline{10.693} & \underline{0.408} & \textbf{1.687} & \underline{0.775} & \underline{3.964} & \underline{10.354} & \underline{0.393} & \textbf{1.702} & \textbf{0.801} & \textbf{4.056} & \textbf{10.891} & \textbf{0.368} & \textbf{1.680} \\
 &  & MAE & 0.804 & 4.787 & 12.965 & 0.247 & \underline{1.680} & 0.812 & 4.812 & 12.729 & 0.140 & 1.700 & 0.830 & 4.898 & 12.496 & 0.132 & \underline{1.711} & 0.864 & 5.071 & 13.250 & 0.142 & 1.688 \\
 &  & TildeQ & \underline{0.757} & 4.320 & 11.892 & 0.322 & 1.681 & \textbf{0.757} & 4.237 & 11.373 & 0.289 & 1.710 & \textbf{0.773} & 4.303 & 11.074 & 0.260 & 1.717 & \underline{0.808} & 4.516 & 11.950 & 0.248 & \underline{1.687} \\
 &  & DBLoss & 0.779 & 4.495 & 12.348 & 0.306 & 1.681 & 0.785 & 4.511 & 12.095 & 0.224 & \underline{1.699} & 0.803 & 4.591 & 11.869 & 0.193 & 1.712 & 0.835 & 4.752 & 12.595 & 0.214 & 1.687 \\
 &  & PS & 0.760 & 4.132 & 11.548 & 0.396 & 1.690 & \underline{0.764} & 4.144 & 11.315 & 0.354 & 1.713 & 0.780 & 4.209 & 11.120 & 0.326 & 1.723 & 0.808 & 4.332 & 11.748 & 0.308 & 1.700 \\
 &  & \textbf{APAL} & 1.025 & \textbf{3.352} & \textbf{7.400} & \textbf{0.567} & 1.683 & 0.931 & \textbf{3.448} & \textbf{7.676} & \textbf{0.522} & 1.717 & 0.835 & \textbf{3.620} & \textbf{8.427} & \textbf{0.482} & 1.728 & 0.818 & \underline{4.304} & \underline{11.183} & \underline{0.356} & 1.695 \\
\cline{2-23}
 & \multirow{6}{*}{TiDE} & MSE & \textbf{0.776} & \underline{4.160} & \underline{11.469} & \underline{0.452} & 1.690 & \textbf{0.778} & \underline{4.125} & \underline{10.839} & \underline{0.399} & 1.686 & \textbf{0.799} & \underline{4.184} & \underline{10.376} & \underline{0.390} & \underline{1.691} & \textbf{0.838} & \underline{4.341} & \underline{11.162} & \textbf{0.392} & 1.673 \\
 &  & MAE & 0.808 & 4.690 & 12.777 & 0.345 & 1.705 & 0.808 & 4.608 & 12.057 & 0.279 & 1.695 & 0.829 & 4.681 & 11.613 & 0.289 & 1.705 & 0.866 & 4.845 & 12.399 & 0.281 & 1.681 \\
 &  & TildeQ & 0.789 & 4.385 & 12.157 & 0.379 & \textbf{1.680} & 0.792 & 4.300 & 11.418 & 0.319 & 1.697 & 0.815 & 4.359 & 10.946 & 0.319 & 1.703 & 0.859 & 4.558 & 11.806 & 0.301 & 1.679 \\
 &  & DBLoss & 0.788 & 4.460 & 12.239 & 0.382 & 1.694 & 0.796 & 4.372 & 11.471 & 0.364 & \textbf{1.675} & 0.810 & 4.454 & 11.062 & 0.326 & 1.695 & 0.846 & 4.611 & 11.829 & 0.321 & 1.678 \\
 &  & PS & \underline{0.784} & 4.396 & 12.137 & 0.397 & \underline{1.684} & \underline{0.785} & 4.333 & 11.414 & 0.346 & 1.683 & \underline{0.807} & 4.409 & 10.966 & 0.336 & 1.692 & \underline{0.843} & 4.562 & 11.739 & 0.330 & \underline{1.673} \\
 &  & \textbf{APAL} & 1.000 & \textbf{3.661} & \textbf{8.610} & \textbf{0.519} & 1.702 & 0.926 & \textbf{3.743} & \textbf{8.378} & \textbf{0.474} & \underline{1.681} & 0.883 & \textbf{3.897} & \textbf{8.619} & \textbf{0.429} & \textbf{1.683} & 0.867 & \textbf{4.268} & \textbf{10.416} & \underline{0.369} & \textbf{1.673} \\
\cline{2-23}
 & \multirow{6}{*}{iTransformer} & MSE & 0.705 & 3.611 & 8.790 & \underline{0.559} & 1.672 & \underline{0.723} & \textbf{3.588} & \underline{8.386} & \textbf{0.578} & \underline{1.680} & \textbf{0.746} & \textbf{3.664} & \underline{8.245} & \textbf{0.560} & \textbf{1.696} & 0.803 & \textbf{3.749} & \underline{8.354} & \textbf{0.561} & \textbf{1.673} \\
 &  & MAE & 0.698 & 3.774 & 9.086 & 0.453 & 1.667 & 0.762 & 4.270 & 10.926 & 0.354 & 1.718 & 0.783 & 4.340 & 10.540 & 0.355 & 1.729 & 0.818 & 4.489 & 10.797 & 0.334 & 1.699 \\
 &  & TildeQ & 0.693 & \underline{3.602} & \underline{8.620} & 0.528 & \underline{1.657} & \textbf{0.721} & \underline{3.705} & 8.855 & \underline{0.532} & 1.692 & \underline{0.747} & \underline{3.803} & 8.626 & \underline{0.507} & 1.709 & \textbf{0.795} & 4.021 & 9.554 & \underline{0.466} & 1.682 \\
 &  & DBLoss & \textbf{0.685} & 3.612 & 8.748 & 0.523 & 1.663 & 0.732 & 3.917 & 9.829 & 0.477 & 1.692 & 0.757 & 4.024 & 9.656 & 0.447 & 1.709 & \underline{0.802} & 4.247 & 10.317 & 0.394 & 1.697 \\
 &  & PS & \underline{0.692} & \textbf{3.555} & 8.775 & 0.558 & 1.666 & 0.728 & 3.768 & 9.550 & 0.521 & 1.681 & 0.752 & 3.912 & 9.512 & 0.465 & 1.705 & 0.804 & 4.176 & 10.382 & 0.409 & 1.712 \\
 &  & \textbf{APAL} & 1.003 & 4.314 & \textbf{7.507} & \textbf{0.565} & \textbf{1.654} & 0.949 & 4.214 & \textbf{6.681} & 0.505 & \textbf{1.679} & 0.926 & 4.105 & \textbf{6.342} & 0.480 & \underline{1.697} & 0.830 & \underline{3.785} & \textbf{7.027} & 0.432 & \underline{1.674} \\
\bottomrule
\end{tabular}
}
\end{table*}

\subsection{Cross-Domain Generalization}
\label{sec:cross_domain}

Table~\ref{tab:mse_apal_comparison} compares MSE vs.\ APAL on public benchmarks under the same protocol. On datasets with recurrent peaks tied to operational constraints (Traffic, Electricity), APAL improves tail fidelity—e.g., on Traffic ($H=720$), it improves $\text{MSE}_{1}$ by \textbf{15\%} and Peak F1 by \textbf{22\%}. On noise-dominated series without operationally meaningful peaks (Exchange, Weather), APAL yields higher aggregate MSE: when tails carry low signal-to-noise, the asymmetric penalty introduces bias that degrades average-case performance. This confirms APAL is best deployed where under-prediction of peaks is more costly than over-prediction. Moreover, APAL is designed to exploit peak structures that are prominent, stable, and partly predictable. Appendix~\ref{app:applicability_diagnostics} provides a dataset diagnostic that checks whether a time series exhibits suitable peak structures for utilizing APAL. Full diagnostic table for all benchmark datasets are in Appendix~\ref{app:additional}.

\begin{table*}[t]
\centering
\caption{MSE vs APAL loss comparison (TSMixer) on $H \in \{96, 192, 336, 720\}$ for public benchmark datasets. Best results in \textbf{bold}.}
\label{tab:mse_apal_comparison}
\resizebox{\textwidth}{!}{%
\begin{tabular}{c|c|l|ccccc|ccccc|ccccc|ccccc}
\toprule
Dataset & Model & Loss & \multicolumn{5}{c|}{96} & \multicolumn{5}{c|}{192} & \multicolumn{5}{c|}{336} & \multicolumn{5}{c}{720} \\
\cmidrule(lr){4-8} \cmidrule(lr){9-13} \cmidrule(lr){14-18} \cmidrule(lr){19-23}
 &  &  & MSE & MSE$_{10}$ & MSE$_{1}$ & Peak F1 & PTE & MSE & MSE$_{10}$ & MSE$_{1}$ & Peak F1 & PTE & MSE & MSE$_{10}$ & MSE$_{1}$ & Peak F1 & PTE & MSE & MSE$_{10}$ & MSE$_{1}$ & Peak F1 & PTE \\
\midrule
\multirow{2}{*}{ETTh1} & \multirow{2}{*}{TSMixer} & MSE & \textbf{0.494} & 0.971 & 1.679 & 0.081 & \textbf{1.865} & \textbf{0.597} & 1.180 & 1.926 & 0.003 & \textbf{1.946} & \textbf{0.677} & 1.327 & 2.090 & 0.000 & \textbf{1.974} & \textbf{0.752} & 1.552 & 2.417 & 0.000 & 1.965 \\
 &  & \textbf{APAL} & 1.205 & \textbf{0.459} & \textbf{0.314} & \textbf{0.359} & 1.975 & 1.170 & \textbf{0.477} & \textbf{0.377} & \textbf{0.366} & 2.065 & 1.133 & \textbf{0.464} & \textbf{0.441} & \textbf{0.341} & 1.999 & 1.008 & \textbf{0.682} & \textbf{1.181} & \textbf{0.277} & \textbf{1.694} \\
\midrule
\multirow{2}{*}{ETTh2} & \multirow{2}{*}{TSMixer} & MSE & \textbf{1.056} & \textbf{1.071} & \textbf{1.558} & \textbf{0.327} & 1.703 & \textbf{2.587} & \textbf{1.242} & \textbf{0.950} & \textbf{0.303} & \textbf{1.657} & \textbf{2.407} & \textbf{1.178} & \textbf{1.072} & \textbf{0.296} & 1.691 & \textbf{2.051} & \textbf{1.198} & \textbf{1.442} & \textbf{0.280} & 1.703 \\
 &  & \textbf{APAL} & 5.194 & 2.818 & 1.824 & 0.315 & \textbf{1.631} & 9.791 & 6.694 & 4.857 & 0.298 & 1.672 & 7.300 & 4.377 & 3.107 & 0.291 & \textbf{1.687} & 5.353 & 2.696 & 2.029 & 0.269 & \textbf{1.702} \\
\midrule
\multirow{2}{*}{ETTm1} & \multirow{2}{*}{TSMixer} & MSE & \textbf{0.479} & 1.139 & 2.050 & 0.121 & 1.918 & \textbf{0.480} & 1.048 & 1.933 & 0.040 & 1.873 & \textbf{0.541} & 1.128 & 2.055 & 0.014 & \textbf{1.856} & \textbf{0.617} & 1.188 & 2.092 & 0.000 & \textbf{1.908} \\
 &  & \textbf{APAL} & 0.773 & \textbf{0.304} & \textbf{0.383} & \textbf{0.350} & \textbf{1.848} & 0.863 & \textbf{0.330} & \textbf{0.376} & \textbf{0.329} & \textbf{1.841} & 0.973 & \textbf{0.371} & \textbf{0.432} & \textbf{0.293} & 1.865 & 1.035 & \textbf{0.441} & \textbf{0.522} & \textbf{0.294} & 1.942 \\
\midrule
\multirow{2}{*}{ETTm2} & \multirow{2}{*}{TSMixer} & MSE & \textbf{0.250} & \textbf{0.234} & \textbf{0.439} & \textbf{0.296} & \textbf{1.732} & \textbf{0.492} & \textbf{0.531} & \textbf{0.910} & 0.254 & \textbf{1.708} & \textbf{0.833} & \textbf{0.956} & 1.424 & 0.238 & \textbf{1.695} & \textbf{2.541} & \textbf{1.659} & \textbf{1.515} & \textbf{0.238} & \textbf{1.706} \\
 &  & \textbf{APAL} & 1.321 & 0.773 & 0.651 & 0.281 & 1.779 & 2.022 & 1.146 & 0.973 & \textbf{0.260} & 1.789 & 2.604 & 1.418 & \textbf{1.222} & \textbf{0.246} & 1.787 & 9.655 & 7.610 & 6.298 & 0.235 & 1.769 \\
\midrule
\multirow{2}{*}{Electricity} & \multirow{2}{*}{TSMixer} & MSE & \textbf{0.204} & \textbf{0.309} & \textbf{0.507} & \textbf{0.571} & 0.769 & \textbf{0.218} & \textbf{0.303} & \textbf{0.480} & \textbf{0.576} & \textbf{0.811} & \textbf{0.240} & \textbf{0.322} & \textbf{0.490} & \textbf{0.549} & \textbf{0.859} & \textbf{0.272} & \textbf{0.370} & \textbf{0.580} & \textbf{0.538} & \textbf{0.983} \\
 &  & \textbf{APAL} & 0.489 & 0.546 & 0.547 & 0.510 & \textbf{0.722} & 0.537 & 0.595 & 0.562 & 0.489 & 0.861 & 0.601 & 0.689 & 0.654 & 0.466 & 0.973 & 0.651 & 0.772 & 0.725 & 0.449 & 1.174 \\
\midrule
\multirow{2}{*}{Traffic} & \multirow{2}{*}{TSMixer} & MSE & \textbf{0.539} & 1.772 & 2.796 & 0.538 & 0.579 & \textbf{0.545} & \textbf{1.673} & 2.514 & 0.750 & 0.670 & \textbf{0.570} & \textbf{1.797} & 2.920 & 0.662 & 0.608 & \textbf{0.621} & \textbf{1.999} & 3.174 & 0.669 & 0.615 \\
 &  & \textbf{APAL} & 1.055 & \textbf{1.645} & \textbf{1.949} & \textbf{0.730} & \textbf{0.402} & 0.994 & 1.939 & \textbf{2.210} & \textbf{0.799} & \textbf{0.416} & 1.081 & 2.233 & \textbf{2.553} & \textbf{0.794} & \textbf{0.481} & 1.021 & 2.190 & \textbf{2.698} & \textbf{0.817} & \textbf{0.401} \\
\midrule
\multirow{2}{*}{Weather} & \multirow{2}{*}{TSMixer} & MSE & \textbf{0.180} & \textbf{0.407} & 0.770 & \textbf{0.266} & \textbf{1.784} & \textbf{0.218} & \textbf{0.464} & \textbf{0.868} & \textbf{0.269} & 1.828 & \textbf{0.261} & \textbf{0.519} & \textbf{0.931} & \textbf{0.257} & 1.832 & \textbf{0.318} & \textbf{0.619} & \textbf{1.050} & \textbf{0.233} & 1.830 \\
 &  & \textbf{APAL} & 0.567 & 0.711 & \textbf{0.706} & 0.257 & 1.810 & 0.682 & 0.793 & 0.872 & 0.231 & \textbf{1.822} & 0.809 & 0.820 & 0.934 & 0.215 & \textbf{1.804} & 0.944 & 0.902 & 1.080 & 0.201 & \textbf{1.817} \\
\midrule
\multirow{2}{*}{Exchange} & \multirow{2}{*}{TSMixer} & MSE & \textbf{0.129} & \textbf{0.146} & \textbf{0.160} & \textbf{0.228} & 1.689 & \textbf{0.223} & \textbf{0.292} & \textbf{0.289} & \textbf{0.205} & \textbf{1.691} & \textbf{0.364} & \textbf{0.532} & \textbf{0.508} & \textbf{0.185} & \textbf{1.709} & \textbf{0.758} & \textbf{0.906} & \textbf{0.750} & \textbf{0.230} & 1.719 \\
 &  & \textbf{APAL} & 0.563 & 0.560 & 0.352 & 0.196 & \textbf{1.688} & 1.221 & 1.117 & 0.731 & 0.179 & 1.711 & 2.499 & 1.707 & 1.078 & 0.169 & 1.723 & 8.384 & 3.817 & 1.759 & 0.158 & \textbf{1.706} \\
\bottomrule
\end{tabular}
}
\end{table*}

\subsection{Ablation Study and Sensitivity Analysis}
\label{sec:ablation}

\begin{figure}[ht]
\centering
\includegraphics[width=0.60\linewidth]{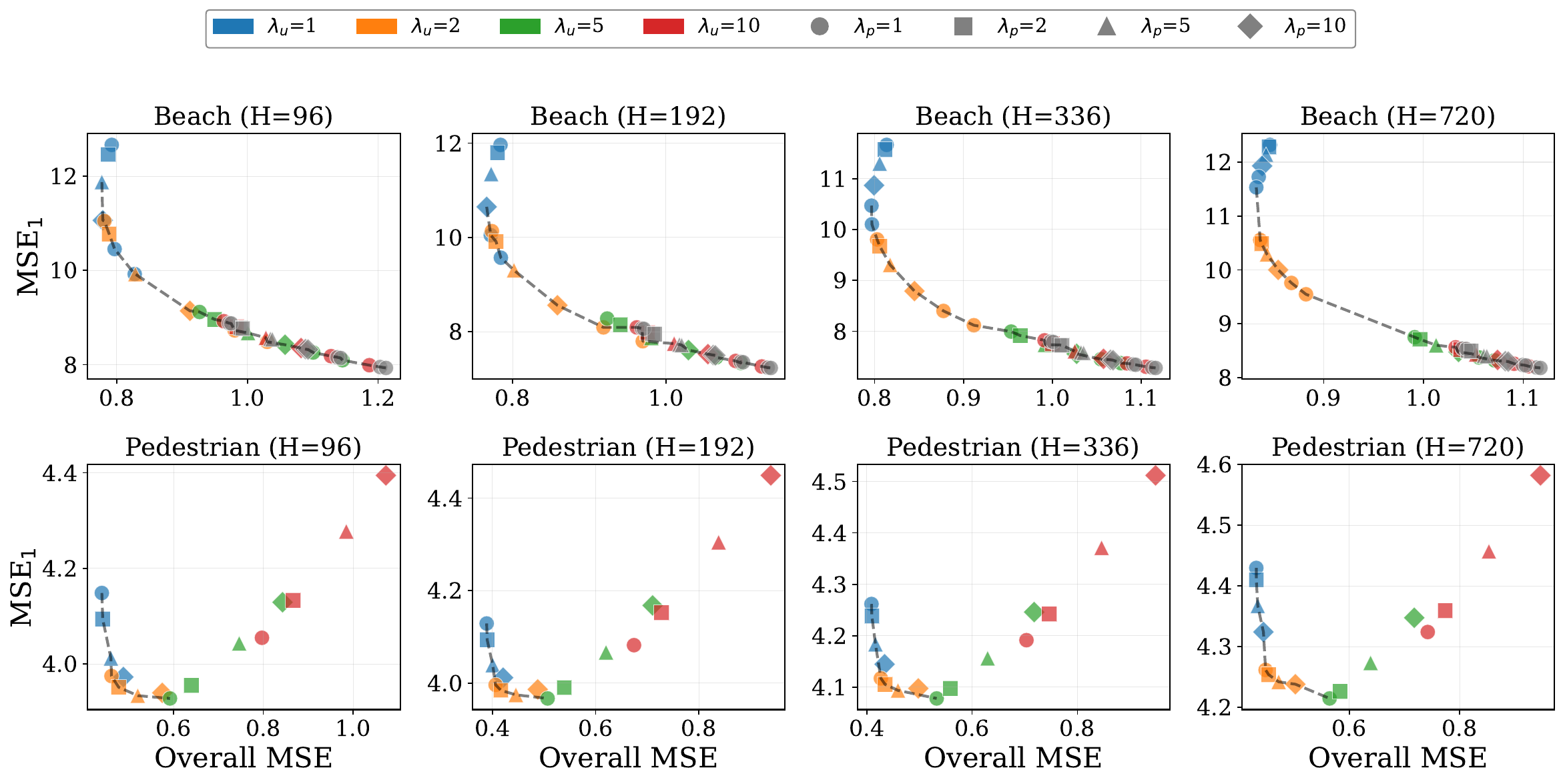}
\caption{
Pareto trade-off between overall MSE and tail $\text{MSE}_{1}$ across APAL configurations with $\tau=0.9$. Each subplot is one dataset-horizon combination; dashed lines mark Pareto-optimal frontiers. $\lambda_u$: upweight factor, $\lambda_p$: peak weight.}
\label{fig:pareto}
\end{figure}

We analyse APAL's sensitivity to ($\lambda_u$, $\lambda_p$, $\tau$) on Pedestrian and Beach datasets using DLinear, with TSMixer cross-domain grids for ETT, Exchange, and Weather datasets in Appendix~\ref{app:cross_domain_ablation}. The grid covers $\lambda_p, \lambda_u \in \{1,2,5,10\}$ and $\tau \in \{0.8,0.9,0.95\}$ for Pedestrian, extended to $\{1,2,5,10,15,20\}$ for Beach. Figure~\ref{fig:pareto} visualises the Pareto trade-off between aggregate MSE and tail $\text{MSE}_1$, exposing achievable operating points so practitioners can match their domain-specific cost model. Detailed marginal-effect plots, heatmaps, and cross-domain ablation summaries are in Appendix~\ref{app:additional}.

Based on our peak-critical ablations, we recommend the conservative setting $\lambda_u = 2$, $\lambda_p = 2$, $\tau = 0.9$ as a starting point rather than as a universal optimum. On the pedestrian-focused ablations this setting gives useful tail improvements with a moderate aggregate-error cost, while the cross-domain grids show that the same default is not uniformly preferable on Exchange or all ETT variants. For applications where peak misses are more costly, increasing $\lambda_u$ or $\lambda_p$ can improve $\text{MSE}_1$ or Peak~F1 at the cost of higher aggregate error and possible overestimation; when overestimation is costly, users should reduce $\lambda_u$ and $\lambda_p$ toward 1 or prefer a symmetric loss. We recommend selecting the final operating point on a held-out validation set using both aggregate and peak-critical metrics, ideally with an application-specific cost model.

\subsection{Channel-wise pedestrian results}
\label{sec:channelwise-results}
\begin{figure}[ht]
    \centering
    \begin{subfigure}[b]{0.49\linewidth}
        \centering
        \includegraphics[width=\linewidth]{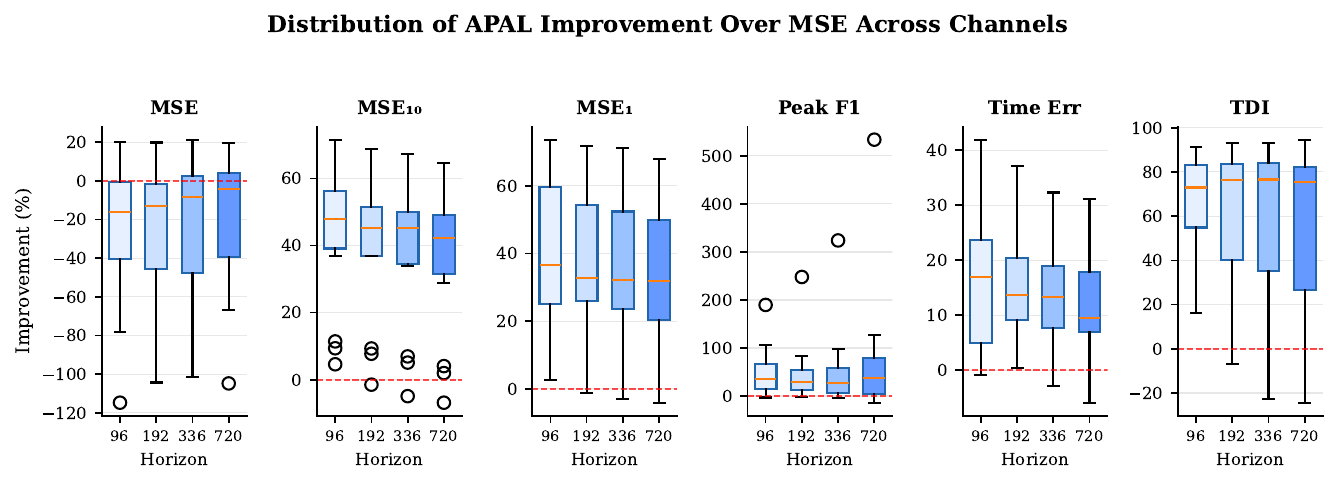}
        \caption{Pedestrian Dataset}
        \label{fig:channelboxPed}
    \end{subfigure}
    \hfill
    \begin{subfigure}[b]{0.49\linewidth}
        \centering
        \includegraphics[width=\linewidth]{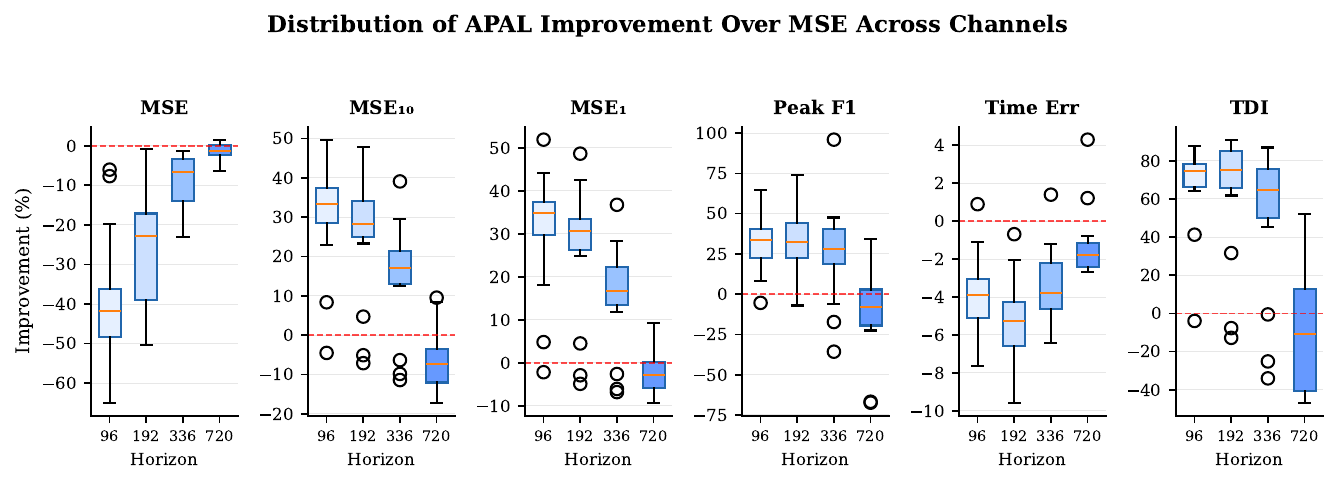}
        \caption{Beach Dataset}
        \label{fig:channelboxBeach}
    \end{subfigure}
     \caption{Channel-wise percentage improvements of APAL over MSE (TSMixer) on datasets \textbf{(a) Pedestrian} and \textbf{(b) Beach}. Each boxplot covers 16 channels at a given horizon; positive values indicate APAL outperforms MSE; the red dashed line marks zero improvement.}
    \label{fig:channelboxCombined}
\end{figure}

We report per-sensor results for Melbourne (Figure~\ref{fig:channelboxPed}) and the Beach Dataset (Figure~\ref{fig:channelboxBeach}), including tail-error and peak-event metrics, to reflect heterogeneity across sensor locations (``channels" in figure). The results show that APAL consistently improves peak-sensitive metrics across channels, with median improvements of 5--15\% for extreme value prediction.

\subsection{Training Stability and Compute}
\label{sec:compute}

APAL adds only element-wise weighing with negligible overhead: $+0.23$\,s/epoch ($\approx 4\%$) vs.\ MSE, compared to DBLoss ($+0.85$\,s, 15\%), PS ($+1.37$\,s, 24\%), and TildeQ ($+2.15$\,s, 37\%). We report the epoch time and total training time in detail in Appendix~\ref{app:compute}. This minor computational overhead compared to MSE makes APAL practical for real-time operational training pipelines.

\subsection{Failure Cases}
\label{sec:limitations}
APAL intentionally changes the optimization target and can therefore be harmful when its cost assumptions do not match the deployment setting. First, because under-predictions are penalized more heavily, APAL can introduce upward bias or false-positive peaks; this is acceptable only when missed peaks are more costly than overestimation. Second, the within-window peak mask can emphasize random fluctuations in flat, low-variance, or sensor-noisy windows, because every window has a relative maximum. Third, APAL can over-commit to historical peak shapes when future peaks shift in timing or magnitude. The current experiments use direct multi-step forecasting, so they do not test recursive rollouts, in which directional bias could accumulate. These limitations motivate conservative validation-based tuning, drift monitoring, and fallback to symmetric losses when peak structure is weak, or overestimation is equally costly.

\section{Conclusion and Future Work}
\label{sec:conclusion}
We studied peak-critical long-term forecasting settings where under-prediction incurs disproportionate operational costs and rare peaks dominate downstream risk.
We introduced Asymmetric Peak-Aware Loss (APAL), a simple model-agnostic objective that penalises under-predictions more heavily and increases training emphasis on peak regions through adaptive within-horizon thresholding.
We also proposed a peak-critical evaluation protocol that complements aggregate metrics with channel-wise tail-error, peak-detection, and peak timing measures.

Across extensive experiments spanning five state-of-the-art backbones, four horizons, and ten datasets, we demonstrated that APAL achieves the best Top-1\% tail error in 39 of 40 peak-critical settings, with reductions of up to 31\% compared to MSE loss. On pedestrian demand forecasting, APAL improves Peak F1 by up to 88\% while incurring only 4\% computational overhead (3--9$\times$ lower than competing peak-aware losses). APAL exposes a controllable trade-off allowing practitioners to tune $(\lambda_u, \lambda_p, \tau)$ to balance aggregate error with peak-critical performance using domain-specific cost functions. We also introduced a quantitative pretraining diagnostic that classifies datasets into Strongly Seasonal and Irregularly Structured and Weakly Structured categories. The benefits of APAL are strongest on datasets with a clear peak structure. When the peaks are not identifiable, learnable, or operationally meaningful, symmetric losses remain preferable. Our findings confirm that incorporating cost-sensitive inductive biases directly into the loss function offers a controllable and effective mechanism for enhancing safety-critical forecasting performance, providing a practical alternative to complex probabilistic models in settings where point forecasts are required for downstream optimisation and an accurate peak forecast is a priority.

Future work will extend APAL's asymmetric weighing principles to probabilistic forecasting frameworks, enabling calibrated uncertainty estimation alongside peak-aware point predictions. We also plan to integrate explicit domain cost models directly into the hyperparameter selection criteria, allowing models to be automatically tuned against real-world financial or safety constraints rather than proxy metrics like MSE. Finally, we aim to rigorously study the robustness of peak-aware objectives under distribution shift and rare-event regime changes, ensuring that models remain stable and reliable even when the statistical properties of extreme events evolve over time.


\bibliographystyle{abbrv}
{\small
\bibliography{example_paper}}

\section*{Impact Statement}
\label{sec:impact}
This work addresses forecasting settings in which forecasting failures during high-demand spikes can have outsized operational consequences, such as crowd management, staffing, and transport planning.
By improving peak-event detection and timing prediction, peak-critical forecasting can support safer resource allocation and more robust contingency planning.

Potential risks arise if forecasts are used to automate decisions without oversight.
Overestimation can cause inefficient allocation, while underestimation during sensitive events can create safety hazards.
In addition, sensor coverage and historical patterns may reflect structural biases, and models may behave differently across locations or demographic contexts.
We recommend that deployments (i) adjust asymmetric penalties using stakeholder-defined costs, (ii) report both aggregate and peak-critical metrics with channel-wise breakdowns, (iii) monitor for distribution shift and sensor changes over time, and (iv) include human-in-the-loop review for high-impact decisions.


\newpage
\appendix
\onecolumn

\section{Datasets}
\label{app:datasets}
This appendix provides comprehensive documentation of all datasets used in our experiments, including sources, preprocessing steps, and train/validation/test splits. Appendix~\ref{app:melbourne_ped} describes the Melbourne Pedestrian Dataset, an open production-ready benchmark for urban crowd forecasting. Appendix~\ref{app:beach} documents the Beach Visitor Count Dataset from Scheveningen, Netherlands. Appendix~\ref{app:public} summarizes eight public benchmarks spanning energy, finance, transportation, and meteorology domains. Finally, Appendix~\ref{app:data_impl} details the normalization and splitting procedures applied across all datasets.

\subsection{Melbourne Pedestrian Dataset (2010--2017)}
\label{app:melbourne_ped}

\paragraph{Official source.}
The raw pedestrian counts are published on the City of Melbourne Open Data Portal as the \emph{Pedestrian Counting System (counts per hour)} dataset, which contains hourly pedestrian counts from sensor devices distributed across the city \cite{melbourne_ped_counts_hourly}.
Sensor metadata (status, location, and directional information) is provided in the companion \emph{Sensor Locations} dataset \cite{melbourne_ped_sensor_locations}.
The City also provides an online visualisation and download interface for the pedestrian counting system \cite{melbourne_ped_website}.

In this study, we release a production-ready multivariate time series derived from the official hourly counts, filtered to 16 sensors in the Melbourne CBD with complete data coverage from 2010-01-01 00:00:01 to 2017-12-31 23:00:01 at hourly frequency.
The released file is \texttt{pedestrian\_tsf\_2010\_2017\_complete.csv} and contains 70{,}128 hourly timestamps (2{,}922 days, 8 years), with no missing values and no gaps in the time index. The processed dataset contains 17 columns: a timestamp column (\texttt{date}, format \texttt{YYYY-MM-DD HH:MM:SS}) and 16 sensor channels.
Each channel records the pedestrian count per hour at one location.
The included sensors are:
\texttt{T1, T2, T3, T4, T5, T6, T7, T8, T9, T10, T11, T12, T14, T15, T17, T18}. A value of zero indicates that no pedestrians passed under a sensor during that hour (as specified by the data provider)~\cite{melbourne_ped_counts_hourly}.

\paragraph{Summary statistics.}
Across all sensors and timestamps, the dataset contains 1{,}122{,}048 observations with mean 670.35, standard deviation 897.68, minimum 0, and maximum 11{,}742.
Per-sensor descriptive statistics (mean, standard deviation, minimum, maximum, median) are reported in Table~\ref{tab:melb_stats}.
We emphasize that different sensors exhibit substantially different scales and tail behavior, motivating channelwise reporting for peak and tail metrics.

\begin{table*}[h]
\centering
\caption{Melbourne pedestrian dataset statistics (2010--2017) for the 16 sensors included in the released complete-coverage subset.}
\label{tab:melb_stats}
\begin{tabular}{lrrrrr}
\toprule
Sensor & Mean & Std & Min & Max & Median \\
\midrule
T1  & 1173.9 & 1211.5 & 0 & 5573  & 690.0 \\
T2  & 1114.4 & 1230.8 & 0 & 6942  & 526.0 \\
T3  & 1231.4 & 982.1  & 0 & 5890  & 1059.0 \\
T4  & 1508.0 & 1271.5 & 0 & 8052  & 1234.0 \\
T5  & 1068.2 & 888.8  & 0 & 7391  & 988.0 \\
T6  & 1157.3 & 997.1  & 0 & 6568  & 1020.0 \\
T7  & 374.3  & 711.7  & 0 & 11742 & 163.0 \\
T8  & 146.6  & 155.4  & 0 & 3009  & 106.0 \\
T9  & 476.9  & 703.8  & 0 & 4272  & 121.0 \\
T10 & 172.8  & 202.3  & 0 & 3113  & 98.0 \\
T11 & 101.6  & 213.7  & 0 & 9805  & 53.0 \\
T12 & 198.8  & 276.9  & 0 & 11284 & 144.0 \\
T14 & 387.5  & 319.5  & 0 & 7304  & 355.0 \\
T15 & 805.5  & 710.9  & 0 & 5559  & 681.0 \\
T17 & 461.4  & 500.3  & 0 & 3889  & 263.0 \\
T18 & 347.1  & 443.7  & 0 & 3759  & 143.5 \\
\bottomrule
\end{tabular}
\end{table*}

\paragraph{Sensor locations.}
We provide sensor coordinates (WGS84, EPSG:4326) by joining the hourly counts with the official sensor locations dataset via \texttt{sensor\_id}~\cite{melbourne_ped_sensor_locations}.
Table~\ref{tab:melb_locations} lists the locations for the sensors in our subset.

\begin{table*}[t]
\centering
\caption{Sensor locations for the 16 sensors in the Melbourne subset (WGS84).}
\label{tab:melb_locations}
\begin{tabular}{llllrr}
\toprule
Sensor & Location name & Latitude & Longitude \\
\midrule
T1  & Bourke Street Mall (North)        & -37.813494 & 144.965153 \\
T2  & Bourke Street Mall (South)        & -37.814067 & 144.965186 \\
T3  & Melbourne Central                 & -37.811015 & 144.964295 \\
T4  & Town Hall (West)                  & -37.814880 & 144.966088 \\
T5  & Princes Bridge                    & -37.818742 & 144.967877 \\
T6  & Flinders Street Station Underpass & -37.818479 & 144.966715 \\
T7  & Birrarung Marr                    & -37.818617 & 144.972484 \\
T8  & Webb Bridge                       & -37.822723 & 144.947677 \\
T9  & Southern Cross Station            & -37.819830 & 144.951026 \\
T10 & Victoria Point                    & -37.818765 & 144.947105 \\
T11 & Waterfront City                   & -37.815650 & 144.939707 \\
T12 & New Quay                          & -37.814580 & 144.942924 \\
T14 & Sandridge Bridge                  & -37.820112 & 144.962919 \\
T15 & State Library                     & -37.810430 & 144.964388 \\
T17 & Collins Place (South)             & -37.813458 & 144.972984 \\
T18 & Collins Place (North)             & -37.813449 & 144.973054 \\
\bottomrule
\end{tabular}
\end{table*}

\paragraph{Intended use and evaluation.}
The dataset exhibits strong weekly seasonality, location heterogeneity, and sharp peaks driven by commuting patterns and events.
These properties make it a realistic dataset for peak-critical objectives, where underestimation and mis-timing of extreme pedestrian volumes can be operationally costly.

\subsection{Beach Visitor Count Dataset}
\label{app:beach}
This study utilizes the historical pedestrian crowd count dataset from Scheveningen Beach, The Hague, Netherlands, provided by a third-party data provider RESONO\footnote{\url{https://reso.no/}}. The dataset comprises hourly visitor counts for 16 different areas, collected from 01 April 2021 to 17 November 2023. These counts serve as a proxy for crowd count, derived from aggregated and anonymised location data from mobile applications on devices of users who have provided consent~\cite{resono2025veelgestelde}. The data is collected by counting devices that share opt-in location data through apps in RESONO's mobile panel and is then scaled using a statistical model to estimate the total number of visitors~\cite{resono2025veelgestelde}.

\subsection{Public Benchmark Dataset}
\label{app:public}
Apart from the \textit{human mobility domain} (Beach, Pedestrian), we also evaluated our proposed APAL loss on eight publicly available time-series forecasting benchmarks spanning diverse application domains, including \textit{energy systems} (ETT, Electricity), \textit{finance} (Exchange), \textit{transportation} (Traffic) and  \textit{meteorology} (Weather).

\paragraph{ETTh1 \& ETTh2 (Electricity Transformer Temperature - Hourly)} datasets~\cite{Zhou_Zhang_Peng_Zhang_Li_Xiong_Zhang_2021} contain hourly measurements from electricity transformers in China, collected from July 2016 to July 2018. Each record includes the oil temperature (target variable) and six power load features (HUFL, HULL, MUFL, MULL, LUFL, LULL) representing high, medium, and low useful/useless load. The two variants correspond to different transformer stations and capture distinct operational patterns. These datasets are widely used benchmarks for evaluating long-term forecasting models.

\paragraph{ETTm1 \& ETTm2 (Electricity Transformer Temperature - Minute)} datasets~\cite{Zhou_Zhang_Peng_Zhang_Li_Xiong_Zhang_2021} are the 15-minute interval counterparts of ETTh1 and ETTh2, providing four times the temporal resolution. The finer granularity enables evaluation of models on short-term fluctuations while maintaining the same feature set. This increased frequency results in 69,680 time steps, making them suitable for assessing scalability.

\paragraph{Electricity Consuming Load (ECL)} dataset~\cite{Zhou_Zhang_Peng_Zhang_Li_Xiong_Zhang_2021} records hourly electricity consumption (in kWh) of 321 clients from 2012 to 2014. This high-dimensional dataset exhibits strong daily and weekly periodicity patterns driven by human activity cycles. The large number of channels (321) makes it particularly challenging for multivariate forecasting methods.

\paragraph{Exchange}dataset~\cite{10.1145/3209978.3210006} contains daily exchange rates of eight countries' currencies (Australian Dollar, British Pound, Canadian Dollar, Swiss Franc, Chinese Yuan, Japanese Yen, New Zealand Dollar, and Singapore Dollar) against the US Dollar, spanning from 1990 to 2016. Unlike other datasets, Exchange exhibits non-stationary trends and lacks clear seasonal patterns, making it a challenging benchmark for trend-sensitive forecasting.

\paragraph{Traffic} dataset~\cite{Zhou_Zhang_Peng_Zhang_Li_Xiong_Zhang_2021} measures hourly road occupancy rates from 862 sensors deployed on San Francisco Bay Area freeways from January 2015 to December 2016. The data captures characteristic rush-hour peaks, weekly commuting patterns, and holiday effects. With 862 channels, it tests the scalability of models on high-dimensional spatial data.

\paragraph{Weather} dataset~\cite{Zhou_Zhang_Peng_Zhang_Li_Xiong_Zhang_2021} contains 21 meteorological indicators recorded every 10 minutes throughout 2020 from the Max Planck Institute weather station in Germany. Features include air temperature, atmospheric pressure, humidity, wind speed and direction, precipitation, and solar radiation. The high sampling frequency (52,696 time steps) and diverse physical quantities make it suitable for evaluating models on complex environmental dynamics.

\subsection{Dataset Implementation details}
\label{app:data_impl}
We evaluated our proposed loss function across ten time-series forecasting benchmark datasets. Table~\ref{tab:datasets} provides a comprehensive summary including the number of variates (channels), sequence length, train/validation/test splits, and sampling frequency. The datasets span multiple application domains: electricity transformer monitoring (ETT), power consumption (ECL), currency exchange (Exchange), road traffic (Traffic), meteorology (Weather), tourism (Beach), and urban mobility (Pedestrian). Following prior work~\cite{autoformer, Zhou_Zhang_Peng_Zhang_Li_Xiong_Zhang_2021, nie2023patchtst}, we apply Z-score normalization using statistics computed solely from the training set, with inverse transformation applied at evaluation time. ETT datasets use fixed chronological splits (12/4/4 months). All other datasets use a 70\%/10\%/20\% train/validation/test chronological split. Z-score normalization is applied using training set statistics only.

\begin{table}[htbp]
\centering
\caption{Summary of Datasets Used in Experiments}
\label{tab:datasets}
\resizebox{\textwidth}{!}{%
\begin{tabular}{lccccl}
\toprule
\textbf{Dataset} & \textbf{Channels} & \textbf{Length} & \textbf{(Train / Val / Test)} & \textbf{Frequency} & \textbf{Description} \\
\midrule
ETTh1 & 7 & 17,420 & (8,640 / 2,880 / 2,880) & 1 hour & Electricity transformer temperature \\
ETTh2 & 7 & 17,420 & (8,640 / 2,880 / 2,880) & 1 hour & Electricity transformer temperature \\
ETTm1 & 7 & 69,680 & (34,560 / 11,520 / 11,520) & 15 min & Electricity transformer temperature \\
ETTm2 & 7 & 69,680 & (34,560 / 11,520 / 11,520) & 15 min & Electricity transformer temperature \\
ECL & 321 & 26,304 & (18,412 / 2,632 / 5,260) & 1 hour & Electricity consumption of 321 clients \\
Exchange & 8 & 7,587 & (5,310 / 760 / 1,517) & 1 day & Daily exchange rates of 8 currencies \\
Traffic & 862 & 17,544 & (12,280 / 1,756 / 3,508) & 1 hour & Road occupancy rates from 862 sensors \\
Weather & 21 & 52,696 & (36,887 / 5,270 / 10,539) & 10 min & 21 meteorological indicators \\
Beach & 16 & 23,031 & (16,121 / 2,304 / 4,606) & 1 hour & Hourly beach visitor counts \\
Pedestrian & 16 & 70,128 & (49,089 / 7,014 / 14,025) & 1 hour & Melbourne pedestrian counts \\
\bottomrule
\end{tabular}%
}
\end{table}

\section{Model Descriptions}
\label{app:models}

This appendix provides detailed descriptions of the five state-of-the-art forecasting backbones used in our experiments. We selected models spanning diverse architectural families, including linear, MLP-based, and Transformer-based architectures, to demonstrate that APAL's benefits are model-agnostic. Training hyperparameters and optimisation settings are reported in Appendix~\ref{sec:hyperparameters}.

\paragraph{DLinear~\cite{10.1609/aaai.v37i9.26317}.}
DLinear is a surprisingly simple yet effective baseline that challenges the necessity of complex architectures for time series forecasting. The model decomposes the input sequence into trend and seasonal components using a moving average kernel, then applies two separate linear layers to each component. The final prediction is the sum of both projections. Despite its simplicity, DLinear achieves competitive performance against Transformer-based models while being significantly faster to train and having fewer parameters. The model demonstrates that for many time series forecasting tasks, the temporal relationships can be effectively captured by simple linear mappings.

\paragraph{PatchTST~\cite{nie2023patchtst}.}
PatchTST (Patch Time Series Transformer) introduces two key innovations for time series forecasting: (1) \textit{patching}, which segments the time series into subseries-level patches that serve as input tokens, reducing computational complexity and enabling the model to capture local semantic information; and (2) \textit{channel-independence}, where each channel is processed independently by a shared Transformer backbone, improving generalization and reducing the risk of overfitting. The patching mechanism also allows the model to attend to longer historical context by reducing the number of tokens while preserving temporal information.

\paragraph{iTransformer~\cite{liu2024itransformer}.}
iTransformer (Inverted Transformer) proposes an inverted view of time series Transformers by applying attention across the variate (channel) dimension rather than the temporal dimension. In this architecture, each time step's multivariate observation is treated as a single token, and the self-attention mechanism learns dependencies between different variates. This design is particularly effective for multivariate forecasting where cross-channel correlations are important. The temporal patterns within each variate are captured through feed-forward networks applied independently to each channel.

\paragraph{TiDE~\cite{das2023tide}.}
TiDE (Time-series Dense Encoder) is an MLP-based model that achieves strong performance through a simple encoder-decoder architecture. The encoder maps the historical time series and covariates to a dense representation, which is then decoded to produce multi-step predictions. TiDE uses residual connections and layer normalization for stable training. The model is designed to be computationally efficient while capturing both temporal dynamics and covariate effects. Its simplicity makes it particularly suitable for large-scale deployment scenarios.

\paragraph{TSMixer~\cite{TSMixer}} 
TSMixer (Time-Series Mixer) is an MLP-based forecasting model inspired by Mixer architectures that alternates mixing operations across the temporal and channel dimensions. Instead of attention, it uses lightweight fully connected blocks to exchange information along time steps and variates, yielding competitive accuracy with high computational efficiency.
This makes TSMixer a strong backbone for isolating the effect of training objectives, since it is simple, fast to train, and widely used as a non-Transformer baseline in long-horizon forecasting.

\subsection{Training Hyperparameters}
\label{sec:hyperparameters}

All experiments are conducted using the Time-Series-Library framework~\cite{wang2024tssurvey} with consistent training protocols. Table~\ref{tab:common_hyperparams} presents the common hyperparameters shared across all models, and Table~\ref{tab:model_hyperparams} details the model-specific configurations. Comparison losses (TILDE-Q, PS, DBLoss) are implemented using author-recommended settings when available and are not additionally tuned beyond shared training hyperparameters.

\paragraph{Common Training Settings.}
All models are trained using the Adam optimizer with a learning rate scheduler (type1: halving the learning rate after each epoch). We use early stopping with a patience of 3 epochs to prevent overfitting. The input sequence length is fixed at 96 time steps with a label length of 48 for decoder-based models. We evaluate four prediction horizons: $\{96, 192, 336, 720\}$ time steps. 
The same model configurations are applied consistently across all datasets to ensure fair comparison.

\begin{table}[ht]
\centering
\caption{Common Training Hyperparameters}
\label{tab:common_hyperparams}
\small
\begin{tabular}{lc}
\toprule
\textbf{Hyperparameter} & \textbf{Value} \\
\midrule
Input sequence length (seq\_len) & 96 \\
Label length (label\_len) & 48 \\
Prediction lengths (pred\_len) & \{96, 192, 336, 720\} \\
Training epochs & 10 or 30$^\ast$ \\
Early stopping patience & 3 \\
Optimizer & Adam \\
Learning rate scheduler & type1 (halving) \\
Random seed & 2021 \\
Features & Multivariate (M) \\
\bottomrule
\end{tabular}
\vspace{1mm}
\\\footnotesize $^\ast$ Training epochs vary by dataset: 10 epochs for ETT, Electricity, Exchange, Traffic, and Weather; 30 epochs for Beach, and Pedestrian.
\end{table}

\begin{table*}[ht]
\centering
\caption{Model-Specific Hyperparameters}
\label{tab:model_hyperparams}
\small
\begin{tabular}{lccccc}
\toprule
\textbf{Hyperparameter} & \textbf{DLinear} & \textbf{PatchTST} & \textbf{iTransformer} & \textbf{TiDE} & \textbf{TSMixer} \\
\midrule
Encoder layers (e\_layers) & 2 & 2 & 3 & 2 & 2 \\
Decoder layers (d\_layers) & 1 & 1 & 1 & 2 & 1 \\
Model dimension (d\_model) & -- & 512 & 512 & 256 & 32 \\
Feed-forward dimension (d\_ff) & -- & 2048 & 512 & 256 & 32 \\
Attention heads (n\_heads) & -- & 4/16$^\dagger$ & 8 & -- & -- \\
Dropout & 0.1 & 0.1 & 0.1 & 0.3 & 0.1 \\
Batch size & 32 & 32/128$^\ddagger$ & 32 & 512 & 32 \\
Learning rate & 0.0001 & 0.0001 & 0.0001 & 0.1 & 0.0001 \\
Factor & 3 & 3 & 3 & -- & 3 \\
\bottomrule
\end{tabular}
\vspace{1mm}
\\\footnotesize $^\dagger$ PatchTST uses 16 heads for pred\_len=192, and 4 heads otherwise.
$^\ddagger$ PatchTST uses batch size 128 for pred\_len $\geq$ 336, and 32 otherwise.
\end{table*}

\section{Baselines and Comparison Loss Functions}

\label{app:loss_baselines}
We compare APAL against MAE, MSE, TILDE-Q~\cite{tildeqloss}, DBLoss~\cite{dbloss}, and Patch-wise Structural (PS) loss~\cite{patchloss} using the authors' recommended settings where available, with a consistent training protocol across losses. Additionally, we also report the results of Pinball Loss~\cite{SMYL2026103637} compared to APAL in Section~\ref{app:pinball}.

\paragraph{MAE and MSE (symmetric point losses).}
MAE and MSE treat over- and under-prediction symmetrically and weight all horizon steps equally. They provide strong average-case performance but can encourage \emph{peak smoothing} when extreme values are rare.

\paragraph{TILDE-Q (shape-aware loss).}
TILDE-Q~\cite{tildeqloss} is a lightweight loss designed to improve \emph{shape fidelity} by reducing sensitivity to temporal distortions. It jointly accounts for amplitude and phase mismatch via transformation-invariant comparisons between predicted and ground-truth trajectories, encouraging the forecast to preserve local extrema and overall temporal structure rather than optimizing only pointwise error.

\paragraph{DBLoss (decomposition-based loss).}
DBLoss~\cite{dbloss} decomposes the target sequence within the prediction horizon into multiple components (e.g., trend and seasonal/residual) using exponential moving averages. Separate loss terms are then applied to each component and combined with user-defined weights. This encourages the model to fit both long-term trend and short-term fluctuations, often improving overall trajectory structure.

\paragraph{Patch-wise Structural (PS) loss.}
PS loss~\cite{patchloss} compares predictions and ground truth at the \emph{patch} (local window) level using simple statistics such as correlation, mean, and variance. By matching local distributional/shape properties across patches, PS encourages structural alignment over short horizons and can be combined with a pointwise term to retain absolute accuracy.

\paragraph{Relation to peak-critical forecasting.}
TILDE-Q, DBLoss, and PS are primarily designed to improve global/patch-level shape agreement and are typically symmetric with respect to under- vs.\ over-prediction. Consequently, their gradient signal can still be diluted on rare extreme points, motivating our peak-aware and asymmetric objective.

\section{Evaluation Metrics: Definitions and Edge Cases}
\label{app:metrics}
This appendix provides formal definitions and implementation details for all evaluation metrics used in our experiments. Appendix~\ref{app:standard} defines standard aggregate metrics (MAE, MSE). Appendix~\ref{app:tail_metrics} introduces tail-focused metrics that restrict error computation to extreme ground-truth values. Appendix~\ref{app:peak_metrics} describes our event-based peak detection and matching protocol. Finally, Appendix~\ref{app:diagnostics} defines supplementary diagnostics including PCC and TDI.

\subsection{Standard metrics}
\label{app:standard}
We report MAE and MSE over all forecast points and channels after inverse transforming predictions to the original scale when normalization is applied.
Let $N$ denote the total number of evaluated samples.
For each sample $n$, let $\hat{y}^{(n)}_{h,c}$ and $y^{(n)}_{h,c}$ denote the prediction and ground truth at horizon step $h$ and channel $c$.
The standard metrics are defined as:
\begin{align}
\mathrm{MAE} &= \frac{1}{N \cdot H \cdot C} \sum_{n=1}^{N} \sum_{h=1}^{H} \sum_{c=1}^{C} \left| y^{(n)}_{h,c} - \hat{y}^{(n)}_{h,c} \right|, \\
\mathrm{MSE} &= \frac{1}{N \cdot H \cdot C} \sum_{n=1}^{N} \sum_{h=1}^{H} \sum_{c=1}^{C} \left( y^{(n)}_{h,c} - \hat{y}^{(n)}_{h,c} \right)^2.
\end{align}
Both metrics treat all time steps and channels equally, which can mask failures on rare extreme values.

\subsection{Tail metrics (Top-10\%, Top-1\%)}
\label{app:tail_metrics}
Tail metrics restrict error computation to extreme ground-truth values, computed per channel and macro-averaged to avoid high-variance channels dominating.

For each channel $c$, let $\mathcal{S}_c = \{y^{(n)}_{h,c}\}$ denote all points flattened over samples and horizon steps, with $|\mathcal{S}_c| = N \cdot H$.
For tail fraction $p \in \{0.10, 0.01\}$, let $k=\lceil p \cdot N \cdot H \rceil$ and $\mathcal{I}^{(p)}_c$ be the indices of the top-$k$ ground-truth values.
The tail metrics are:
\begin{align}
\mathrm{MSE}^{(p)}_c &= \frac{1}{k}\sum_{i \in \mathcal{I}^{(p)}_c} \left(y_{i,c} - \hat{y}_{i,c}\right)^2, &
\mathrm{MAE}^{(p)}_c &= \frac{1}{k}\sum_{i \in \mathcal{I}^{(p)}_c} \left|y_{i,c} - \hat{y}_{i,c}\right|,
\end{align}
with reported values as macro-averages: $\mathrm{MSE}_{p} = \frac{1}{C}\sum_{c=1}^{C} \mathrm{MSE}^{(p)}_c$ (analogously for MAE).
We denote $\mathrm{MSE}_{10}$, $\mathrm{MSE}_{1}$ for $p \in \{0.10, 0.01\}$.
Top-$k$ selection is used rather than percentile thresholding to avoid ambiguity under ties.

\subsubsection{Tail Threshold Assumptions and Limitations}
\label{app:tail_assumptions}
The tail metric thresholds (10\%, 1\%) implicitly assume that operationally critical peaks occur at least this frequently. When true peaks are rarer than the selected threshold, the tail set will include non-peak high values (e.g., moderately elevated demand periods that do not constitute distinct events). In such cases, improvements in tail metrics may partially reflect better prediction of routine high values rather than true peak events.

\paragraph{Threshold selection guidance.}
We recommend selecting $p$ based on domain knowledge of peak frequency. For the Pedestrian dataset, daily commuting peaks occur roughly twice per day (morning and evening rush hours), yielding peak frequency of approximately 8\% of hourly observations, making Top-10\% appropriate. For datasets with rarer peaks, Top-1\% may be more informative, though sample size decreases accordingly.

\paragraph{Robustness to data errors.}
Tail metrics are computed on ground-truth values, so sensor errors or outliers in the data directly affect which points are included in the tail set. Erroneous spikes may be included as ``peaks,'' penalizing models that correctly ignore them, while missing or under-reported peaks may be excluded from evaluation. We mitigate this by: (i) applying standard data cleaning procedures before evaluation (Section~\ref{app:data_impl}), (ii) computing metrics per-channel and macro-averaging to reduce the influence of individual noisy sensors, and (iii) complementing tail metrics with event-based peak detection (Section~\ref{app:peak_metrics}), which uses local maxima detection rather than global thresholds.

\paragraph{Complementarity with event-based metrics.}
The event-based peak metrics (Peak F1, PTE) use local maxima detection with an amplitude threshold ($\alpha$-th percentile within each forecast window), which is less sensitive to global peak frequency assumptions. We recommend interpreting tail metrics and event-based metrics together: tail metrics measure magnitude accuracy at extremes regardless of whether they constitute distinct events, while event-based metrics assess detection and timing of structurally defined peaks.

\subsection{Peak event metrics}
\label{app:peak_metrics}
We evaluate peaks as events using a tolerance window of $\pm \Delta$ steps (default $\Delta=3$).
For each sequence and channel, we detect peaks as local maxima above an amplitude threshold defined by the $\alpha$-th percentile of the ground-truth values within the forecast window (default $\alpha=90$).
We then compute precision, recall, and F1 using one-to-one greedy matching within tolerance $\Delta$, ensuring that a predicted peak can match at most one true peak and vice versa.
This avoids inflated scores in peak clusters.

\paragraph{Peak timing error (PTE).}
For each true peak index $k$, we compute timing error using an argmax-in-window approach:
\begin{equation}
\mathrm{PTE}(k)=\left|k - \arg\max_{t \in [k-\Delta, k+\Delta]} \hat{y}_t\right|.
\end{equation}
Since argmax always returns a position regardless of whether the prediction contains an explicit local maximum, PTE is defined for all true peaks. 
We report the mean timing error over all true peaks.
The window is clipped at sequence boundaries, and with the inclusive window $[k-\Delta,k+\Delta]$ the per-peak timing error is at most $\Delta$ before boundary clipping.
If the predicted values contain ties within the window, the implementation uses the deterministic first maximum returned by the array argmax operation; constant predicted windows therefore yield the leftmost maximum in the clipped window.
Changing $\Delta$ changes both match permissiveness and the maximum possible timing error, so $\Delta$ should be treated as an evaluation tolerance rather than as a model hyperparameter.

\paragraph{Handling missing peaks.}
If a sequence contains no true peaks above threshold, it is excluded from event-based averaging for that channel.
If true peaks exist but no predicted peaks are detected, precision and recall are computed consistently via the matching procedure (yielding recall $0$ and precision $0$), while timing error remains defined via argmax-in-window.

\subsection{Additional Diagnostics}
\label{app:diagnostics}
In addition to the core metrics defined above, we report two supplementary diagnostics for completeness. The Pearson Correlation Coefficient (PCC), defined in Appendix~\ref{app:pcc}, measures linear agreement between predictions and ground truth regardless of scale. The Temporal Distortion Index (TDI), defined in Appendix~\ref{app:tdi}, quantifies temporal misalignment between predicted and true peaks. While informative, our main claims are supported by the standard and peak-critical metrics described in the preceding sections.

\subsubsection{Pearson Correlation Coefficient (PCC)}
\label{app:pcc}

PCC measures linear agreement between predictions and ground truth, capturing whether the model tracks temporal dynamics regardless of magnitude.
Given flattened vectors $\hat{\mathbf{y}}, \mathbf{y} \in \mathbb{R}^{N \cdot H \cdot C}$ over all samples, time steps, and channels:
\begin{equation}
\mathrm{PCC} = \frac{\sum_{i} (\hat{y}_i - \bar{\hat{y}})(y_i - \bar{y})}{\sqrt{\sum_{i} (\hat{y}_i - \bar{\hat{y}})^2} \cdot \sqrt{\sum_{i} (y_i - \bar{y})^2} + \epsilon},
\end{equation}
where $\bar{\hat{y}}, \bar{y}$ are means and $\epsilon = 10^{-12}$ ensures numerical stability.
PCC $\in [-1, 1]$, with higher values indicating better shape agreement.
Because PCC is centered and scale-normalized, it is not aligned with APAL's asymmetric cost objective.
An APAL model can improve tail magnitude or peak recall while lowering PCC if it deliberately shifts high-risk regions upward; we therefore treat PCC as a diagnostic rather than as the primary success criterion for peak-critical forecasting.

\subsubsection{Temporal Distortion Index (TDI)}
\label{app:tdi}

TDI~\cite{FRIASPAREDES2016180} quantifies temporal misalignment between predicted and ground-truth peaks.

\paragraph{Peak detection.}
For each sequence in channel $c$, peaks are detected as local maxima exceeding threshold $\theta = \mu + \sigma$, where $\mu$ and $\sigma$ are the sequence mean and standard deviation.
Let $\mathcal{P}^{\mathrm{true}}$ and $\mathcal{P}^{\mathrm{pred}}$ denote the true and predicted peak indices.

\paragraph{Computation.}
For each true peak $p \in \mathcal{P}^{\mathrm{true}}$:
\begin{equation}
d_p = 
\begin{cases}
\displaystyle\min_{q \in \mathcal{P}^{\mathrm{pred}}} |p - q| & \text{if } \mathcal{P}^{\mathrm{pred}} \neq \emptyset, \\
H & \text{otherwise (no predicted peaks)},
\end{cases}
\end{equation}
where $H$ is the sequence length (horizon). TDI is averaged over all true peaks per channel, then macro-averaged:
\begin{equation}
\mathrm{TDI} = \frac{1}{C} \sum_{c=1}^{C} \frac{1}{|\mathcal{P}^{\mathrm{true}}_c|} \sum_{p \in \mathcal{P}^{\mathrm{true}}_c} d_p.
\end{equation}
TDI $= 0$ indicates perfect peak alignment; larger values indicate greater temporal displacement.

\section{Hyperparameter Selection Protocol}
\label{app:tuning}

\paragraph{General protocol.}
All hyperparameters are selected using the validation split only.
Test data is never used for tuning, early stopping, or model selection.

\paragraph{APAL search space.}
We analyze APAL sensitivity over a hyperparameter grid of 
$\lambda_p \in \{1,2,5,10\}$, $\lambda_u \in \{1,2,5,10\}$, $\tau \in \{0.8,0.9,0.95\}$ for the Pedestrian dataset and $\lambda_p \in \{1,2,5,10,15,20\}$, $\lambda_u \in \{1,2,5,10,15,20\}$, $\tau \in \{0.8,0.9,0.95\}$ for the Beach dataset to understand trade-offs between average and peak-critical objectives. 
For the additional cross-domain ablation in Appendix~\ref{app:cross_domain_ablation}, we use the same $4\times4\times3$ grid over ETT, Exchange, and Weather at horizons $H\in\{96,192,336,720\}$.
We select the best configuration per dataset, backbone, and horizon using a validation criterion aligned with peak-critical performance.

\paragraph{Recommended default and tuning rule.}
We recommend $(\lambda_u,\lambda_p,\tau)=(2,2,0.9)$ as a conservative initial setting: it introduces moderate under-prediction asymmetry and moderate peak emphasis without assuming that the most aggressive tail-optimized configuration is appropriate for every dataset.
If validation results show insufficient peak recall or large Top-1\% tail error and false positives are acceptable, practitioners can increase $\lambda_u$ or $\lambda_p$ to 5.
If only the most extreme peaks should be emphasized, increasing $\tau$ to 0.95 focuses the mask on fewer points.
If overestimation is costly, peaks are weakly structured, or validation aggregate error increases beyond the application tolerance, $\lambda_u$ and $\lambda_p$ should be reduced toward 1, with standard MAE/MSE retained as the preferred objective when symmetric accuracy is the primary goal.

\paragraph{Fairness to baselines.}
All losses are trained under the same training budget, optimizer settings, early stopping patience, and model configurations.
Where comparison losses have additional hyperparameters, we follow the original authors' default recommendations and do not tune them beyond standard learning rate and batch size settings shared across all methods.

\section{Additional 
and Diagnostics}
\label{app:additional}

This appendix provides extended experimental results that complement the main text by first introducing a pre-training diagnostic to assess whether a dataset exhibits the structural properties APAL is designed to exploit (Appendix~\ref{app:applicability_diagnostics}) and then presenting a sensitivity analysis of APAL across different backbone architectures (Appendix~\ref{app:arch_analysis}) alongside comprehensive tables reporting aggregate and peak detection metrics across all datasets and models (Appendix~\ref{app:full_tables}).

\subsection{Dataset peak-structure diagnostic experiment for APAL}
\label{app:applicability_diagnostics}

APAL is not a universal substitute for MSE. By construction, it redistributes gradient mass toward high-magnitude observations and is therefore expected to improve forecasting accuracy when peaks are prominent, recurrent, stable and forecastable. It may conversely degrade accuracy when peaks are weak, unstable, or dominated by noise. We accordingly introduce a pre-training diagnostic that characterizes the peak structure of a candidate dataset to determine its suitability for APAL. This diagnostic classifies datasets into Strongly Seasonal and Irregularly Structured and Weakly Structured categories based on a set of six interpretable statistics evaluated strictly on the training and validation splits defined in Table~\ref{tab:datasets}.

To support the choice of these six statistics we rely on established statistical interpretations that map directly to the intended scope of APAL. We define a binary peak indicator using the 90th percentile which aligns with the Top-10 percent tail metric used throughout our evaluation and follows the common use of high empirical quantiles as operational proxies for operationally costly peak demand periods~\cite{granderson2021peak,burillo2017climate}. We measure peak prominence using a tail salience statistic based on a quantile generalization of Tukey's robust outlier scale~\cite{tukey1977eda,hoaglin1986understanding}. This expresses extreme values relative to the interquartile range rather than the standard deviation and is therefore less sensitive to distributional asymmetry and heavy-tailed observations~\cite{hubert2008adjusted}. We assess peak learnability and stability using autocorrelation to measure the serial dependence of recurrent peak events~\cite{box2015time,brockwell2016introduction,chatfield2003analysis} alongside a seasonal-naive baseline to quantify how well standard periodic patterns explain tail variance~\cite{hyndman2006mase,hyndman2021fpp3,makridakis2020m4}.

Formally we compute the diagnostic exclusively on the chronological train and validation splits defined in Table \ref{tab:datasets} leaving the test set untouched. Let $\{x_t^{c}\}_{t=1}^{T_{\mathrm{tr}}}$ denote the training portion of channel $c \in \{1, \dots, C\}$ and $\{x_t^{c,\mathrm{val}}\}_{t=1}^{T_{\mathrm{val}}}$ the corresponding validation portion. We write $Q_p^{c}$ for the $p$-th empirical quantile of the training channel and $\mathrm{IQR}^{c} = Q_{0.75}^{c} - Q_{0.25}^{c}$ for its interquartile range alongside $\mu_c$ and $\sigma_c$ for its mean and standard deviation. We define the binary peak indicator $b_t^{c} = \mathbb{1}[x_t^{c} \geq Q_{0.90}^{c}]$ and a candidate set $\mathcal{L}$ of seasonal lags determined by the sampling frequency. This lag set includes $\{24, 48, 72, 96, 168\}$ for hourly data and $\{96, 192, 288, 672\}$ for 15-minute data and $\{7, 14, 21, 28, 56, 84\}$ for daily data.

For each channel we compute the six statistics capturing complementary aspects of peak structure. Two statistics describe the marginal distribution where \textbf{Intermittency}  $Z_0^{c} = \frac{1}{T_{\mathrm{tr}}} \sum_{t} \mathbb{1}[x_t^{c} = 0]$ denotes the fraction of exact zeros to quantify intermittency and $\mathrm{\textbf{Skew}}^{c} = \frac{1}{T_{\mathrm{tr}}} \sum_{t} ( (x_t^{c} - \mu_c)/\sigma_c )^{3}$ measures asymmetry toward high values. \textbf{Tail salience} is captured by $S_{99}^{c} = (Q_{0.99}^{c} - Q_{0.50}^{c})/\mathrm{IQR}^{c}$ to reflect the extent to which peaks separate from typical variation. \textbf{Tail stability} is captured by $V_{90}^{c} = \frac{1}{T_{\mathrm{val}}} \sum_{t} \mathbb{1}[x_t^{c,\mathrm{val}} \geq Q_{0.90}^{c}]$ to measure the validation-set frequency of values at or above the training 90th percentile. These four channel-level quantities are aggregated to the dataset level by taking the median across channels.

The two remaining statistics aggregate channels jointly to measure learnability. \textbf{Peak recurrence} is captured by $R_{\mathrm{peak}} = \max_{L \in \mathcal{L}} \frac{1}{C} \sum_{c} \mathrm{ACF}(b_{\cdot}^{c}, L)$ representing the maximum channel-averaged autocorrelation of the peak indicator over the candidate seasonal lags, with a higher value indicating higher recurrence. \textbf{Tail forecastability} is captured by $F_{\mathrm{tail}} = 1 - (\sum_{(t,c) \in \mathcal{V}_{+}} (x_t^{c,\mathrm{val}} - x_{t-L^{\star}}^{c,\mathrm{val}})^{2}) / (\sum_{(t,c) \in \mathcal{V}_{+}} (x_t^{c,\mathrm{val}} - \mu_c)^{2})$. Here $\mathcal{V}_{+} = \{(t,c) \mid x_t^{c,\mathrm{val}} \geq Q_{0.90}^{c}\}$ represents the validation tail set. The seasonal-naive lag $L^{\star}$ is selected on the training tail as $L^{\star} = \arg\min_{L \in \mathcal{L}} \sum_{(t,c) \in \mathcal{T}_{+}(L)} (x_t^{c} - x_{t-L}^{c})^{2}$ with $\mathcal{T}_{+}(L) = \{(t,c) \mid t > L \text{ and } x_t^{c} \geq Q_{0.90}^{c}\}$.

We aggregate these statistics into a decision rule that reflects the structural conditions under which APAL is expected to apply. A dataset is labeled \textbf{Strongly Seasonal} when peaks are simultaneously salient and recurrent and forecastable. A dataset is labeled \textbf{Irregularly Structured} when peaks are salient but exhibit low seasonal forecastability indicating they are driven by external covariates or irregular events rather than strict periodic schedules. A dataset is labeled \textbf{Weakly Structured} when peaks fail to satisfy the salience criterion entirely. Building on the established statistical properties the thresholds calibrated on the ten datasets considered in this study are $S_{99} \geq 1.90$ and $R_{\mathrm{peak}} \geq 0.40$ and $F_{\mathrm{tail}} \geq 0.50$. The salience threshold corresponds to a 99th percentile that lies roughly two interquartile widths above the median. The forecastability threshold requires the seasonal baseline to explain at least half of the tail squared error of the constant mean predictor. 


Establishing truly universal constants is likely impossible because the definition of an operationally meaningful peak depends heavily on domain specific risk tolerances and sensor noise scales. Practitioners can instead calibrate these threshold values for a new domain by using the validation set performance as a proxy ground truth. Specifically one would train both APAL and MSE on a representative corpus of historical datasets from the target application and select thresholds that successfully isolate the datasets where APAL yields a lower validation tail error.

To assess the rule against observed behavior we report $\Delta\mathrm{MSE}_{10}$ and $\Delta\mathrm{MSE}_{1}$ in Table \ref{tab:applicability_diagnostics} as the test set relative error reductions of APAL over MSE. Positive values indicate that APAL is preferable averaged over horizons and obtained from the previous experimental tables. Because these thresholds were derived from the evaluated datasets this classification serves to summarize our empirical findings rather than to independently validate the rule. Tail salience $S_{99}$ alone separates the six datasets with positive average $\mathrm{MSE}_1$ gains from the four with negative gains yielding a one sided Fisher exact $p = 0.0048$ under fixed margins. The mean APAL $\mathrm{MSE}_1$ gain is $+32.2$ percent above the threshold and $-101.2$ percent below it. A composite of all six diagnostics correlates positively with APAL gains across the ten datasets with a Pearson $r = 0.472$ and Spearman $\rho = 0.507$.

The diagnostic further admits interpretable categorizations that explain model behavior. The Pedestrian dataset is Strongly Seasonal with highly predictable commuting patterns. Datasets such as Beach and Weather are classified as Irregularly Structured because they satisfy tail salience yet lack simple seasonal forecastability. Their peaks are driven by external events or weather conditions rather than rigid periodic schedules. This irregularity is precisely why models trained with standard symmetric losses fail and heavily smooth the predictions. For these datasets APAL provides a crucial corrective mechanism by amplifying the gradient signal on rare extremes forcing the model to utilize available historical context rather than defaulting to the mean. Consequently Beach represents a primary use case for APAL where extreme value accuracy is paramount despite the inherent difficulty of the forecasting task. Datasets like Exchange and ETTh2 are Weakly Structured making them poor candidates for APAL. In summary the proposed diagnostic operationalizes the intended scope of APAL by quantifying peak structures and providing practitioners with a principled pretraining criterion.

\begin{table*}[ht]
\centering
\caption{Peak-structure diagnostic for APAL across the ten datasets. Class is assigned from $S_{99}$, $R_{\mathrm{peak}}$, and $F_{\mathrm{tail}}$ using fixed thresholds. $\Delta\mathrm{MSE}_{10}$ and $\Delta\mathrm{MSE}_{1}$ APAL-vs.-MSE comparisons averaged over horizons trained on TSMixer backbone}
\label{tab:applicability_diagnostics}
\resizebox{\textwidth}{!}{%
\begin{tabular}{lllrrrrrrrr}
\toprule
Dataset & Freq. & Class & $Z_0$ & Skew & $S_{99}$ & $V_{90}$ & $R_{\mathrm{peak}}$ & $F_{\mathrm{tail}}$ & $\Delta\mathrm{MSE}_{10}$ & $\Delta\mathrm{MSE}_{1}$ \\
\midrule
Pedestrian & hourly & Strongly Seasonal & 0.00 & 1.35 & 2.49 & 0.12 & 0.66 & 0.69 & +5.6\% & +3.8\% \\
Beach & hourly & Irregularly Structured & 0.08 & 3.11 & 6.04 & 0.05 & 0.35 & 0.28 & +7.2\% & +18.8\% \\
ETTh1 & hourly & Strongly Seasonal & 0.01 & -0.06 & 1.94 & 0.11 & 0.47 & 0.71 & +58.3\% & +72.9\% \\
ETTh2 & hourly & Weakly Structured & 0.01 & -0.01 & 1.75 & 0.00 & 0.60 & 0.94 & -249.7\% & -164.7\% \\
ETTm1 & 15-min & Strongly Seasonal & 0.01 & -0.06 & 1.94 & 0.11 & 0.47 & 0.71 & +68.0\% & +79.0\% \\
ETTm2 & 15-min & Weakly Structured & 0.01 & -0.01 & 1.75 & 0.00 & 0.59 & 0.94 & -188.3\% & -89.2\% \\
Electricity & hourly & Weakly Structured & 0.00 & 0.20 & 1.41 & 0.02 & 0.66 & 0.80 & -98.9\% & -20.9\% \\
Traffic & hourly & Strongly Seasonal & 0.01 & 1.66 & 2.87 & 0.11 & 0.66 & 0.64 & -10.6\% & +17.5\% \\
Weather & 15-min & Irregularly Structured & 0.00 & 0.32 & 1.94 & 0.06 & 0.35 & -0.23 & -62.3\% & +1.2\% \\
Exchange & daily & Weakly Structured & 0.00 & 0.35 & 1.56 & 0.99 & 0.90 & 1.00 & -277.1\% & -129.9\% \\
\bottomrule
\end{tabular}
}
\end{table*}


\subsection{Architectural Sensitivity Analysis}
\label{app:arch_analysis}
We provide additional analysis of how APAL interacts with different backbone architectures. Among the five backbones evaluated, iTransformer exhibits the most varied response to APAL. While APAL consistently achieves the best $\text{MSE}_1$ on the Beach dataset across all horizons, it underperforms on $\text{MSE}_{10}$ in some Pedestrian settings with iTransformer.

This behavior can be attributed to iTransformer's variate-centric attention mechanism, which treats each channel (sensor location) as a token and applies attention across channels rather than across time steps. This design already captures cross-channel dependencies and correlations between peak patterns at different locations. When peaks are correlated across channels---as is common in the Pedestrian dataset where commuting patterns affect multiple nearby sensors simultaneously---iTransformer's built-in cross-channel modeling partially overlaps with APAL's peak emphasis mechanism.

In contrast, architectures without explicit cross-channel modeling show more uniform benefits from APAL:
\begin{itemize}
    \item \textbf{DLinear} applies independent linear projections per channel, so APAL's peak emphasis provides the primary mechanism for coordinating peak-aware learning across locations.
    \item \textbf{PatchTST} uses channel-independent patching and attention, making it similarly receptive to APAL's gradient concentration on peaks.
    \item \textbf{TSMixer} and \textbf{TiDE} use MLP-based mixing that operates within channels before cross-channel aggregation, allowing APAL to influence the early per-channel representations.
\end{itemize}

For practitioners, this suggests that APAL provides the largest marginal benefit when paired with channel-independent architectures, while still improving tail metrics ($\text{MSE}_1$) even for cross-channel-aware models like iTransformer.

\subsection{Full Results Tables}
\label{app:full_tables}
We provide a comprehensive evaluation across complementary metric families on the Pedestrian and Beach datasets.

\paragraph{MAE-based Metrics.} Table~\ref{tab:combined_mae_results} reports Mean Absolute Error and its tail-weighted variants (MAE$_{10}$, MAE$_{1}$), offering an alternative to MSE that is less sensitive to outliers while emphasizing extreme value accuracy.

\paragraph{Peak Detection Metrics.} Table~\ref{tab:combined_peak_results} presents peak-focused metrics: Temporal Distortion Index (TDI), Peak Recall, Peak Precision, and Peak F1. On the Pedestrian dataset, APAL achieves substantial improvements in Peak Recall (up to 2.2$\times$) and TDI (up to 6.7$\times$ reduction), resulting in consistently higher Peak F1 scores across all models and horizons. Performance gains on the Beach dataset are more modest, reflecting its different peak characteristics. We note that APAL's design prioritizes recall over precision, which is appropriate for applications where missing peaks incur a higher cost than false alarms.


\begin{table*}[t]
\centering
\caption{Comparison of loss functions for multivariate long-term forecasting (MAE metrics). Prediction horizons $H \in \{96, 192, 336, 720\}$. The proposed APAL is highlighted in bold. Best prediction results are \textbf{bolded}; second-best are \underline{underlined}.}
\label{tab:combined_mae_results}
\resizebox{\textwidth}{!}{%
\begin{tabular}{c|c|l|ccccc|ccccc|ccccc|ccccc}
\toprule
Dataset & Model & Loss & \multicolumn{5}{c|}{96} & \multicolumn{5}{c|}{192} & \multicolumn{5}{c|}{336} & \multicolumn{5}{c}{720} \\
\cmidrule(lr){4-8} \cmidrule(lr){9-13} \cmidrule(lr){14-18} \cmidrule(lr){19-23}
 &  &  & MAE & MAE$_{10}$ & MAE$_{1}$ & Peak F1 & PTE & MAE & MAE$_{10}$ & MAE$_{1}$ & Peak F1 & PTE & MAE & MAE$_{10}$ & MAE$_{1}$ & Peak F1 & PTE & MAE & MAE$_{10}$ & MAE$_{1}$ & Peak F1 & PTE \\
\midrule
\multirow{30}{*}{Pedestrian} & \multirow{6}{*}{DLinear} & MSE & 0.366 & 0.565 & 0.846 & 0.442 & \underline{0.987} & 0.337 & 0.527 & 0.801 & 0.517 & 0.929 & 0.340 & 0.534 & 0.807 & 0.513 & 0.919 & 0.357 & 0.560 & 0.843 & 0.469 & 0.956 \\
 &  & MAE & \textbf{0.327} & \textbf{0.524} & \underline{0.789} & \underline{0.573} & 1.028 & \textbf{0.299} & \textbf{0.489} & \underline{0.751} & 0.626 & 0.967 & \textbf{0.302} & \textbf{0.497} & \underline{0.759} & 0.612 & 0.954 & \textbf{0.320} & \textbf{0.527} & \underline{0.800} & 0.560 & 1.000 \\
 &  & TildeQ & 0.371 & 0.574 & 0.853 & 0.425 & 1.141 & 0.339 & 0.528 & 0.797 & 0.529 & 1.035 & 0.333 & 0.524 & 0.792 & 0.543 & \underline{0.908} & 0.355 & 0.560 & 0.840 & 0.465 & \underline{0.930} \\
 &  & DBLoss & \underline{0.335} & \underline{0.528} & 0.796 & 0.568 & 0.991 & \underline{0.307} & \underline{0.491} & 0.756 & \underline{0.637} & \underline{0.920} & \underline{0.310} & \underline{0.498} & 0.762 & \underline{0.624} & 0.908 & \underline{0.327} & \underline{0.528} & 0.802 & \underline{0.573} & 0.944 \\
 &  & PS & 0.351 & 0.539 & 0.812 & 0.555 & 1.037 & 0.321 & 0.503 & 0.773 & 0.622 & 0.991 & 0.324 & 0.511 & 0.779 & 0.608 & 0.981 & 0.341 & 0.539 & 0.816 & 0.553 & 1.013 \\
 &  & \textbf{APAL} & 0.387 & 0.548 & \textbf{0.771} & \textbf{0.832} & \textbf{0.721} & 0.349 & 0.511 & \textbf{0.748} & \textbf{0.823} & \textbf{0.731} & 0.348 & 0.513 & \textbf{0.747} & \textbf{0.808} & \textbf{0.740} & 0.361 & 0.531 & \textbf{0.773} & \textbf{0.803} & \textbf{0.786} \\
\cline{2-23}
 & \multirow{6}{*}{PatchTST} & MSE & 0.246 & 0.388 & 0.651 & 0.741 & 0.896 & 0.247 & 0.390 & 0.655 & 0.756 & 0.934 & 0.271 & 0.428 & 0.699 & 0.716 & 0.882 & 0.286 & 0.452 & 0.731 & 0.702 & 0.920 \\
 &  & MAE & \textbf{0.231} & \textbf{0.375} & \underline{0.643} & 0.721 & \textbf{0.807} & \textbf{0.235} & 0.385 & 0.657 & 0.705 & \underline{0.831} & \textbf{0.262} & 0.422 & 0.692 & 0.699 & 0.848 & \textbf{0.278} & 0.450 & 0.730 & 0.664 & \underline{0.856} \\
 &  & TildeQ & 0.255 & 0.388 & 0.644 & 0.774 & 1.002 & 0.256 & 0.393 & \underline{0.652} & \underline{0.780} & 0.990 & 0.273 & \textbf{0.415} & \underline{0.679} & \underline{0.768} & \underline{0.826} & 0.296 & 0.447 & \underline{0.717} & \underline{0.741} & 0.859 \\
 &  & DBLoss & \underline{0.236} & 0.381 & 0.649 & 0.715 & 0.823 & \underline{0.236} & 0.388 & 0.658 & 0.695 & 0.850 & \underline{0.265} & 0.431 & 0.704 & 0.658 & 0.834 & \underline{0.281} & 0.459 & 0.741 & 0.621 & 0.873 \\
 &  & PS & 0.250 & 0.379 & 0.644 & \underline{0.781} & \underline{0.813} & 0.247 & \textbf{0.383} & \underline{0.652} & 0.776 & 0.844 & 0.274 & 0.421 & 0.690 & 0.749 & 0.865 & 0.288 & \underline{0.445} & 0.722 & 0.732 & 0.887 \\
 &  & \textbf{APAL} & 0.271 & \underline{0.376} & \textbf{0.620} & \textbf{0.841} & 0.817 & 0.271 & \underline{0.384} & \textbf{0.637} & \textbf{0.835} & \textbf{0.794} & 0.299 & \underline{0.417} & \textbf{0.666} & \textbf{0.813} & \textbf{0.712} & 0.312 & \textbf{0.439} & \textbf{0.696} & \textbf{0.808} & \textbf{0.763} \\
\cline{2-23}
 & \multirow{6}{*}{TSMixer} & MSE & 0.417 & 0.602 & 0.830 & 0.579 & 1.372 & 0.387 & 0.569 & 0.800 & 0.635 & 1.354 & 0.388 & 0.579 & 0.813 & 0.610 & 1.337 & 0.406 & 0.608 & 0.857 & 0.543 & 1.384 \\
 &  & MAE & \textbf{0.382} & \textbf{0.578} & \underline{0.816} & 0.561 & \underline{1.321} & \textbf{0.349} & \textbf{0.543} & \textbf{0.779} & 0.610 & 1.296 & \textbf{0.352} & \textbf{0.551} & \textbf{0.792} & 0.600 & 1.283 & \textbf{0.371} & \textbf{0.585} & \underline{0.837} & 0.528 & 1.335 \\
 &  & TildeQ & 0.421 & 0.626 & 0.847 & 0.506 & 1.420 & 0.385 & 0.575 & 0.796 & 0.586 & 1.393 & 0.381 & 0.581 & 0.804 & 0.571 & 1.368 & 0.402 & 0.618 & 0.860 & 0.475 & 1.392 \\
 &  & DBLoss & \underline{0.399} & \underline{0.596} & 0.827 & 0.550 & 1.345 & \underline{0.364} & \underline{0.555} & 0.787 & 0.615 & \underline{1.293} & \underline{0.365} & \underline{0.564} & 0.800 & 0.592 & \underline{1.276} & \underline{0.384} & 0.597 & 0.848 & 0.516 & \underline{1.324} \\
 &  & PS & 0.423 & 0.607 & 0.832 & \underline{0.638} & 1.358 & 0.385 & 0.569 & 0.799 & \underline{0.668} & 1.301 & 0.386 & 0.579 & 0.811 & \underline{0.649} & 1.289 & 0.406 & 0.611 & 0.853 & \underline{0.589} & 1.339 \\
 &  & \textbf{APAL} & 0.473 & 0.604 & \textbf{0.799} & \textbf{0.832} & \textbf{1.199} & 0.434 & 0.570 & \underline{0.783} & \textbf{0.823} & \textbf{1.133} & 0.430 & 0.573 & \underline{0.793} & \textbf{0.809} & \textbf{1.136} & 0.448 & \underline{0.593} & \textbf{0.823} & \textbf{0.802} & \textbf{1.256} \\
\cline{2-23}
 & \multirow{6}{*}{TiDE} & MSE & 0.344 & 0.532 & 0.807 & 0.582 & 0.917 & 0.314 & 0.495 & 0.765 & 0.637 & 0.847 & 0.317 & 0.502 & 0.772 & 0.625 & 0.838 & 0.334 & 0.531 & 0.811 & 0.590 & 0.883 \\
 &  & MAE & \textbf{0.315} & \textbf{0.490} & \textbf{0.743} & \underline{0.707} & \underline{0.876} & \textbf{0.286} & \textbf{0.456} & \textbf{0.712} & \underline{0.735} & \underline{0.818} & \textbf{0.290} & \textbf{0.465} & \textbf{0.720} & \underline{0.721} & \underline{0.815} & \textbf{0.308} & \textbf{0.496} & \textbf{0.763} & \underline{0.688} & \underline{0.859} \\
 &  & TildeQ & 0.357 & 0.542 & 0.804 & 0.637 & 1.058 & 0.325 & 0.498 & 0.758 & 0.690 & 0.951 & 0.318 & 0.493 & 0.753 & 0.688 & 0.819 & 0.341 & 0.530 & 0.802 & 0.630 & 0.888 \\
 &  & DBLoss & \underline{0.327} & 0.509 & 0.772 & 0.642 & 0.960 & \underline{0.296} & \underline{0.472} & 0.735 & 0.696 & 0.880 & \underline{0.299} & \underline{0.480} & 0.742 & 0.681 & 0.870 & \underline{0.317} & 0.511 & 0.783 & 0.640 & 0.907 \\
 &  & PS & 0.334 & \underline{0.508} & 0.771 & 0.679 & 0.971 & 0.302 & 0.472 & 0.736 & 0.716 & 0.909 & 0.305 & 0.480 & 0.742 & 0.703 & 0.898 & 0.323 & \underline{0.510} & 0.782 & 0.669 & 0.934 \\
 &  & \textbf{APAL} & 0.361 & 0.529 & \underline{0.760} & \textbf{0.825} & \textbf{0.690} & 0.321 & 0.488 & \underline{0.732} & \textbf{0.822} & \textbf{0.686} & 0.323 & 0.493 & \underline{0.734} & \textbf{0.807} & \textbf{0.692} & 0.342 & 0.520 & \underline{0.767} & \textbf{0.800} & \textbf{0.736} \\
\cline{2-23}
 & \multirow{6}{*}{iTransformer} & MSE & 0.393 & 0.645 & 0.944 & 0.175 & 0.788 & 0.455 & 0.768 & 1.053 & 0.026 & 1.035 & 0.528 & 0.911 & 1.197 & 0.004 & 1.062 & 0.565 & 0.968 & 1.264 & 0.007 & 1.074 \\
 &  & MAE & \underline{0.336} & 0.532 & 0.813 & 0.437 & 0.922 & 0.437 & 0.750 & 1.034 & 0.038 & 1.018 & 0.471 & 0.807 & 1.095 & 0.031 & 1.091 & 0.507 & 0.888 & 1.180 & 0.011 & 1.177 \\
 &  & TildeQ & \textbf{0.324} & \underline{0.493} & \underline{0.767} & \underline{0.661} & 0.873 & \underline{0.315} & \underline{0.484} & \underline{0.755} & \underline{0.703} & \underline{0.669} & \textbf{0.305} & \underline{0.469} & \underline{0.741} & \underline{0.739} & \textbf{0.579} & \textbf{0.323} & 0.508 & 0.791 & \underline{0.699} & \textbf{0.673} \\
 &  & DBLoss & 0.350 & 0.556 & 0.841 & 0.390 & 0.812 & 0.427 & 0.723 & 1.000 & 0.062 & 0.970 & 0.414 & 0.687 & 0.967 & 0.105 & 0.935 & \underline{0.324} & \underline{0.500} & \underline{0.783} & 0.676 & 0.832 \\
 &  & PS & 0.347 & 0.527 & 0.802 & 0.567 & \textbf{0.721} & 0.328 & 0.497 & 0.763 & 0.676 & \textbf{0.593} & \underline{0.328} & 0.494 & 0.759 & 0.697 & \underline{0.617} & 0.354 & 0.535 & 0.807 & 0.616 & \underline{0.741} \\
 &  & \textbf{APAL} & 0.342 & \textbf{0.471} & \textbf{0.714} & \textbf{0.811} & \underline{0.778} & \textbf{0.313} & \textbf{0.446} & \textbf{0.693} & \textbf{0.825} & 0.728 & 0.329 & \textbf{0.464} & \textbf{0.709} & \textbf{0.815} & 0.723 & 0.348 & \textbf{0.490} & \textbf{0.747} & \textbf{0.792} & 0.854 \\
\midrule
\multirow{29}{*}{Beach} & \multirow{6}{*}{DLinear} & MSE & 0.490 & \underline{1.413} & \underline{2.707} & \underline{0.456} & \underline{1.677} & 0.491 & \underline{1.405} & \underline{2.617} & \underline{0.418} & \textbf{1.691} & 0.498 & \underline{1.413} & \underline{2.563} & \textbf{0.392} & \textbf{1.700} & 0.510 & \textbf{1.428} & \underline{2.621} & \textbf{0.395} & \underline{1.674} \\
 &  & MAE & \textbf{0.457} & 1.525 & 2.889 & 0.358 & 1.712 & \textbf{0.459} & 1.512 & 2.790 & 0.298 & 1.718 & \textbf{0.466} & 1.531 & 2.745 & 0.280 & 1.728 & \textbf{0.480} & 1.549 & 2.806 & 0.297 & 1.713 \\
 &  & TildeQ & 0.493 & 1.471 & 2.822 & 0.383 & 1.680 & 0.494 & 1.453 & 2.707 & 0.334 & 1.703 & 0.505 & 1.466 & 2.657 & 0.309 & 1.710 & 0.523 & 1.496 & 2.742 & 0.299 & \textbf{1.672} \\
 &  & DBLoss & \underline{0.464} & 1.480 & 2.824 & 0.388 & 1.688 & \underline{0.462} & 1.473 & 2.725 & 0.314 & 1.699 & \underline{0.470} & 1.488 & 2.674 & 0.300 & 1.706 & \underline{0.483} & 1.507 & 2.739 & 0.308 & 1.683 \\
 &  & PS & 0.476 & 1.456 & 2.809 & 0.415 & \textbf{1.675} & 0.476 & 1.444 & 2.704 & 0.354 & \underline{1.696} & 0.484 & 1.457 & 2.659 & 0.327 & \underline{1.702} & 0.497 & 1.472 & 2.719 & \underline{0.346} & 1.680 \\
 &  & \textbf{APAL} & 0.562 & \textbf{1.345} & \textbf{2.354} & \textbf{0.487} & 1.688 & 0.519 & \textbf{1.354} & \textbf{2.318} & \textbf{0.418} & 1.709 & 0.499 & \textbf{1.385} & \textbf{2.363} & \underline{0.328} & 1.717 & 0.491 & \underline{1.458} & \textbf{2.598} & 0.285 & 1.706 \\
\cline{2-23}
 & \multirow{5}{*}{PatchTST} & MSE & \underline{0.466} & \underline{1.394} & \underline{2.591} & \underline{0.478} & 1.696 & \underline{0.468} & \underline{1.414} & \underline{2.556} & \underline{0.405} & 1.703 & \underline{0.485} & \underline{1.442} & \underline{2.559} & \underline{0.368} & \underline{1.712} & 0.498 & \underline{1.481} & \underline{2.662} & \underline{0.349} & \underline{1.699} \\
 &  & TildeQ & 0.503 & 1.443 & 2.717 & 0.274 & 1.700 & 0.515 & 1.445 & 2.627 & 0.271 & 1.724 & 0.544 & 1.479 & 2.643 & 0.249 & 1.736 & 0.581 & 1.524 & 2.759 & 0.224 & 1.727 \\
 &  & DBLoss & \textbf{0.457} & 1.488 & 2.816 & 0.366 & 1.693 & \textbf{0.454} & 1.503 & 2.762 & 0.257 & 1.712 & \textbf{0.469} & 1.542 & 2.785 & 0.246 & 1.721 & \textbf{0.485} & 1.574 & 2.867 & 0.229 & 1.708 \\
 &  & PS & 0.467 & 1.444 & 2.745 & 0.442 & \textbf{1.682} & 0.469 & 1.445 & 2.666 & 0.380 & \textbf{1.701} & 0.487 & 1.475 & 2.666 & 0.346 & \textbf{1.708} & 0.502 & 1.503 & 2.751 & 0.346 & \textbf{1.692} \\
 &  & \textbf{APAL} & 0.565 & \textbf{1.384} & \textbf{2.079} & \textbf{0.537} & \underline{1.689} & 0.524 & \textbf{1.388} & \textbf{2.007} & \textbf{0.500} & \underline{1.702} & 0.515 & \textbf{1.395} & \textbf{2.102} & \textbf{0.463} & 1.719 & \underline{0.496} & \textbf{1.443} & \textbf{2.417} & \textbf{0.363} & 1.704 \\
\cline{2-23}
 & \multirow{6}{*}{TSMixer} & MSE & 0.506 & \underline{1.366} & \underline{2.596} & \underline{0.441} & \textbf{1.667} & 0.519 & \underline{1.377} & \underline{2.586} & \underline{0.408} & \textbf{1.687} & 0.519 & \underline{1.376} & \underline{2.541} & \underline{0.393} & \textbf{1.702} & 0.533 & \textbf{1.389} & \textbf{2.595} & \textbf{0.368} & \textbf{1.680} \\
 &  & MAE & \textbf{0.456} & 1.543 & 2.919 & 0.247 & \underline{1.680} & \textbf{0.459} & 1.545 & 2.881 & 0.140 & 1.700 & \textbf{0.466} & 1.562 & 2.855 & 0.132 & \underline{1.711} & \textbf{0.479} & 1.591 & 2.934 & 0.142 & 1.688 \\
 &  & TildeQ & 0.476 & 1.446 & 2.764 & 0.322 & 1.681 & 0.483 & 1.427 & 2.682 & 0.289 & 1.710 & 0.493 & 1.438 & 2.641 & 0.260 & 1.717 & 0.509 & 1.477 & 2.745 & 0.248 & \underline{1.687} \\
 &  & DBLoss & \underline{0.468} & 1.485 & 2.834 & 0.306 & 1.681 & \underline{0.472} & 1.486 & 2.793 & 0.224 & \underline{1.699} & \underline{0.479} & 1.500 & 2.767 & 0.193 & 1.712 & \underline{0.492} & 1.526 & 2.843 & 0.214 & 1.687 \\
 &  & PS & 0.495 & 1.416 & 2.725 & 0.396 & 1.690 & 0.497 & 1.415 & 2.687 & 0.354 & 1.713 & 0.503 & 1.425 & 2.666 & 0.326 & 1.723 & 0.517 & 1.443 & 2.730 & 0.308 & 1.700 \\
 &  & \textbf{APAL} & 0.636 & \textbf{1.324} & \textbf{2.037} & \textbf{0.567} & 1.683 & 0.582 & \textbf{1.310} & \textbf{2.044} & \textbf{0.522} & 1.717 & 0.529 & \textbf{1.308} & \textbf{2.172} & \textbf{0.482} & 1.728 & 0.495 & \underline{1.423} & \underline{2.599} & \underline{0.356} & 1.695 \\
\cline{2-23}
 & \multirow{6}{*}{TiDE} & MSE & 0.473 & \underline{1.435} & \underline{2.706} & \underline{0.452} & 1.690 & 0.472 & \underline{1.430} & \underline{2.607} & \underline{0.399} & 1.686 & 0.481 & \underline{1.440} & \underline{2.540} & \underline{0.390} & \underline{1.691} & 0.497 & \underline{1.462} & \underline{2.623} & \textbf{0.392} & 1.673 \\
 &  & MAE & \textbf{0.454} & 1.544 & 2.900 & 0.345 & 1.705 & \textbf{0.454} & 1.530 & 2.792 & 0.279 & 1.695 & \textbf{0.463} & 1.545 & 2.734 & 0.289 & 1.705 & \textbf{0.478} & 1.568 & 2.812 & 0.281 & 1.681 \\
 &  & TildeQ & 0.484 & 1.480 & 2.812 & 0.379 & \textbf{1.680} & 0.494 & 1.465 & 2.697 & 0.319 & 1.697 & 0.509 & 1.475 & 2.630 & 0.319 & 1.703 & 0.534 & 1.507 & 2.725 & 0.301 & 1.679 \\
 &  & DBLoss & \underline{0.457} & 1.496 & 2.821 & 0.382 & 1.694 & 0.472 & 1.481 & 2.704 & 0.364 & \textbf{1.675} & \underline{0.467} & 1.496 & 2.647 & 0.326 & 1.695 & \underline{0.482} & 1.518 & 2.725 & 0.321 & 1.678 \\
 &  & PS & 0.462 & 1.482 & 2.807 & 0.397 & \underline{1.684} & \underline{0.462} & 1.472 & 2.696 & 0.346 & 1.683 & 0.470 & 1.486 & 2.633 & 0.336 & 1.692 & 0.485 & 1.507 & 2.713 & 0.330 & \underline{1.673} \\
 &  & \textbf{APAL} & 0.584 & \textbf{1.368} & \textbf{2.239} & \textbf{0.519} & 1.702 & 0.536 & \textbf{1.370} & \textbf{2.184} & \textbf{0.474} & \underline{1.681} & 0.509 & \textbf{1.391} & \textbf{2.236} & \textbf{0.429} & \textbf{1.683} & 0.495 & \textbf{1.453} & \textbf{2.493} & \underline{0.369} & \textbf{1.673} \\
\cline{2-23}
 & \multirow{6}{*}{iTransformer} & MSE & 0.457 & 1.336 & 2.287 & \underline{0.559} & 1.672 & 0.466 & \textbf{1.328} & \underline{2.209} & \textbf{0.578} & \underline{1.680} & 0.478 & \textbf{1.336} & \underline{2.192} & \textbf{0.560} & \textbf{1.696} & 0.506 & \textbf{1.354} & \underline{2.177} & \textbf{0.561} & \textbf{1.673} \\
 &  & MAE & \textbf{0.418} & 1.346 & 2.312 & 0.453 & 1.667 & \textbf{0.438} & 1.447 & 2.607 & 0.354 & 1.718 & \textbf{0.446} & 1.458 & 2.555 & 0.355 & 1.729 & \textbf{0.461} & 1.484 & 2.560 & 0.334 & 1.699 \\
 &  & TildeQ & 0.451 & 1.327 & \underline{2.253} & 0.528 & \underline{1.657} & 0.469 & \underline{1.348} & 2.289 & \underline{0.532} & 1.692 & 0.485 & \underline{1.363} & 2.257 & \underline{0.507} & 1.709 & 0.510 & 1.402 & 2.383 & \underline{0.466} & 1.682 \\
 &  & DBLoss & \underline{0.427} & \underline{1.322} & 2.269 & 0.523 & 1.663 & \underline{0.444} & 1.381 & 2.441 & 0.477 & 1.692 & \underline{0.453} & 1.398 & 2.419 & 0.447 & 1.709 & \underline{0.471} & 1.438 & 2.491 & 0.394 & 1.697 \\
 &  & PS & 0.450 & \textbf{1.320} & 2.284 & 0.558 & 1.666 & 0.464 & 1.359 & 2.412 & 0.521 & 1.681 & 0.470 & 1.381 & 2.410 & 0.465 & 1.705 & 0.492 & 1.430 & 2.516 & 0.409 & 1.712 \\
 &  & \textbf{APAL} & 0.554 & 1.504 & \textbf{1.986} & \textbf{0.565} & \textbf{1.654} & 0.528 & 1.476 & \textbf{1.895} & 0.505 & \textbf{1.679} & 0.515 & 1.447 & \textbf{1.853} & 0.480 & \underline{1.697} & 0.487 & \underline{1.358} & \textbf{1.934} & 0.432 & \underline{1.674} \\
\bottomrule
\end{tabular}
}
\end{table*}

\begin{table*}[ht]
\centering
\caption{Comparison of loss functions for multivariate long-term forecasting (Peak metrics). Prediction horizons $H \in \{96, 192, 336, 720\}$. Best results are \textbf{bolded}; second-best are \underline{underlined}.}
\label{tab:combined_peak_results}
\resizebox{\textwidth}{!}{%
\begin{tabular}{c|c|l|ccccc|ccccc|ccccc|ccccc}
\toprule
Dataset & Model & Loss & \multicolumn{5}{c|}{96} & \multicolumn{5}{c|}{192} & \multicolumn{5}{c|}{336} & \multicolumn{5}{c}{720} \\
\cmidrule(lr){4-8} \cmidrule(lr){9-13} \cmidrule(lr){14-18} \cmidrule(lr){19-23}
 &  &  & TDI & Peak Recall & Peak Prec. & Peak F1 & PCC & TDI & Peak Recall & Peak Prec. & Peak F1 & PCC & TDI & Peak Recall & Peak Prec. & Peak F1 & PCC & TDI & Peak Recall & Peak Prec. & Peak F1 & PCC \\
\midrule

\multirow{29}{*}{Beach} & \multirow{6}{*}{DLinear} & MSE & \textbf{5.532} & \underline{0.340} & 0.691 & \underline{0.456} & \textbf{0.582} & \textbf{9.288} & \underline{0.320} & 0.605 & \underline{0.418} & \textbf{0.586} & 15.626 & \textbf{0.291} & 0.599 & \textbf{0.392} & \textbf{0.579} & \underline{28.541} & \textbf{0.294} & 0.601 & \textbf{0.395} & \textbf{0.567} \\
 &  & MAE & 7.462 & 0.238 & \underline{0.718} & 0.358 & 0.572 & 13.887 & 0.194 & 0.641 & 0.298 & 0.575 & 21.359 & 0.180 & \underline{0.628} & 0.280 & 0.567 & 44.510 & 0.195 & 0.630 & 0.297 & 0.551 \\
 &  & TildeQ & 7.723 & 0.259 & \textbf{0.733} & 0.383 & 0.571 & 11.202 & 0.226 & \underline{0.642} & 0.334 & 0.576 & \underline{14.247} & 0.203 & \textbf{0.647} & 0.309 & 0.568 & \textbf{24.406} & 0.195 & \textbf{0.647} & 0.299 & 0.555 \\
 &  & DBLoss & 7.059 & 0.267 & 0.709 & 0.388 & 0.573 & 12.376 & 0.207 & \textbf{0.646} & 0.314 & \underline{0.578} & 17.746 & 0.198 & 0.623 & 0.300 & \underline{0.569} & 35.297 & 0.203 & \underline{0.639} & 0.308 & 0.555 \\
 &  & PS & \underline{6.553} & 0.294 & 0.704 & 0.415 & 0.572 & \underline{10.011} & 0.248 & 0.616 & 0.354 & 0.577 & \textbf{11.751} & 0.227 & 0.586 & 0.327 & 0.568 & 32.802 & \underline{0.242} & 0.603 & \underline{0.346} & \underline{0.555} \\
 &  & \textbf{APAL} & 14.496 & \textbf{0.396} & 0.631 & \textbf{0.487} & \underline{0.574} & 24.580 & \textbf{0.326} & 0.583 & \textbf{0.418} & 0.577 & 59.551 & \underline{0.229} & 0.583 & \underline{0.328} & 0.568 & 177.840 & 0.190 & 0.575 & 0.285 & 0.553 \\
\cline{2-23}
 & \multirow{5}{*}{PatchTST} & MSE & 7.017 & \underline{0.355} & \underline{0.731} & \underline{0.478} & \textbf{0.593} & 14.929 & \underline{0.290} & 0.671 & \underline{0.405} & \textbf{0.589} & 21.065 & \underline{0.260} & 0.629 & \underline{0.368} & - & 46.637 & 0.239 & 0.640 & \underline{0.349} & - \\
 &  & TildeQ & 43.487 & 0.169 & 0.724 & 0.274 & 0.569 & 48.904 & 0.170 & \underline{0.708} & 0.271 & 0.565 & 88.951 & 0.157 & \textbf{0.651} & 0.249 & 0.533 & 197.756 & 0.137 & \underline{0.653} & 0.224 & 0.493 \\
 &  & DBLoss & 8.903 & 0.243 & \textbf{0.737} & 0.366 & 0.579 & 18.264 & 0.157 & \textbf{0.710} & 0.257 & 0.581 & 20.946 & 0.152 & \underline{0.648} & 0.246 & 0.558 & \underline{45.166} & 0.139 & \textbf{0.653} & 0.229 & 0.539 \\
 &  & PS & \underline{6.590} & 0.325 & 0.691 & 0.442 & \underline{0.586} & \underline{12.717} & 0.272 & 0.631 & 0.380 & \underline{0.584} & \underline{14.192} & 0.247 & 0.577 & 0.346 & \textbf{0.563} & \textbf{39.925} & \underline{0.243} & 0.597 & 0.346 & \underline{0.544} \\
 &  & \textbf{APAL} & \textbf{6.476} & \textbf{0.458} & 0.651 & \textbf{0.537} & 0.585 & \textbf{7.235} & \textbf{0.426} & 0.606 & \textbf{0.500} & 0.580 & \textbf{14.006} & \textbf{0.377} & 0.602 & \textbf{0.463} & \underline{0.561} & 61.219 & \textbf{0.256} & 0.627 & \textbf{0.363} & \textbf{0.544} \\
\cline{2-23}
 & \multirow{6}{*}{TSMixer} & MSE & \underline{22.100} & \underline{0.377} & 0.531 & \underline{0.441} & \underline{0.585} & \underline{42.651} & \underline{0.360} & 0.470 & \underline{0.408} & 0.575 & \underline{64.549} & \underline{0.337} & 0.472 & \underline{0.393} & 0.576 & \textbf{101.023} & \textbf{0.306} & 0.461 & \textbf{0.368} & 0.566 \\
 &  & MAE & 52.390 & 0.154 & \textbf{0.617} & 0.247 & 0.580 & 103.564 & 0.080 & \textbf{0.586} & 0.140 & 0.580 & 198.958 & 0.075 & \underline{0.531} & 0.132 & 0.572 & 445.101 & 0.081 & \textbf{0.580} & 0.142 & 0.558 \\
 &  & TildeQ & 40.027 & 0.219 & \underline{0.609} & 0.322 & \textbf{0.587} & 68.755 & 0.194 & \underline{0.563} & 0.289 & \textbf{0.589} & 122.524 & 0.171 & \textbf{0.542} & 0.260 & \textbf{0.583} & 304.810 & 0.163 & \underline{0.520} & 0.248 & \textbf{0.569} \\
 &  & DBLoss & 44.137 & 0.206 & 0.598 & 0.306 & 0.580 & 85.551 & 0.140 & 0.557 & 0.224 & 0.580 & 156.180 & 0.119 & 0.513 & 0.193 & 0.571 & 379.732 & 0.135 & 0.512 & 0.214 & 0.558 \\
 &  & PS & 29.558 & 0.310 & 0.551 & 0.396 & 0.585 & 59.699 & 0.274 & 0.500 & 0.354 & \underline{0.586} & 110.913 & 0.246 & 0.485 & 0.326 & \underline{0.580} & 233.766 & 0.229 & 0.470 & 0.308 & \underline{0.569} \\
 &  & \textbf{APAL} & \textbf{4.304} & \textbf{0.575} & 0.559 & \textbf{0.567} & 0.581 & \textbf{6.552} & \textbf{0.543} & 0.502 & \textbf{0.522} & 0.579 & \textbf{15.495} & \textbf{0.474} & 0.492 & \textbf{0.482} & 0.574 & \underline{119.491} & \underline{0.280} & 0.492 & \underline{0.356} & 0.562 \\
\cline{2-23}
 & \multirow{6}{*}{TiDE} & MSE & \textbf{7.762} & \underline{0.333} & 0.702 & \underline{0.452} & \textbf{0.572} & \underline{12.894} & \underline{0.291} & 0.633 & \underline{0.399} & \textbf{0.575} & \underline{17.540} & \underline{0.286} & 0.614 & \underline{0.390} & \textbf{0.564} & \underline{35.033} & \textbf{0.285} & 0.632 & \textbf{0.392} & \textbf{0.546} \\
 &  & MAE & 11.797 & 0.223 & \textbf{0.754} & 0.345 & 0.566 & 21.126 & 0.176 & \textbf{0.668} & 0.279 & 0.569 & 27.387 & 0.185 & \textbf{0.656} & 0.289 & 0.559 & 57.189 & 0.177 & \textbf{0.677} & 0.281 & 0.542 \\
 &  & TildeQ & \underline{8.729} & 0.256 & \underline{0.731} & 0.379 & 0.558 & 13.636 & 0.210 & \underline{0.658} & 0.319 & 0.558 & \textbf{15.431} & 0.213 & 0.631 & 0.319 & 0.545 & \textbf{24.883} & 0.197 & 0.643 & 0.301 & 0.520 \\
 &  & DBLoss & 9.528 & 0.259 & 0.724 & 0.382 & 0.569 & 15.094 & 0.257 & 0.626 & 0.364 & 0.567 & 23.248 & 0.218 & \underline{0.641} & 0.326 & 0.563 & 47.532 & 0.213 & \underline{0.650} & 0.321 & 0.545 \\
 &  & PS & 9.146 & 0.275 & 0.712 & 0.397 & 0.570 & 16.640 & 0.235 & 0.654 & 0.346 & \underline{0.573} & 23.108 & 0.228 & 0.637 & 0.336 & \underline{0.563} & 44.831 & 0.221 & 0.644 & 0.330 & \underline{0.545} \\
 &  & \textbf{APAL} & 9.042 & \textbf{0.455} & 0.605 & \textbf{0.519} & \underline{0.570} & \textbf{11.864} & \textbf{0.436} & 0.519 & \textbf{0.474} & 0.569 & 21.312 & \textbf{0.365} & 0.520 & \textbf{0.429} & 0.557 & 64.403 & \underline{0.272} & 0.574 & \underline{0.369} & 0.541 \\
\cline{2-23}
 & \multirow{6}{*}{iTransformer} & MSE & 8.832 & \underline{0.433} & 0.786 & \underline{0.559} & 0.625 & \underline{13.326} & \textbf{0.476} & 0.733 & \textbf{0.578} & \textbf{0.619} & \underline{14.381} & \textbf{0.458} & 0.720 & \textbf{0.560} & \textbf{0.604} & \textbf{15.164} & \textbf{0.466} & 0.704 & \textbf{0.561} & \underline{0.584} \\
 &  & MAE & 12.925 & 0.311 & \textbf{0.833} & 0.453 & \textbf{0.641} & 19.805 & 0.231 & \underline{0.765} & 0.354 & 0.604 & 24.988 & 0.232 & \underline{0.754} & 0.355 & 0.596 & 43.715 & 0.213 & \underline{0.770} & 0.334 & 0.582 \\
 &  & TildeQ & 11.277 & 0.387 & \underline{0.831} & 0.528 & 0.628 & 17.503 & 0.401 & \textbf{0.789} & \underline{0.532} & 0.611 & 18.716 & \underline{0.376} & \textbf{0.778} & \underline{0.507} & 0.596 & 24.002 & \underline{0.333} & \textbf{0.775} & \underline{0.466} & 0.569 \\
 &  & DBLoss & 10.478 & 0.384 & 0.819 & 0.523 & \underline{0.640} & 16.978 & 0.351 & 0.746 & 0.477 & 0.612 & 20.044 & 0.326 & 0.710 & 0.447 & 0.599 & 31.858 & 0.275 & 0.698 & 0.394 & 0.577 \\
 &  & PS & \underline{8.395} & 0.431 & 0.792 & 0.558 & 0.635 & 13.567 & 0.410 & 0.714 & 0.521 & 0.613 & 16.728 & 0.352 & 0.684 & 0.465 & 0.600 & 27.691 & 0.298 & 0.652 & 0.409 & 0.573 \\
 &  & \textbf{APAL} & \textbf{3.510} & \textbf{0.469} & 0.709 & \textbf{0.565} & 0.627 & \textbf{5.167} & \underline{0.411} & 0.653 & 0.505 & \underline{0.613} & \textbf{7.114} & 0.370 & 0.683 & 0.480 & \underline{0.604} & \underline{15.278} & 0.309 & 0.714 & 0.432 & \textbf{0.592} \\
\bottomrule
\end{tabular}
}
\end{table*}

\subsection{Cross-domain APAL hyperparameter ablation}
\label{app:cross_domain_ablation}

To assess hyperparameter transfer beyond the two peak-critical crowd datasets, we analyze the APAL grids for ETT, Exchange, and Weather datasets using TSMixer architecture. The grid uses $\lambda_u,\lambda_p\in\{1,2,5,10\}$ and $\tau\in\{0.8,0.9,0.95\}$ for each horizon $H\in\{96,192,336,720\}$.
Table~\ref{tab:cross_domain_ablation} reports, for each dataset--horizon pair, the best observed $\mathrm{MSE}_1$ configuration and the corresponding aggregate-MSE change relative to the neutral APAL setting $(\lambda_u,\lambda_p)=(1,1)$, which reduces to MAE and makes $\tau$ inactive. These comparisons are therefore within the APAL grid and complement, rather than replace, the MSE-vs.-APAL benchmark in Table~\ref{tab:mse_apal_comparison}.

The results reinforce three practical points. First, strong peak weights can produce large tail-error reductions on the ETTh1 and ETTm1 datasets, but usually at a substantial aggregate MSE cost. Second, the Exchange dataset is a boundary case: the average best $\mathrm{MSE}_1$ gain is 17.7\%, but gains are small at shorter horizons and Peak F1 changes little, indicating limited practical value unless tail error is the dominant objective. Third, Weather dataset has consistent tail gains across horizons (15.7--28.2\%), but the best tail settings increase aggregate MSE by 69.0--131.2\%. Thus, the Weather dataset supports the controllability claim while also showing why APAL should be selected based on validation trade-offs rather than adopted blindly.

\begin{table*}[ht]
\centering
\caption{APAL hyperparameter ablation on ETT, Exchange, and Weather dataset with TSMixer architecture. Neutral denotes $(\lambda_u,\lambda_p)=(1,1)$, where APAL reduces to MAE. Gain and aggregate-MSE change are relative to the neutral APAL setting for the same dataset and horizon. Best Peak F1 reports the maximum Peak F1 observed in the same grid.}
\label{tab:cross_domain_ablation}
\resizebox{\textwidth}{!}{%
\begin{tabular}{llcccccclc}
\toprule
Dataset & $H$ & Neutral $\mathrm{MSE}_1$ & Best $(\lambda_u,\lambda_p,\tau)$ & Best $\mathrm{MSE}_1$ & Gain & Neutral MSE & Best MSE & MSE Change & Best Peak F1 \\
\midrule
\multirow{4}{*}{ETTh1} & 96 & 1.791 & (5, 10, 0.95) & 0.278 & 84.5\% & 0.503 & 1.256 & 149.9\% & 0.489 \\
 & 192 & 1.961 & (5, 10, 0.90) & 0.357 & 81.8\% & 0.597 & 1.323 & 121.7\% & 0.453 \\
 & 336 & 2.160 & (5, 10, 0.90) & 0.416 & 80.8\% & 0.682 & 1.194 & 75.1\% & 0.422 \\
 & 720 & 2.475 & (10, 10, 0.80) & 0.557 & 77.5\% & 0.754 & 1.883 & 149.7\% & 0.415 \\
\midrule
\multirow{4}{*}{ETTh2} & 96 & 0.707 & (1, 5, 0.80) & 0.460 & 34.9\% & 0.815 & 1.268 & 55.6\% & 0.368 \\
 & 192 & 0.598 & (1, 2, 0.90) & 0.590 & 1.4\% & 1.894 & 1.933 & 2.0\% & 0.309 \\
 & 336 & 0.836 & (1, 10, 0.90) & 0.785 & 6.1\% & 2.076 & 2.256 & 8.6\% & 0.297 \\
 & 720 & 1.217 & (2, 5, 0.90) & 1.087 & 10.7\% & 1.906 & 2.652 & 39.1\% & 0.272 \\
\midrule
\multirow{4}{*}{ETTm1} & 96 & 1.999 & (10, 10, 0.95) & 0.276 & 86.2\% & 0.454 & 1.044 & 129.9\% & 0.428 \\
 & 192 & 2.039 & (10, 5, 0.90) & 0.315 & 84.6\% & 0.486 & 1.061 & 118.2\% & 0.416 \\
 & 336 & 2.097 & (10, 10, 0.95) & 0.362 & 82.8\% & 0.532 & 1.275 & 139.7\% & 0.387 \\
 & 720 & 2.193 & (5, 10, 0.80) & 0.411 & 81.3\% & 0.621 & 1.332 & 114.4\% & 0.369 \\
\midrule
\multirow{4}{*}{ETTm2} & 96 & 0.416 & (2, 10, 0.95) & 0.310 & 25.5\% & 0.238 & 0.402 & 69.1\% & 0.331 \\
 & 192 & 0.624 & (2, 5, 0.95) & 0.441 & 29.3\% & 0.296 & 0.531 & 79.5\% & 0.280 \\
 & 336 & 0.885 & (1, 10, 0.90) & 0.556 & 37.2\% & 0.504 & 0.841 & 67.0\% & 0.258 \\
 & 720 & 1.038 & (1, 10, 0.95) & 0.758 & 27.0\% & 1.981 & 2.090 & 5.5\% & 0.239 \\
\midrule
\multirow{4}{*}{Exchange} & 96 & 0.120 & (1, 2, 0.90) & 0.117 & 2.1\% & 0.111 & 0.118 & 6.0\% & 0.211 \\
 & 192 & 0.201 & (1, 2, 0.80) & 0.192 & 4.3\% & 0.190 & 0.215 & 13.1\% & 0.195 \\
 & 336 & 0.383 & (1, 10, 0.90) & 0.285 & 25.5\% & 0.299 & 0.415 & 38.9\% & 0.182 \\
 & 720 & 0.746 & (1, 10, 0.80) & 0.458 & 38.6\% & 0.801 & 0.860 & 7.4\% & 0.278 \\
\midrule
\multirow{4}{*}{Weather} & 96 & 0.813 & (2, 10, 0.80) & 0.583 & 28.2\% & 0.175 & 0.354 & 102.5\% & 0.303 \\
 & 192 & 0.916 & (2, 10, 0.90) & 0.718 & 21.6\% & 0.212 & 0.359 & 69.0\% & 0.310 \\
 & 336 & 0.985 & (5, 5, 0.95) & 0.822 & 16.6\% & 0.256 & 0.593 & 131.2\% & 0.282 \\
 & 720 & 1.128 & (5, 5, 0.95) & 0.951 & 15.7\% & 0.312 & 0.673 & 115.9\% & 0.249 \\
\bottomrule
\end{tabular}
}
\end{table*}

\subsection{Ablation details}
\begin{figure}[ht]
    \centering
    \includegraphics[width=\linewidth/2]{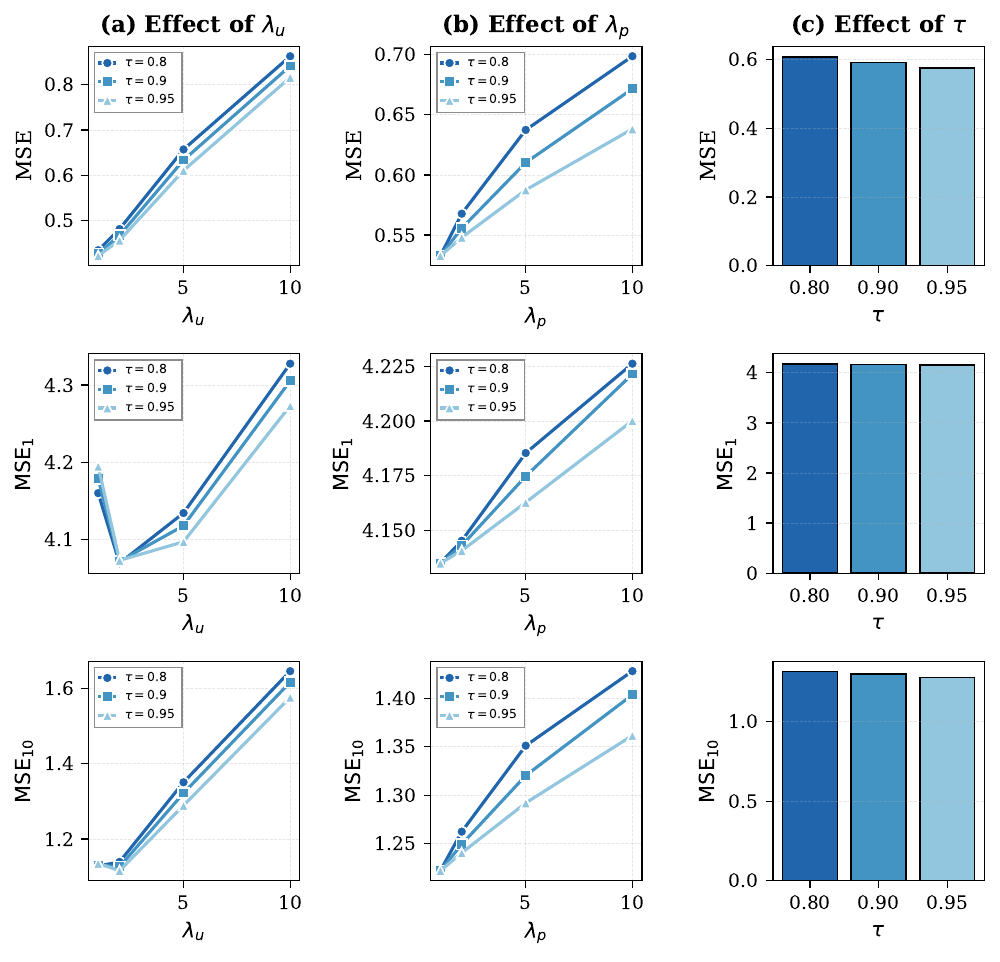}
    \caption{Marginal effects of APAL hyperparameters on MSE, $\mathrm{MSE}_1$, and $\mathrm{MSE}_{10}$ (Pedestrian Dataset, aggregated across horizons). Columns: (a)~effect of $\lambda_u$, (b)~effect of $\lambda_p$, (c)~effect of $\tau$. Lines in (a) and (b) represent different peak thresholds $\tau \in \{0.8, 0.9, 0.95\}$.}
    \label{fig:ablationMarginalCombined}
\end{figure}

\begin{figure}[ht]
    \centering
    \includegraphics[width=0.75\linewidth]{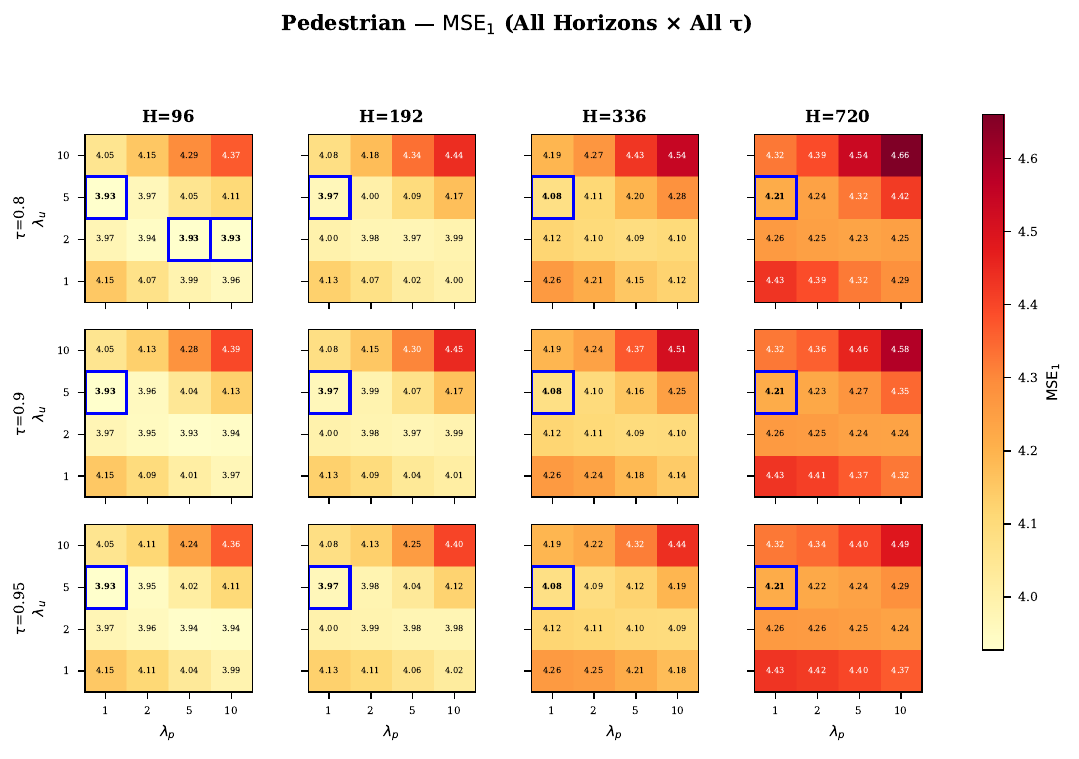}
    \caption{Ablation study of APAL hyperparameters on the Pedestrian dataset, evaluated using $\mathrm{MSE}_1$ (mean squared error on the top 1\% peak values). Each panel shows a heatmap of performance across the underestimation penalty, $\lambda_u$ (rows), and the peak weight,$\lambda_p$ (columns). Rows correspond to peak threshold values $\tau \in \{0.80, 0.90, 0.95\}$, and columns correspond to forecast horizons $H \in \{96, 192, 336, 720\}$. Lower values (darker) indicate better peak prediction accuracy. Blue boxes highlight optimal configurations within each panel.}
    \label{fig:ablation_heatmap_mse_1_ped}
\end{figure}

In addition to the main text, we provide marginal effect plots (Figure~\ref{fig:ablationMarginalCombined}) and heatmaps (Figure~\ref{fig:ablation_heatmap_mse_1_ped}) for different hyperparameters of APAL $(\lambda_u,\lambda_p,\tau)$ on the Pedestrian Dataset.

Figure~\ref{fig:ablationMarginalCombined} shows the marginal effects of each APAL hyperparameter ($\lambda_u$, $\lambda_p$, $\tau$) on MSE, $\text{MSE}_1$, and $\text{MSE}_{10}$, aggregated across all horizons on the Pedestrian dataset; Beach results in Figure~\ref{fig:pareto} exhibit similar trends.

The under-prediction penalty $\lambda_u$ has the strongest influence on tail metrics: increasing $\lambda_u$ from 1 to 2 yields substantial reductions in both $\text{MSE}_1$ and $\text{MSE}_{10}$ with only a modest increase in overall MSE, while larger values improve tails further at the cost of aggregate error. The peak emphasis weight $\lambda_p$ has a smaller marginal effect, suggesting that asymmetric under-prediction penalties contribute more to tail improvements than explicit peak weighing alone. Increasing the peak threshold $\tau$ generally reduces error for both average and extreme cases, as higher thresholds focus on fewer, more extreme peak regions. 

\subsection{Comparison to Pinball Loss}
\label{app:pinball}
A natural baseline for asymmetric objectives is the quantile (pinball) loss~\cite{SMYL2026103637, 10230996}, widely used in probabilistic forecasting. Since APAL penalizes under-prediction more heavily (via $\lambda_u > 1$), one might expect pinball loss at a high quantile to achieve similar benefits. We compare APAL against pinball loss at $q \in \{0.90, 0.95\}$ across all datasets and horizons (Table~\ref{tab:pinball_vs_apal}).

\paragraph{Pinball loss formulation.}
For prediction $\hat{y}_{h,c}$ and ground truth $y_{h,c}$, pinball loss at quantile $q$ is:
\begin{equation}
    \mathcal{L}_{\text{pinball}}(\hat{y}_{h,c}, y_{h,c}; q) = \max\bigl(q \cdot (y_{h,c} - \hat{y}_{h,c}),\; (q - 1) \cdot (y_{h,c} - \hat{y}_{h,c})\bigr).
\end{equation}
At $q = 0.90$, under-predictions receive $9\times$ the weight of over-predictions. Unlike APAL, pinball loss applies uniform asymmetry across all time steps without any notion of peak regions.

Table~\ref{tab:pinball_vs_apal} reveals three patterns:
\emph{(1) APAL achieves superior aggregate MSE.}
On the Pedestrian dataset at $H=96$, APAL achieves MSE of 0.537 vs.\ 0.763 for pinball ($q=0.90$), a 29.6\% improvement. This arises because pinball loss with high $q$ systematically biases predictions upward, inflating errors in non-peak regions, whereas APAL's peak weighting emphasises local maxima only.

\emph{(2) APAL excels on peak-critical datasets.}
On Pedestrian and Beach datasets, APAL consistently achieves the best MSE, MSE$_{10}$, and MSE$_{1}$ across all horizons, confirming that peak emphasis provides benefits beyond pure asymmetry.

\emph{(3) Dataset characteristics modulate relative advantage.}
On datasets with less pronounced peaks (ETTh1, ETTm2), pinball ($q=0.90$) achieves competitive tail metrics but at the cost of degraded aggregate MSE. On Exchange, pinball becomes competitive only at $H=720$, where peak structure is less predictable.

Increasing $q$ from 0.90 to 0.95 consistently degrades performance, indicating that the optimal quantile is data-dependent and not interpretable in terms of peak-criticality. In contrast, APAL's hyperparameters ($\lambda_u$, $\lambda_p$, $\tau$) directly encode peak-aware objectives and are tunable via validation. These results confirm that APAL's dual mechanism---asymmetry \emph{and} peak emphasis---provides complementary benefits that cannot be replicated by asymmetric losses alone.

\begin{table*}[ht]
\centering
\caption{Comparison of Pinball Loss ($q \in \{0.9, 0.95\}$) and APAL Loss. Lower values are better. Best results are \textbf{bold}, second best are \underline{underlined}. Model used shown in parentheses.}
\label{tab:pinball_vs_apal}
\resizebox{\textwidth}{!}{%
\begin{tabular}{c|l|ccc|ccc|ccc|ccc}
\toprule
\multirow{2}{*}{Dataset} & \multirow{2}{*}{Loss Function} & \multicolumn{3}{c|}{$H=96$} & \multicolumn{3}{c|}{$H=192$} & \multicolumn{3}{c|}{$H=336$} & \multicolumn{3}{c}{$H=720$} \\
\cmidrule(lr){3-5} \cmidrule(lr){6-8} \cmidrule(lr){9-11} \cmidrule(lr){12-14}
 &  & MSE & MSE$_{10}$ & MSE$_{1}$ & MSE & MSE$_{10}$ & MSE$_{1}$ & MSE & MSE$_{10}$ & MSE$_{1}$ & MSE & MSE$_{10}$ & MSE$_{1}$ \\
\midrule
 \multirow{3}{*}{Beach (DLinear)} & Pinball ($\tau=0.90$) & \underline{0.961} & \underline{3.468} & \underline{8.936} & \underline{0.969} & \underline{3.448} & \underline{8.200} & \underline{0.988} & \underline{3.453} & \underline{7.831} & \underline{1.029} & \underline{3.551} & \underline{8.574} \\
  & Pinball ($\tau=0.95$) & 0.975 & \textbf{3.461} & \textbf{8.878} & 0.978 & \textbf{3.446} & \textbf{8.176} & 1.001 & \textbf{3.447} & \textbf{7.789} & 1.042 & \textbf{3.545} & \textbf{8.530} \\
  & \textbf{APAL} & \textbf{0.888} & 3.590 & 9.259 & \textbf{0.826} & 3.685 & 9.111 & \textbf{0.812} & 3.897 & 9.348 & \textbf{0.830} & 4.321 & 11.017 \\
\midrule
 \multirow{3}{*}{Pedestrian (DLinear)} & Pinball ($\tau=0.90$) & \underline{0.763} & \underline{1.436} & \underline{4.028} & \underline{0.669} & \underline{1.358} & \underline{4.102} & \underline{0.674} & \underline{1.375} & \underline{4.140} & \underline{0.712} & \underline{1.428} & \underline{4.271} \\
  & Pinball ($\tau=0.95$) & 0.971 & 1.718 & 4.212 & 0.874 & 1.649 & 4.305 & 0.872 & 1.656 & 4.329 & 0.902 & 1.690 & 4.441 \\
  & \textbf{APAL} & \textbf{0.537} & \textbf{1.166} & \textbf{3.929} & \textbf{0.476} & \textbf{1.122} & \textbf{4.013} & \textbf{0.481} & \textbf{1.143} & \textbf{4.064} & \textbf{0.499} & \textbf{1.181} & \textbf{4.206} \\
\midrule
 \multirow{3}{*}{ETTh1 (iTransformer)} & Pinball ($\tau=0.90$) & \underline{1.123} & \textbf{0.402} & \textbf{0.369} & \underline{1.222} & \textbf{0.432} & \textbf{0.411} & \underline{1.262} & \textbf{0.438} & \textbf{0.439} & \underline{1.261} & \textbf{0.471} & \textbf{0.530} \\
  & Pinball ($\tau=0.95$) & 1.799 & 0.739 & 0.482 & 1.925 & 0.777 & 0.545 & 2.058 & 0.780 & 0.575 & 3.035 & 1.131 & 0.795 \\
  & \textbf{APAL} & \textbf{1.047} & \underline{0.477} & \underline{0.408} & \textbf{1.089} & \underline{0.498} & \underline{0.464} & \textbf{1.059} & \underline{0.467} & \underline{0.481} & \textbf{0.994} & \underline{0.472} & \underline{0.594} \\
\midrule
 \multirow{3}{*}{ETTm1 (TSMixer)} & Pinball ($\tau=0.90$) & \textbf{0.694} & \underline{0.316} & 0.564 & \textbf{0.814} & \textbf{0.327} & 0.499 & \textbf{0.943} & \underline{0.383} & 0.579 & \underline{1.090} & \underline{0.476} & 0.675 \\
  & Pinball ($\tau=0.95$) & 1.066 & 0.362 & \textbf{0.375} & 1.186 & 0.408 & \underline{0.466} & 1.297 & 0.461 & \underline{0.539} & 1.516 & 0.523 & \underline{0.612} \\
  & \textbf{APAL} & \underline{0.773} & \textbf{0.304} & \underline{0.383} & \underline{0.863} & \underline{0.330} & \textbf{0.376} & \underline{0.973} & \textbf{0.371} & \textbf{0.432} & \textbf{1.035} & \textbf{0.441} & \textbf{0.522} \\
\midrule
 \multirow{3}{*}{ETTm2 (TSMixer)} & Pinball ($\tau=0.90$) & \underline{1.574} & \underline{0.892} & \underline{0.729} & \underline{2.079} & \textbf{1.143} & \textbf{0.960} & \underline{2.657} & \textbf{1.394} & \textbf{1.172} & \textbf{9.269} & \textbf{7.053} & \textbf{5.753} \\
  & Pinball ($\tau=0.95$) & 2.012 & 1.198 & 0.948 & 2.475 & 1.386 & 1.114 & 3.283 & 1.813 & 1.469 & 10.973 & 8.579 & 7.067 \\
  & \textbf{APAL} & \textbf{1.321} & \textbf{0.773} & \textbf{0.651} & \textbf{2.022} & \underline{1.146} & \underline{0.973} & \textbf{2.604} & \underline{1.418} & \underline{1.222} & \underline{9.655} & \underline{7.610} & \underline{6.298} \\
\midrule
 \multirow{3}{*}{Exchange (TSMixer)} & Pinball ($\tau=0.90$) & \underline{0.718} & \underline{0.784} & \underline{0.524} & \underline{1.375} & \underline{1.291} & \underline{0.852} & \textbf{2.494} & \underline{1.734} & \textbf{1.069} & \textbf{6.835} & \textbf{3.314} & \textbf{1.750} \\
  & Pinball ($\tau=0.95$) & 1.561 & 1.934 & 1.294 & 2.951 & 3.087 & 1.957 & 5.413 & 4.300 & 2.411 & 16.058 & 9.280 & 5.003 \\
  & \textbf{APAL} & \textbf{0.563} & \textbf{0.560} & \textbf{0.352} & \textbf{1.221} & \textbf{1.117} & \textbf{0.731} & \underline{2.499} & \textbf{1.707} & \underline{1.078} & \underline{8.384} & \underline{3.817} & \underline{1.759} \\
\bottomrule
\end{tabular}
}
\end{table*}



\subsection{Qualitative examples and failure cases}

\begin{figure}[ht]
    \centering
    \begin{subfigure}[b]{0.49\textwidth}
        \centering
        \includegraphics[width=\linewidth]{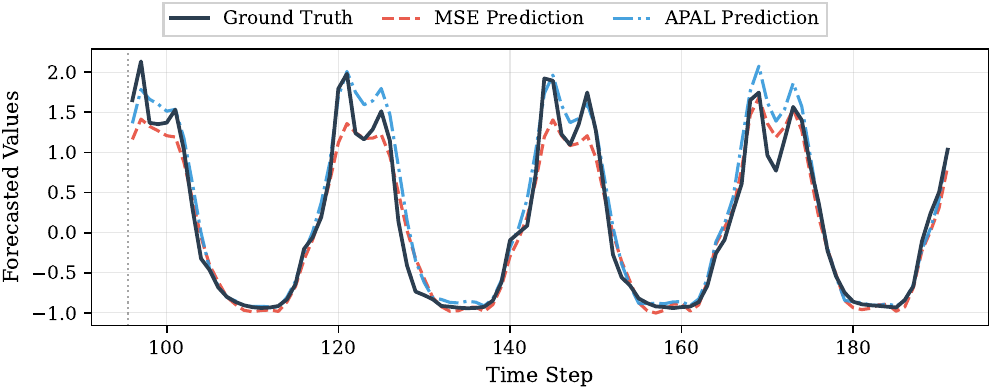}
        \caption{Pedestrian dataset}
        \label{fig:qual_peaks1}
    \end{subfigure}
    \hfill 
    \begin{subfigure}[b]{0.49\textwidth}
        \centering
        \includegraphics[width=\linewidth]{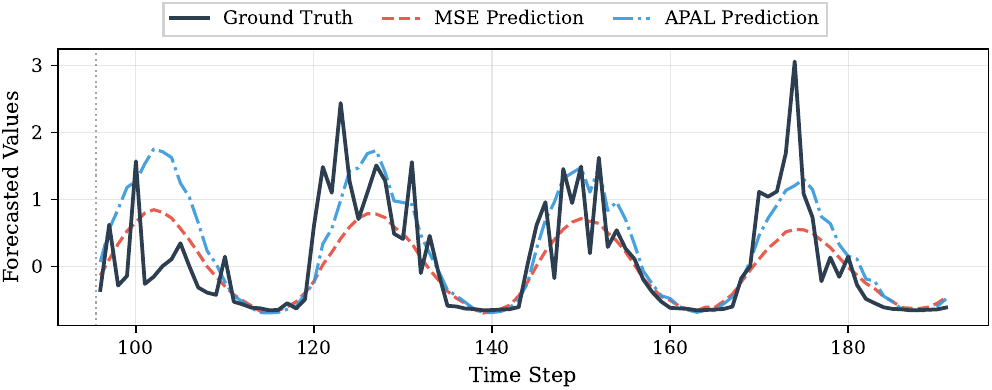}
        \caption{Beach dataset}
        \label{fig:qual_peaks2}
    \end{subfigure}
    
    \caption{
    Qualitative comparison of long-horizon forecasts trained with MSE versus APAL loss. APAL reduces underestimation at high-demand periods in both the (a) Pedestrian and (b) Beach datasets. Values shown are standardized (z-score normalized).
    }
    \label{fig:qual_peaks_combined}
\end{figure}

Figure~\ref{fig:qual_peaks_combined} illustrates how APAL improves peak fidelity compared to MSE. On Pedestrian (Figure~\ref{fig:qual_peaks1}), MSE systematically underestimates the regular periodic peaks, while APAL closely tracks ground-truth amplitudes. On Beach (Figure~\ref{fig:qual_peaks2}), peaks are more irregular with greater amplitude variability; MSE underestimates peak magnitude throughout, with largest errors at extreme peaks. APAL reduces this underestimation, consistent with the larger $\text{MSE}_1$ reductions observed on Beach in Table~\ref{tab:combined_results}.

\section{Computation Costs}
\label{app:compute}

\paragraph{Hardware and software.}
All the experiments were implemented in Python (version 3.12) using PyTorch (version 2.4) and  using standard Time Series Library (TSLib)~\cite{wang2024tssurvey} run on an NVIDIA L4 GPU (24\,GB) and 12 cores Intel Xeon CPU. 

\paragraph{Training time and overhead.}
APAL adds only elementwise operations to the standard loss computation and therefore incurs negligible overhead relative to the model forward pass.
Table~\ref{tab:computational_overhead} reports the average per-epoch training time (in seconds) for TSMixer across four datasets (Pedestrian, Beach, Weather, and Exchange) and four prediction horizons (96, 192, 336, 720).
We compare APAL against MSE, MAE, and three peak-sensitive baselines: TildeQ, DBLoss, and PS Loss.

Among all loss functions, APAL exhibits the lowest computational overhead, adding only $+0.23$\,s per epoch on average compared to MSE, a relative increase of approximately $4\%$.
In contrast, the other peak-aware losses incur substantially higher overhead: DBLoss adds $+0.85$\,s ($15\%$), PS adds $+1.37$\,s ($24\%$), and TildeQ adds $+2.15$\,s ($37\%$) per epoch.
MAE, which like MSE involves only simple elementwise operations, adds a negligible $+0.09$\,s ($1.6\%$).

These results demonstrate that APAL achieves its peak-sensitivity benefits at a computational cost comparable to standard losses, making it practical for large-scale time-series forecasting applications where training efficiency is critical.

\begin{table*}[ht]
\caption{Computational overhead of various loss functions compared to MSE loss. 
Per-epoch overhead measures the additional computation time per training epoch (in seconds).}
\label{tab:computational_overhead}
\vskip 0.15in
\begin{center}
\begin{small}
\begin{tabular}{llcccccc}
\toprule
Dataset & Horizon & MSE & MAE & TildeQ & DBLoss & PS & APAL \\
\midrule
Pedestrian & 96 & 8.59 & 8.77 (+0.18) & 12.70 (+4.11) & 10.47 (+1.88) & 11.21 (+2.62) & 9.12 (+0.53) \\
 & 192 & 8.94 & 9.04 (+0.10) & 13.08 (+4.14) & 10.71 (+1.77) & 11.47 (+2.53) & 9.47 (+0.53) \\
 & 336 & 9.36 & 9.47 (+0.11) & 13.48 (+4.12) & 11.13 (+1.77) & 11.71 (+2.35) & 9.94 (+0.58) \\
 & 720 & 10.32 & 10.42 (+0.10) & 14.54 (+4.22) & 12.14 (+1.82) & 12.77 (+2.45) & 10.88 (+0.56) \\
\textit{Avg.} & -- & 9.30 & 9.43 (+0.12) & 13.45 (+4.15) & 11.11 (+1.81) & 11.79 (+2.49) & 9.85 (+0.55) \\
\cmidrule{1-8}
Beach & 96 & 2.94 & 3.02 (+0.08) & 4.08 (+1.14) & 3.38 (+0.44) & 3.60 (+0.66) & 3.10 (+0.16) \\
 & 192 & 3.28 & 3.35 (+0.07) & 4.28 (+1.00) & 3.60 (+0.32) & 3.80 (+0.52) & 3.33 (+0.05) \\
 & 336 & 3.77 & 3.83 (+0.06) & 4.96 (+1.19) & 4.20 (+0.43) & 4.33 (+0.56) & 3.93 (+0.16) \\
 & 720 & 4.74 & 4.79 (+0.05) & 5.88 (+1.14) & 5.27 (+0.53) & 5.27 (+0.53) & 4.90 (+0.16) \\
\textit{Avg.} & -- & 3.68 & 3.75 (+0.06) & 4.80 (+1.12) & 4.11 (+0.43) & 4.25 (+0.57) & 3.81 (+0.13) \\
\cmidrule{1-8}
Weather & 96 & 6.74 & 6.96 (+0.22) & 9.33 (+2.59) & 7.64 (+0.90) & 8.53 (+1.79) & 7.11 (+0.37) \\
 & 192 & 7.46 & 7.53 (+0.07) & 9.93 (+2.47) & 8.23 (+0.77) & 8.84 (+1.38) & 7.74 (+0.28) \\
 & 336 & 9.18 & 9.29 (+0.11) & 11.83 (+2.65) & 10.14 (+0.96) & 10.62 (+1.44) & 9.36 (+0.18) \\
 & 720 & 12.26 & 12.41 (+0.15) & 15.92 (+3.66) & 13.55 (+1.29) & 13.52 (+1.26) & 12.31 (+0.05) \\
\textit{Avg.} & -- & 8.91 & 9.05 (+0.14) & 11.75 (+2.84) & 9.89 (+0.98) & 10.38 (+1.47) & 9.11 (+0.20) \\
\cmidrule{1-8}
Exchange & 96 & 1.07 & 1.11 (+0.04) & 1.55 (+0.48) & 1.25 (+0.18) & 1.69 (+0.62) & 1.12 (+0.05) \\
 & 192 & 1.13 & 1.15 (+0.02) & 1.52 (+0.39) & 1.31 (+0.18) & 1.89 (+0.76) & 1.15 (+0.02) \\
 & 336 & 1.19 & 1.30 (+0.11) & 1.77 (+0.58) & 1.39 (+0.20) & 1.65 (+0.46) & 1.22 (+0.03) \\
 & 720 & 1.29 & 1.32 (+0.03) & 1.84 (+0.55) & 1.42 (+0.13) & 3.34 (+2.05) & 1.33 (+0.04) \\
\textit{Avg.} & -- & 1.17 & 1.22 (+0.05) & 1.67 (+0.50) & 1.34 (+0.17) & 2.14 (+0.97) & 1.21 (+0.04) \\
\midrule
\textbf{Overall} & -- & \textbf{5.77} & \textbf{5.86} (+0.09) & \textbf{7.92} (+2.15) & \textbf{6.61} (+0.85) & \textbf{7.14} (+1.37) & \textbf{5.99} (+0.23) \\
\bottomrule
\end{tabular}
\end{small}
\end{center}
\vskip -0.1in
\end{table*}

\section{Reproducibility Details}
\label{app:repro}
We fix random seeds for data shuffling, initialization, and training, and report the seed used in all experiments. We use chronological splits. Normalization statistics are computed on the training split only. Inverse transformation is applied prior to metric computation. We provide scripts to reproduce training, evaluation, and plotting. We include example commands for each dataset and backbone, and document all hyper-parameters. All reported results are obtained from the checkpoint with best validation performance under the specified selection criterion. The scripts used for training and testing the model will be shared as a GitHub repository to ensure reproducibility of the experiments.

\section{Dataset Statement and Reproducibility}
\label{app:private}
The Beach Visitor Count dataset is provided by a third-party data provider (RESONO) and derived from aggregated, anonymised, opt-in mobile location signals \cite{resono2025veelgestelde}.
Due to data-sharing restrictions, raw data cannot be released publicly.
To support reproducibility, we release the full preprocessing code, the evaluation code, and all trained model hyperparameters to enable replication on alternative datasets.
We also report results on multiple public benchmarks and an open pedestrian dataset to ensure that conclusions do not rely solely on private data.

\newpage
\section*{NeurIPS Paper Checklist}

The checklist is designed to encourage best practices for responsible machine learning research, addressing issues of reproducibility, transparency, research ethics, and societal impact. Do not remove the checklist: {\bf The papers not including the checklist will be desk rejected.} The checklist should follow the references and follow the (optional) supplemental material.  The checklist does NOT count towards the page
limit. 

Please read the checklist guidelines carefully for information on how to answer these questions. For each question in the checklist:
\begin{itemize}
    \item You should answer \answerYes{}, \answerNo{}, or \answerNA{}.
    \item \answerNA{} means either that the question is Not Applicable for that particular paper or the relevant information is Not Available.
    \item Please provide a short (1--2 sentence) justification right after your answer (even for \answerNA). 
\end{itemize}

{\bf The checklist answers are an integral part of your paper submission.} They are visible to the reviewers, area chairs, senior area chairs, and ethics reviewers. You will also be asked to include it (after eventual revisions) with the final version of your paper, and its final version will be published with the paper.

The reviewers of your paper will be asked to use the checklist as one of the factors in their evaluation. While \answerYes{} is generally preferable to \answerNo{}, it is perfectly acceptable to answer \answerNo{} provided a proper justification is given (e.g., error bars are not reported because it would be too computationally expensive'' or ``we were unable to find the license for the dataset we used''). In general, answering \answerNo{} or \answerNA{} is not grounds for rejection. While the questions are phrased in a binary way, we acknowledge that the true answer is often more nuanced, so please just use your best judgment and write a justification to elaborate. All supporting evidence can appear either in the main paper or the supplemental material, provided in appendix. If you answer \answerYes{} to a question, in the justification please point to the section(s) where related material for the question can be found.

IMPORTANT, please:
\begin{itemize}
    \item {\bf Delete this instruction block, but keep the section heading ``NeurIPS Paper Checklist"},
    \item  {\bf Keep the checklist subsection headings, questions/answers and guidelines below.}
    \item {\bf Do not modify the questions and only use the provided macros for your answers}.
\end{itemize}


\begin{enumerate}

\item {\bf Claims}
    \item[] Question: Do the main claims made in the abstract and introduction accurately reflect the paper's contributions and scope?
    \item[] Answer: \answerYes{}
    \item[] Justification: The abstract and introduction state the paper's main contributions: APAL, the peak-critical evaluation protocol, and the empirical trade-off between aggregate and peak performance. These are supported by the method and experimental sections (Sections~\ref{sec:prelim}, \ref{sec:results}, and \ref{sec:ablation}).
    \item[] Guidelines:
    \begin{itemize}
        \item The answer \answerNA{} means that the abstract and introduction do not include the claims made in the paper.
        \item The abstract and/or introduction should clearly state the claims made, including the contributions made in the paper and important assumptions and limitations. A \answerNo{} or \answerNA{} answer to this question will not be perceived well by the reviewers. 
        \item The claims made should match theoretical and experimental results, and reflect how much the results can be expected to generalize to other settings. 
        \item It is fine to include aspirational goals as motivation as long as it is clear that these goals are not attained by the paper. 
    \end{itemize}

\item {\bf Limitations}
    \item[] Question: Does the paper discuss the limitations of the work performed by the authors?
    \item[] Answer: \answerYes{}
    \item[] Justification:  The paper explicitly discusses failure cases and limits of applicability, including false-positive peaks, sensitivity to distribution shift, and weaker benefits on noise-dominated datasets (Section~\ref{sec:limitations} and Appendix Section~\ref{app:additional}.
    \item[] Guidelines:
    \begin{itemize}
        \item The answer \answerNA{} means that the paper has no limitation while the answer \answerNo{} means that the paper has limitations, but those are not discussed in the paper. 
        \item The authors are encouraged to create a separate ``Limitations'' section in their paper.
        \item The paper should point out any strong assumptions and how robust the results are to violations of these assumptions (e.g., independence assumptions, noiseless settings, model well-specification, asymptotic approximations only holding locally). The authors should reflect on how these assumptions might be violated in practice and what the implications would be.
        \item The authors should reflect on the scope of the claims made, e.g., if the approach was only tested on a few datasets or with a few runs. In general, empirical results often depend on implicit assumptions, which should be articulated.
        \item The authors should reflect on the factors that influence the performance of the approach. For example, a facial recognition algorithm may perform poorly when image resolution is low or images are taken in low lighting. Or a speech-to-text system might not be used reliably to provide closed captions for online lectures because it fails to handle technical jargon.
        \item The authors should discuss the computational efficiency of the proposed algorithms and how they scale with dataset size.
        \item If applicable, the authors should discuss possible limitations of their approach to address problems of privacy and fairness.
        \item While the authors might fear that complete honesty about limitations might be used by reviewers as grounds for rejection, a worse outcome might be that reviewers discover limitations that aren't acknowledged in the paper. The authors should use their best judgment and recognize that individual actions in favor of transparency play an important role in developing norms that preserve the integrity of the community. Reviewers will be specifically instructed to not penalize honesty concerning limitations.
    \end{itemize}

\item {\bf Theory assumptions and proofs}
    \item[] Question: For each theoretical result, does the paper provide the full set of assumptions and a complete (and correct) proof?
    \item[] Answer:\answerNA{}
    \item[] Justification: The paper does not contain formal theoretical results such as theorems, lemmas, or convergence/generalisation guarantees that would require a separate set of assumptions and proofs. The equations in the main manuscript (Section~\ref{sec:prelim}) are constructive definitions of the APAL loss, its peak-detection mask, and the peak-critical evaluation metrics. Each component is derived directly from standard operations (smoothed envelopes, asymmetric Huber-style penalties, weighted aggregation) and is fully specified inline, with no claims requiring separate proof.
    
    \item[] Guidelines:
    \begin{itemize}
        \item The answer \answerNA{} means that the paper does not include theoretical results. 
        \item All the theorems, formulas, and proofs in the paper should be numbered and cross-referenced.
        \item All assumptions should be clearly stated or referenced in the statement of any theorems.
        \item The proofs can either appear in the main paper or the supplemental material, but if they appear in the supplemental material, the authors are encouraged to provide a short proof sketch to provide intuition. 
        \item Inversely, any informal proof provided in the core of the paper should be complemented by formal proofs provided in appendix or supplemental material.
        \item Theorems and Lemmas that the proof relies upon should be properly referenced. 
    \end{itemize}

    \item {\bf Experimental result reproducibility}
    \item[] Question: Does the paper fully disclose all the information needed to reproduce the main experimental results of the paper to the extent that it affects the main claims and/or conclusions of the paper (regardless of whether the code and data are provided or not)?
    \item[] Answer: \answerYes{}
    \item[] Justification: In the manuscript, we specify datasets, evaluation metrics, model families, tuning protocol, training settings, random seed policy, chronological splits, and reproducibility details (Sections~\ref{sec:experiments}--\ref{sec:ablation}, Appendix~\ref{app:repro}). The scripts for training, evaluation, and plotting will be provided as supplementary material.
    \item[] Guidelines:
    \begin{itemize}
        \item The answer \answerNA{} means that the paper does not include experiments.
        \item If the paper includes experiments, a \answerNo{} answer to this question will not be perceived well by the reviewers: Making the paper reproducible is important, regardless of whether the code and data are provided or not.
        \item If the contribution is a dataset and\slash or model, the authors should describe the steps taken to make their results reproducible or verifiable. 
        \item Depending on the contribution, reproducibility can be accomplished in various ways. For example, if the contribution is a novel architecture, describing the architecture fully might suffice, or if the contribution is a specific model and empirical evaluation, it may be necessary to either make it possible for others to replicate the model with the same dataset, or provide access to the model. In general. releasing code and data is often one good way to accomplish this, but reproducibility can also be provided via detailed instructions for how to replicate the results, access to a hosted model (e.g., in the case of a large language model), releasing of a model checkpoint, or other means that are appropriate to the research performed.
        \item While NeurIPS does not require releasing code, the conference does require all submissions to provide some reasonable avenue for reproducibility, which may depend on the nature of the contribution. For example
        \begin{enumerate}
            \item If the contribution is primarily a new algorithm, the paper should make it clear how to reproduce that algorithm.
            \item If the contribution is primarily a new model architecture, the paper should describe the architecture clearly and fully.
            \item If the contribution is a new model (e.g., a large language model), then there should either be a way to access this model for reproducing the results or a way to reproduce the model (e.g., with an open-source dataset or instructions for how to construct the dataset).
            \item We recognize that reproducibility may be tricky in some cases, in which case authors are welcome to describe the particular way they provide for reproducibility. In the case of closed-source models, it may be that access to the model is limited in some way (e.g., to registered users), but it should be possible for other researchers to have some path to reproducing or verifying the results.
        \end{enumerate}
    \end{itemize}

\item {\bf Open access to data and code}
    \item[] Question: Does the paper provide open access to the data and code, with sufficient instructions to faithfully reproduce the main experimental results, as described in supplemental material?
    \item[] Answer: \answerYes{}
    \item[] Justification: The code, preprocessing, evaluation scripts, and hyperparameters will be released (Appendix~\ref{app:repro}, Appendix~\ref{app:private}) , except for one core dataset (the Beach Visitor Count dataset), which cannot be made public due to data-sharing restrictions. Reproducibility is therefore partial rather than fully open.
    \item[] Guidelines:
    \begin{itemize}
        \item The answer \answerNA{} means that paper does not include experiments requiring code.
        \item Please see the NeurIPS code and data submission guidelines (\url{https://neurips.cc/public/guides/CodeSubmissionPolicy}) for more details.
        \item While we encourage the release of code and data, we understand that this might not be possible, so \answerNo{} is an acceptable answer. Papers cannot be rejected simply for not including code, unless this is central to the contribution (e.g., for a new open-source benchmark).
        \item The instructions should contain the exact command and environment needed to run to reproduce the results. See the NeurIPS code and data submission guidelines (\url{https://neurips.cc/public/guides/CodeSubmissionPolicy}) for more details.
        \item The authors should provide instructions on data access and preparation, including how to access the raw data, preprocessed data, intermediate data, and generated data, etc.
        \item The authors should provide scripts to reproduce all experimental results for the new proposed method and baselines. If only a subset of experiments are reproducible, they should state which ones are omitted from the script and why.
        \item At submission time, to preserve anonymity, the authors should release anonymized versions (if applicable).
        \item Providing as much information as possible in supplemental material (appended to the paper) is recommended, but including URLs to data and code is permitted.
    \end{itemize}

\item {\bf Experimental setting/details}
    \item[] Question: Does the paper specify all the training and test details (e.g., data splits, hyperparameters, how they were chosen, type of optimizer) necessary to understand the results?
    \item[] Answer: \answerYes{}
    \item[] Justification: The experimental sections and appendix describe the model backbones, shared and model-specific hyperparameters, optimizer and scheduler, early stopping, prediction horizons, data splits, and validation-based hyperparameter selection (Section~\ref{sec:experiments}, Section~\ref{sec:models}, Appendix~\ref{app:data_impl}, and Appendix~\ref{app:repro}).
    \item[] Guidelines:
    \begin{itemize}
        \item The answer \answerNA{} means that the paper does not include experiments.
        \item The experimental setting should be presented in the core of the paper to a level of detail that is necessary to appreciate the results and make sense of them.
        \item The full details can be provided either with the code, in appendix, or as supplemental material.
    \end{itemize}

\item {\bf Experiment statistical significance}
    \item[] Question: Does the paper report error bars suitably and correctly defined or other appropriate information about the statistical significance of the experiments?
    \item[] Answer: \answerYes{}
    \item[] Justification: We report extensive comparisons, ablations, and per-channel analyses with error bars for the main results.
    \item[] Guidelines:
    \begin{itemize}
        \item The answer \answerNA{} means that the paper does not include experiments.
        \item The authors should answer \answerYes{} if the results are accompanied by error bars, confidence intervals, or statistical significance tests, at least for the experiments that support the main claims of the paper.
        \item The factors of variability that the error bars are capturing should be clearly stated (for example, train/test split, initialization, random drawing of some parameter, or overall run with given experimental conditions).
        \item The method for calculating the error bars should be explained (closed form formula, call to a library function, bootstrap, etc.)
        \item The assumptions made should be given (e.g., Normally distributed errors).
        \item It should be clear whether the error bar is the standard deviation or the standard error of the mean.
        \item It is OK to report 1-sigma error bars, but one should state it. The authors should preferably report a 2-sigma error bar than state that they have a 96\% CI, if the hypothesis of Normality of errors is not verified.
        \item For asymmetric distributions, the authors should be careful not to show in tables or figures symmetric error bars that would yield results that are out of range (e.g., negative error rates).
        \item If error bars are reported in tables or plots, the authors should explain in the text how they were calculated and reference the corresponding figures or tables in the text.
    \end{itemize}

\item {\bf Experiments compute resources}
    \item[] Question: For each experiment, does the paper provide sufficient information on the computer resources (type of compute workers, memory, time of execution) needed to reproduce the experiments?
    \item[] Answer: \answerYes{}
    \item[] Justification: We report the software stack, GPU/CPU type, GPU memory, and per-epoch training-time overheads for several datasets and horizons, which together provide practical compute requirements for reproduction (Section~\ref{sec:compute}, Appendix~\ref{app:compute}).
    \item[] Guidelines:
    \begin{itemize}
        \item The answer \answerNA{} means that the paper does not include experiments.
        \item The paper should indicate the type of compute workers CPU or GPU, internal cluster, or cloud provider, including relevant memory and storage.
        \item The paper should provide the amount of compute required for each of the individual experimental runs as well as estimate the total compute. 
        \item The paper should disclose whether the full research project required more compute than the experiments reported in the paper (e.g., preliminary or failed experiments that didn't make it into the paper). 
    \end{itemize}
    
\item {\bf Code of ethics}
    \item[] Question: Does the research conducted in the paper conform, in every respect, with the NeurIPS Code of Ethics \url{https://neurips.cc/public/EthicsGuidelines}?
    \item[] Answer: \answerYes{}
    \item[] Justification: The work uses aggregated, anonymised, opt-in mobility data for the private dataset and includes an impact statement discussing risks, deployment caveats, and mitigation measures, with no apparent conflict with the NeurIPS Code of Ethics (Section~\ref{sec:impact}, Appendix~\ref{app:private}).
    \item[] Guidelines:
    \begin{itemize}
        \item The answer \answerNA{} means that the authors have not reviewed the NeurIPS Code of Ethics.
        \item If the authors answer \answerNo, they should explain the special circumstances that require a deviation from the Code of Ethics.
        \item The authors should make sure to preserve anonymity (e.g., if there is a special consideration due to laws or regulations in their jurisdiction).
    \end{itemize}

\item {\bf Broader impacts}
    \item[] Question: Does the paper discuss both potential positive societal impacts and negative societal impacts of the work performed?
    \item[] Answer: \answerYes{}
    \item[] Justification: The impact statement describes beneficial uses such as safer resource allocation and contingency planning, while also discussing risks from automated decision-making, sensor bias, and deployment under shift, along with mitigation recommendations (Section~\ref{sec:impact}).
    \item[] Guidelines:
    \begin{itemize}
        \item The answer \answerNA{} means that there is no societal impact of the work performed.
        \item If the authors answer \answerNA{} or \answerNo, they should explain why their work has no societal impact or why the paper does not address societal impact.
        \item Examples of negative societal impacts include potential malicious or unintended uses (e.g., disinformation, generating fake profiles, surveillance), fairness considerations (e.g., deployment of technologies that could make decisions that unfairly impact specific groups), privacy considerations, and security considerations.
        \item The conference expects that many papers will be foundational research and not tied to particular applications, let alone deployments. However, if there is a direct path to any negative applications, the authors should point it out. For example, it is legitimate to point out that an improvement in the quality of generative models could be used to generate Deepfakes for disinformation. On the other hand, it is not needed to point out that a generic algorithm for optimizing neural networks could enable people to train models that generate Deepfakes faster.
        \item The authors should consider possible harms that could arise when the technology is being used as intended and functioning correctly, harms that could arise when the technology is being used as intended but gives incorrect results, and harms following from (intentional or unintentional) misuse of the technology.
        \item If there are negative societal impacts, the authors could also discuss possible mitigation strategies (e.g., gated release of models, providing defenses in addition to attacks, mechanisms for monitoring misuse, mechanisms to monitor how a system learns from feedback over time, improving the efficiency and accessibility of ML).
    \end{itemize}
    
\item {\bf Safeguards}
    \item[] Question: Does the paper describe safeguards that have been put in place for responsible release of data or models that have a high risk for misuse (e.g., pre-trained language models, image generators, or scraped datasets)?
    \item[] Answer: \answerNA{}
    \item[] Justification: The paper does not release a high-risk generative model or dataset. The only non-public dataset remains restricted rather than openly released due to data-sharing restrictions, so special release safeguards are not central to this work (Appendix~\ref{app:private}).
    \item[] Guidelines:
    \begin{itemize}
        \item The answer \answerNA{} means that the paper poses no such risks.
        \item Released models that have a high risk for misuse or dual-use should be released with necessary safeguards to allow for controlled use of the model, for example by requiring that users adhere to usage guidelines or restrictions to access the model or implementing safety filters. 
        \item Datasets that have been scraped from the Internet could pose safety risks. The authors should describe how they avoided releasing unsafe images.
        \item We recognize that providing effective safeguards is challenging, and many papers do not require this, but we encourage authors to take this into account and make a best faith effort.
    \end{itemize}

\item {\bf Licenses for existing assets}
    \item[] Question: Are the creators or original owners of assets (e.g., code, data, models), used in the paper, properly credited and are the license and terms of use explicitly mentioned and properly respected?
    \item[] Answer: \answerYes{}
    \item[] Justification: All existing assets are credited via citation. The codebase builds on the Time Series Library (TSLib, MIT License)~\cite{wang2024tssurvey}, and we use PyTorch (BSD-style license). Public benchmark datasets (ETT, Electricity, Exchange, Traffic, Weather, PEMS) are used under the licenses provided by their respective releases and are cited with their original sources (Appendix~\ref{app:public}). The Melbourne Pedestrian Counting data is provided by the City of Melbourne under its open-data terms (Creative Commons Attribution 4.0) and is credited in Appendix~\ref{app:melbourne_ped}. Baseline loss implementations (TILDE-Q, PS, DBLoss) follow author-recommended settings and are cited in Appendix~\ref{app:loss_baselines}.
    \item[] Guidelines:
    \begin{itemize}
        \item The answer \answerNA{} means that the paper does not use existing assets.
        \item The authors should cite the original paper that produced the code package or dataset.
        \item The authors should state which version of the asset is used and, if possible, include a URL.
        \item The name of the license (e.g., CC-BY 4.0) should be included for each asset.
        \item For scraped data from a particular source (e.g., website), the copyright and terms of service of that source should be provided.
        \item If assets are released, the license, copyright information, and terms of use in the package should be provided. For popular datasets, \url{paperswithcode.com/datasets} has curated licenses for some datasets. Their licensing guide can help determine the license of a dataset.
        \item For existing datasets that are re-packaged, both the original license and the license of the derived asset (if it has changed) should be provided.
        \item If this information is not available online, the authors are encouraged to reach out to the asset's creators.
    \end{itemize}

\item {\bf New assets}
    \item[] Question: Are new assets introduced in the paper well documented and is the documentation provided alongside the assets?
    \item[] Answer: \answerYes{}
    \item[] Justification: The new assets introduced by this paper, namely the APAL loss implementation, the peak-critical evaluation protocol, and the training, evaluation, and plotting scripts, are released as supplementary material with documentation covering dependencies, usage commands, dataset preparation, and hyperparameter configurations (Appendix~\ref{app:repro}, Appendix~\ref{app:data_impl}). The Beach Visitor Count dataset is not released due to data-sharing restrictions and is documented as such (Appendix~\ref{app:private}). The Melbourne Pedestrian Count dataset used in the study is already publicly available, and we only use a production-ready subset of it. The preprocessing, sensor selection, and train/validation/test splits used to construct this subset are fully documented in Appendix~\ref{app:melbourne_ped} and Appendix~\ref{app:data_impl}, alongside the released code in the supplementary material. The processed subset is also included in the supplementary material for direct reproducibility. The code will be released upon acceptance of the paper.
    \item[] Guidelines:
    \begin{itemize}
        \item The answer \answerNA{} means that the paper does not release new assets.
        \item Researchers should communicate the details of the dataset\slash code\slash model as part of their submissions via structured templates. This includes details about training, license, limitations, etc. 
        \item The paper should discuss whether and how consent was obtained from people whose asset is used.
        \item At submission time, remember to anonymize your assets (if applicable). You can either create an anonymized URL or include an anonymized zip file.
    \end{itemize}

\item {\bf Crowdsourcing and research with human subjects}
    \item[] Question: For crowdsourcing experiments and research with human subjects, does the paper include the full text of instructions given to participants and screenshots, if applicable, as well as details about compensation (if any)? 
    \item[] Answer: \answerNA{}
    \item[] Justification: The study does not conduct crowdsourcing experiments or direct research with human participants.
    \item[] Guidelines:
    \begin{itemize}
        \item The answer \answerNA{} means that the paper does not involve crowdsourcing nor research with human subjects.
        \item Including this information in the supplemental material is fine, but if the main contribution of the paper involves human subjects, then as much detail as possible should be included in the main paper. 
        \item According to the NeurIPS Code of Ethics, workers involved in data collection, curation, or other labor should be paid at least the minimum wage in the country of the data collector. 
    \end{itemize}

\item {\bf Institutional review board (IRB) approvals or equivalent for research with human subjects}
    \item[] Question: Does the paper describe potential risks incurred by study participants, whether such risks were disclosed to the subjects, and whether Institutional Review Board (IRB) approvals (or an equivalent approval/review based on the requirements of your country or institution) were obtained?
    \item[] Answer: \answerNA{}
    \item[] Justification: The work does not involve direct human-subject experiments or interventions by the authors, so IRB-style approval reporting is not applicable to the study as presented.
    \item[] Guidelines:
    \begin{itemize}
        \item The answer \answerNA{} means that the paper does not involve crowdsourcing nor research with human subjects.
        \item Depending on the country in which research is conducted, IRB approval (or equivalent) may be required for any human subjects research. If you obtained IRB approval, you should clearly state this in the paper. 
        \item We recognize that the procedures for this may vary significantly between institutions and locations, and we expect authors to adhere to the NeurIPS Code of Ethics and the guidelines for their institution. 
        \item For initial submissions, do not include any information that would break anonymity (if applicable), such as the institution conducting the review.
    \end{itemize}

\item {\bf Declaration of LLM usage}
    \item[] Question: Does the paper describe the usage of LLMs if it is an important, original, or non-standard component of the core methods in this research? Note that if the LLM is used only for writing, editing, or formatting purposes and does \emph{not} impact the core methodology, scientific rigor, or originality of the research, declaration is not required.
    \item[] Answer: \answerNA{}
    \item[] Justification: The core methodology is a loss function for time-series forecasting and does not use LLMs as part of the scientific method or experimental pipeline.
    \item[] Guidelines:
    \begin{itemize}
        \item The answer \answerNA{} means that the core method development in this research does not involve LLMs as any important, original, or non-standard components.
        \item Please refer to our LLM policy in the NeurIPS handbook for what should or should not be described.
    \end{itemize}

\end{enumerate}

\end{document}